\pdfoutput=1
\documentclass[11pt]{article}
\usepackage{bbding}
\usepackage[]{emnlp2022}
\usepackage{times}
\usepackage{latexsym}
\usepackage{float}
\usepackage{pdfpages}

\usepackage[T1]{fontenc}
\usepackage{times}
\usepackage{pdfpages}
\usepackage{latexsym}
\usepackage{graphicx}
\usepackage{multirow}
\usepackage{enumitem}
\usepackage[utf8]{inputenc}
\usepackage{microtype}
\usepackage{amsmath}
\usepackage[normalem]{ulem}
\usepackage{subfigure}
\usepackage{makecell}
\usepackage{graphicx} 
\usepackage{xcolor}
\usepackage{amsmath,amssymb,booktabs}
\usepackage{microtype}

\usepackage{enumitem}
\usepackage{comment}

\usepackage{xparse}
\NewDocumentCommand{\heng}
{ mO{} }{\textcolor{red}{\textsuperscript{\textit{Heng}}\textsf{\textbf{\small[#1]}}}}

\newif\ifshowcomment
    \showcommenttrue

\ifshowcomment
    \newcommand{\ganqu}[1]{\textcolor{purple}{[{ganqu: #1}]}}
    \newcommand{\yang}[1]{\textcolor{blue}{[yang: #1]}}

\else
    \newcommand{\todo}[1]{}
    \newcommand{\ganqu}[1]{}
    \newcommand{\yang}[1]{}

    \newcommand{\focus}[1]{}
\fi

\title{\includegraphics[width=0.04\textwidth]{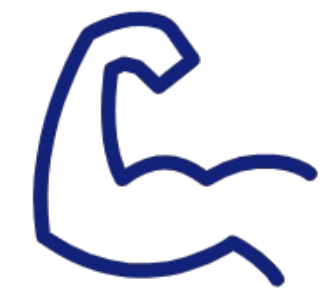}From Adversarial Arms Race to Model-centric Evaluation\includegraphics[width=0.04\textwidth]{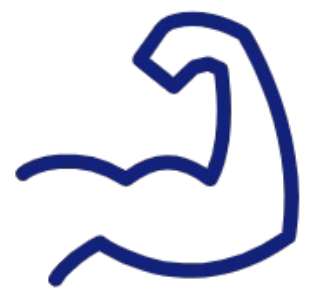} \\ Motivating a Unified Automatic Robustness Evaluation Framework }

  \author{
Yangyi Chen$^{1,2}$\thanks{\ \ Indicates equal contribution. Work done during internship at Tsinghua University.}\hspace{0.3em},
Hongcheng Gao$^{1,3*}$,
Ganqu Cui$^{1*}$,
Lifan Yuan$^{1,4}$, Dehan Kong$^{5}$, Hanlu Wu$^{5}$\\
{\bf Ning Shi$^{5}$, Bo Yuan$^{5}$, Longtao Huang$^{5}$, Hui Xue$^{5}$, Zhiyuan Liu$^{1,6}$\thanks{\ \ Corresponding Author.}, Maosong Sun$^{1,6}$\footnotemark[2], Heng Ji$^{2}$
}
\\ 
$^{1}$NLP Group, DCST, IAI, BNRIST, Tsinghua University, Beijing \\
$^{2}$UIUC
$^{3}$Chongqing University 
$^{4}$HUST
$^{5}$Alibaba Group\\
$^{6}$ Jiangsu Collaborative Innovation Center for Language Ability, Jiangsu Normal University, Xuzhou\\
{\tt yangyic3@illinois.edu,}
{\tt gaohongcheng2000@gmail.com}
\\
{\tt cgq22@mails.tsinghua.edu.cn,}
{\tt \{liuzy,sms\}@tsinghua.edu.cn} 
}

\begin{document}
\maketitle
\begin{abstract}

Textual adversarial attacks can discover models' weaknesses by adding semantic-preserved but misleading perturbations to the inputs. The long-lasting adversarial attack-and-defense arms race in Natural Language Processing (NLP) is algorithm-centric, providing valuable techniques for automatic robustness evaluation. However, the existing practice of robustness evaluation may exhibit issues of incomprehensive evaluation, impractical evaluation protocol, and invalid adversarial samples. In this paper, we aim to set up a unified automatic robustness evaluation framework, shifting towards model-centric evaluation to further exploit the advantages of adversarial attacks. To address the above challenges, we first determine robustness evaluation dimensions based on model capabilities and specify the reasonable algorithm to generate adversarial samples for each dimension. Then we establish the evaluation protocol, including evaluation settings and metrics, under realistic demands. Finally, we use the perturbation degree of adversarial samples to control the sample validity. We implement a toolkit \textbf{RobTest} that realizes our automatic robustness evaluation framework. In our experiments, we conduct a robustness evaluation of RoBERTa models to demonstrate the effectiveness of our evaluation framework, and further show the rationality of each component in the framework.
The code will be made public at \url{https://github.com/thunlp/RobTest}. 

\end{abstract}
\section{Introduction}

\begin{figure}[ht]
    \centering
    \includegraphics[width=0.8\linewidth]{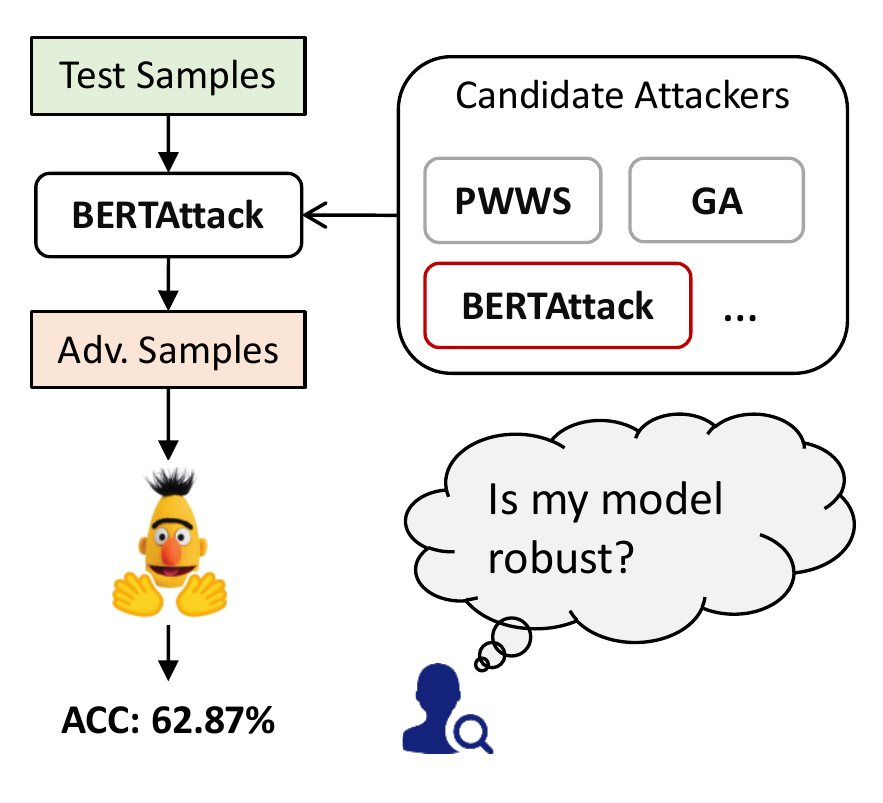}
    \caption{The original evaluation pipeline. The attacker is usually selected by intuition and practitioners get little information from scores.}
    \label{fig:hello_figure}
    \vspace{-20pt}
\end{figure}

Pre-trained language models (PLMs) are vulnerable to textual adversarial attacks that fool the models by adding semantic-preserved perturbations to the inputs~\citep{zhang2020adversarial}.
Compared to the static evaluation benchmarks~\cite{wang2018glue, wang2019superglue}, attack methods can continually generate diverse adversarial samples to reveal models' weaknesses, rendering a more comprehensive and rigorous model evaluation. 
Previous work explores adversarial NLP in both the attack~\citep{gao2018black, alzantot-etal-2018-generating} and the defense~\citep{mozes-etal-2021-frequency, huang-etal-2019-achieving} sides, leading to a long-lasting adversarial arms race.

\looseness=-1
The arms race is algorithm-centric.
It continually motivates stronger attack and defense methods to explore and fix models' weaknesses, providing useful techniques for robustness evaluation.
However, existing work on model robustness evaluation naturally follows the previous evaluation practice, and doesn't fully consider the real-world needs of robustness evaluation~\citep{zeng-etal-2021-openattack, wang-etal-2021-textflint, goel2021robustness} (See Figure~\ref{fig:hello_figure}).
We identify three weaknesses in previous robustness evaluation:
(1) Relying on a single attack method~\citep{zang-etal-2020-word} or static challenging datasets~\citep{nie2019adversarial, wang2021adversarial}, which can only measure a limited number of aspects of models' capabilities; 
(2) Directly inheriting the evaluation settings and metrics in the arms race era, which may result in impractical evaluation~\citep{zeng-etal-2021-openattack, morris-etal-2020-textattack}; 
(3) Designing invalid adversarial sample\footnote{Detailed explanation for validity is in Appendix~\ref{appendix:validity}.} filtering rules based on certain thresholds (e.g., sentence similarity), which cannot generalize to all kinds of adversarial samples~\citep{wang-etal-2021-textflint, zeng-etal-2021-openattack}.



\begin{figure}[hpt]
    \centering
    \includegraphics[width=\linewidth]{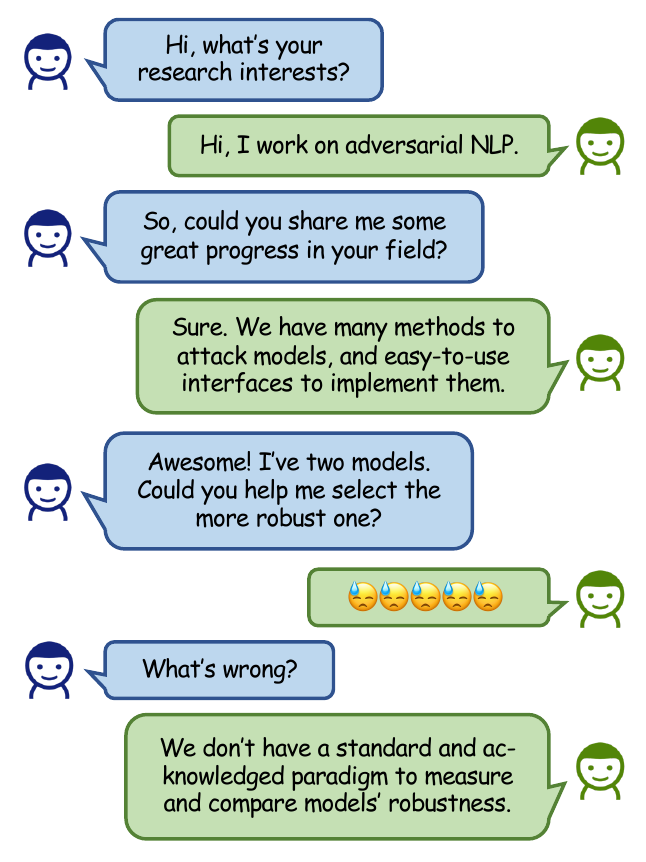}
      \caption{The current dilemma in adversarial NLP.}
    \label{fig:delimma}

\end{figure}

Thus, we propose to shift towards the model-centric evaluation, which should satisfy the following characteristics accordingly:
(1) \textbf{Comprehensively} measuring NLP models' robustness;
(2) Establishing a \textbf{reasonable} evaluation protocol considering practical scenarios;
(3) Filtering out invalid adversarial samples for \textbf{reliable} robustness estimation.
Given these challenges, a standard and acknowledged framework for employing adversarial attacks to automatically measure and compare NLP models' robustness is lacking (See Figure~\ref{fig:delimma}).


%


%
In this paper, we motivate a unified model-centric automatic robustness evaluation framework based on the foundation of the adversarial arms race. 
To achieve \textbf{comprehensive} evaluation, we define eight robustness dimensions from top to down, constituting a evaluation of multi-dimensional robustness towards sentence-level, word-level, and char-level transformations.
For each robustness dimension, we specify the concrete algorithm to generate adversarial samples.
Then we set up a \textbf{reasonable} evaluation protocol by specifying evaluation settings and metrics under realistic demands. 
%
Finally, we rely on the perturbation degree to control the validity of generated adversarial samples for more \textbf{reliable} robustness evaluation.
Our intuition is that adversarial samples with smaller perturbation degrees are more likely to be valid, which is justified through human annotation experiments.



\begin{figure}[hpt]
    \centering
    \includegraphics[width=\linewidth]{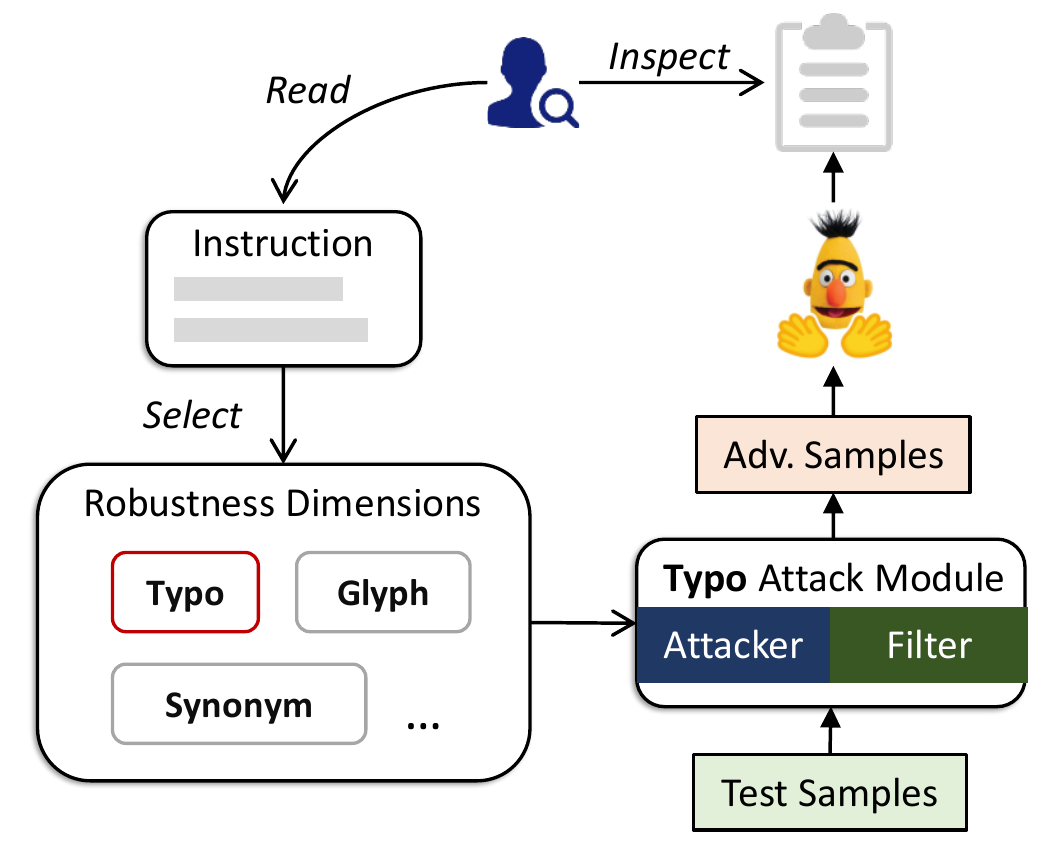}
       \caption{Our robustness evaluation framework. The ``Instruction'' refers to the written guidance for robustness evaluation.}
    \label{fig:conceptual_framework}

    \vspace{-15pt}
\end{figure}

We implement a toolkit \textbf{RobTest} to realize our robustness evaluation framework (See Figure~\ref{fig:conceptual_framework}).
We highlight four core features in RobTest, including basic adversarial attack methods, robustness report generation, general user instructions, and adversarial data augmentation.
In experiments, we use RobTest to measure the robustness of RoBERTa models~\cite{liu2019roberta} to demonstrate the effectiveness of our evaluation framework in addressing the core challenges.
Further, we show the rationality of each component in our robustness evaluation framework through detailed analysis.

\begin{table*}[t]
\centering
\resizebox{0.85\textwidth}{!}{
\begin{tabular}{llccl}
\toprule
Granularity                     & Dimension   & \multicolumn{1}{l}{General?} & \multicolumn{1}{l}{Malicious?} & \multicolumn{1}{c}{Case}                                               \\ \midrule
\multirow{3}{*}{Char-level}     & Typo        & Yes                          & Yes                            & I watch a smart, \color[HTML]{FE0000} swet adn \color[HTML]{000000} playful romantic comedy.                     \\
                                & Glyph       & Yes                          & Yes                            & I watch a \color[HTML]{FE0000}ŝmârt\color[HTML]{000000}, sweet and playful \color[HTML]{FE0000}romaňtîĉ \color[HTML]{000000}  comedy.                         \\
                                 & Phonetic       & Yes                          & Yes                            & I \color[HTML]{FE0000}wotch \color[HTML]{000000} a smart, sweet and playful \color[HTML]{FE0000}romentic  \color[HTML]{000000}comedy.                         \\
                                 \midrule
\multirow{3}{*}{Word-level}     & Synonym     & Yes                          & No                             & I watch a smart, sweet and \color[HTML]{FE0000} naughty \color[HTML]{000000} romantic comedy.                    \\
                                & Contextual  & Yes                          & No                             & \color[HTML]{FE0000} We \color[HTML]{000000} watch a smart, sweet and playful romantic \color[HTML]{FE0000} teleplay\color[HTML]{000000}.                 \\
                                & Inflection  & Yes                          & No                             & I \color[HTML]{FE0000} watched \color[HTML]{000000} a smart, sweet and playful romantic \color[HTML]{FE0000} comedies\color[HTML]{000000}.                \\ \midrule
\multirow{2}{*}{Sentence-level} & Syntax      & Yes                          & No                             & \color[HTML]{FE0000} In my eyes will be a witty, sweet romantic comedy.                    \\
                                & Distraction & No                           & Yes                            & I watch a smart, sweet and playful romantic comedy. \color[HTML]{FE0000} True is not False\color[HTML]{000000}. \\ \bottomrule
\end{tabular}
}
\caption{\label{tab:robustness_dimensions} 
 The robustness dimensions included in our framework. We also attach general and malicious robustness tags to each dimension. The original sentence is ``I watch a smart, sweet and playful romantic comedy.''}
\vspace{-15pt}

\end{table*}

\section{Model-centric Robustness Evaluation}
In this section, we motivate the first model-centric automatic robustness evaluation framework. 
We first define robustness evaluation dimensions and specify corresponding attack algorithms (Sec.~\ref{sec:robust_eval_dimen}).
Then we discuss the evaluation protocol under realistic demands (Sec.~\ref{sec:eval_protocol}). 
Finally, we provide solutions to filter out invalid adversarial samples for more reliable robustness evaluation (Sec.~\ref{sec:reliable_eval}).

\subsection{Robustness Evaluation Dimension}
\label{sec:robust_eval_dimen}

\looseness=-1
\paragraph{Motivation.}
Existing research designs adversarial attacks based on observations~\cite{le2022perturbations} or intuitions~\cite{li-etal-2020-bert-attack} and adopts the proposed method to test the robustness of evaluated models. 
In this procedure, the robustness evaluation is restricted to the specific attack method without considering samples from other potential distributions.
We argue that considering only one single dimension cannot comprehensively describe the models' robustness (See Sec.~\ref{sec:analysis_of_framework} for verification).


\paragraph{Selection Criteria.}
We build our model-centric robustness evaluation framework based on the foundation of adversarial NLP but aim to cover a more comprehensive set of robustness dimensions. 
We integrate previous adversarial attack methods in a systematic way. 
We focus on task-agnostic robustness dimensions\footnote{Task-specific robustness dimensions are also essential, and we leave it for future work.}, and define them from top to down (See Table~\ref{tab:robustness_dimensions}). 
The selection criteria of robustness evaluation dimensions and attack methods are: 
(1) \textbf{Important and practical}: Methods that can reasonably simulate common inputs from real-world users or attackers; 
(2) \textbf{Representative}: Methods that have been studied for a long time in the adversarial arms race stage and have many homogeneous counterparts; 
(3) \textbf{Diversified}: Methods that explore various aspects of model capabilities. 

Note that we don't consider the ``imperceptible perturbations'' requirement in the selection of robustness dimensions, although previous work repeatably emphasizes this requirement~\cite{goodfellow2014explaining, ren-etal-2019-generating, zang-etal-2020-word}.
We give our justification in Appendix~\ref{sec:justify}. 
%

%
%

\paragraph{Dimensions.}
 We start from a high-level categorization, considering char-level, word-level, and sentence-level transformations, differing in the perturbation granularity (See Table~\ref{tab:robustness_dimensions}). 
%
\textbf{Char-level} transformations add perturbations to characters in the word units. 
We include the following dimensions in our framework:
(1) \textbf{Typo}~\cite{li2018textbugger, eger-benz-2020-hero} considers five basic operations to add typos in the inputs, including randomly delete, insert, replace, swap, or repeat one character; 
(2) \textbf{Glyph}~\cite{li2018textbugger, Eger-2019-viper} replaces characters with visually-similar ones;
(3) \textbf{Phonetic}~\citep{le2022perturbations} replaces characters but makes the whole word sound similar to the origin. 
\textbf{Word-level} transformations modify word units as a whole. 
We include the following dimensions in our framework: 
(1) \textbf{Synonym} \cite{ren-etal-2019-generating, zang-etal-2020-word} replaces words with their synonymous substitutes according to external knowledge sources. 
We consider WordNet \cite{miller1995wordnet} and HowNet \cite{dong2003hownet} in our implementation;
(2) \textbf{Contextual} \cite{li-etal-2020-bert-attack, garg2020bae} replaces words with their context-similar substitutes, which are generated by masked language models;
(3) \textbf{Inflection}~\cite{tan2020s} perturbs the inflectional morphology of words.
\textbf{Sentence-level} transformations generate adversarial samples directly from the entire original sentences.
We include the following dimensions in our framework: 
(1) \textbf{Syntax}~\cite{iyyer-etal-2018-adversarial,huang-chang-2021-generating, sun2021aesop} transforms the syntactic patterns of original samples; 
(2) \textbf{Distraction} \cite{naik-etal-2018-stress, ribeiro-etal-2020-beyond,chen2022can} appends some irrelevant contents to the end of sentences. 

\looseness=-1
\paragraph{Malicious \& General Tags.}
For each robustness dimension, we also attach the general or malicious tag to characterize the intended simulated agents.
The general (malicious) tag indicates that the generated samples mainly come from benign users (malicious attackers). 
For example, Synonym and Distraction are representative types of general and malicious dimensions respectively. 
Note that we attach both tags to three char-level transformations since both benign users and malicious attackers can produce these kinds of samples.
%


	

\subsection{Evaluation Protocol}
\label{sec:eval_protocol}

\begin{table*}[t]
\centering
\resizebox{\textwidth}{!}{
\begin{tabular}{ll}
\toprule
Original        & \color[HTML]{000000} I love the way that it took chances and really asks you to take these great leaps of faith and pays off.  \\ \midrule
BERT-Attack~\cite{li-etal-2020-bert-attack} & \color[HTML]{000000} I \color[HTML]{FE0000} hate\color[HTML]{000000}~the way that it took chances and \color[HTML]{FE0000}jesus asking \color[HTML]{000000}  you to take these \color[HTML]{FE0000}grand \color[HTML]{000000} leaps of faith and pays off. \\ \midrule
GA~\cite{alzantot-etal-2018-generating}  &\color[HTML]{000000}  I \color[HTML]{FE0000} screw \color[HTML]{000000} the way that it \color[HTML]{FE0000} read \color[HTML]{000000} chances and really asks you to \color[HTML]{FE0000} remove \color[HTML]{000000} these great leaps of faith and pays off. \\ \midrule
Textbugger~\cite{li2018textbugger} & I \color[HTML]{FE0000} lve \color[HTML]{000000} the way that it took \color[HTML]{FE0000} cances \color[HTML]{000000} and really \color[HTML]{FE0000} a sks \color[HTML]{000000} you to take these \color[HTML]{FE0000} grwat lezps \color[HTML]{000000} of \color[HTML]{FE0000} fith \color[HTML]{000000} and \color[HTML]{FE0000} pay5 \color[HTML]{000000}off. \\ \bottomrule

\end{tabular}
}
\caption{\label{tab:adv_invalid_case} 
Cases of invalid adversarial samples crafted by three popular attack methods. The original label is positive.}
\vspace{-15pt}

\end{table*}

\paragraph{Motivation.}
Previous work in adversarial NLP naturally follows the early attempts~\cite{szegedy2013intriguing, goodfellow2014explaining, liang2017deep,gao2018black} to establish the evaluation protocol.
However, \citet{chen2022should} categorize and summarize four different roles of textual adversarial samples, and argue for a different evaluating protocol for each role. 
In our framework, we reconsider the robustness evaluation protocol when employing adversarial attack methods for model evaluation. 
We first describe the evaluation setting, and then the evaluation metrics in our framework.

\paragraph{Evaluation Setting (available information from evaluated models).}
Most existing attack methods assume the accessibility to confidence scores only~\cite{alzantot-etal-2018-generating, ren-etal-2019-generating,  zang-etal-2020-word, li-etal-2020-bert-attack, chen2021multi}.
We acknowledge the rationality of this assumption since the size of models may become too large nowadays~\cite{radford2019language, brown2020language}, resulting in inefficient evaluation if also requiring the gradients information for adversarial samples generation~\citep{goodfellow2014explaining}. 
However, in practice, we as practitioners mostly have all access to the evaluated models, including the parameters and gradient information, for better robustness evaluation.

Thus, we implement three evaluation settings in our framework, assuming different available information from evaluated models. 
The settings include rule-based, score-based, and gradient-based attacks.
Rule-based attacks don't assume any information from the evaluated models and generate adversarial samples based on pre-defined rules.
Score-based and gradient-based attacks assume access to the confidence scores and gradients information respectively from evaluated models for more rigorous evaluation.
They first compute the saliency maps that give the importance scores to each word for samples and then perform selective perturbations based on the scores.
Specifically, for score-based attacks, we employ the difference in confidence scores when iteratively masking each word as the important score for that word.
For gradient-based attacks, we employ integrated gradient (IG)~\cite{sundararajan2017axiomatic} to compute the saliency map.
IG computes the average gradient along the linear path of varying the input from a baseline value to itself.
Besides, we use greedy search since it can achieve satisfying performance within a reasonable time~\cite{yoo2020searching}. 


\paragraph{Evaluation Metrics.}
Most previous work considers the ``is robust'' problem~\citep{li-etal-2020-bert-attack, li-etal-2021-contextualized, chen2021multi}.
They generate adversarial samples for each original sample and test if at least one of them can successfully attack the evaluated models.
Then the final score is computed as the percentage of samples that are not attacked successfully.
This is the \textbf{worst performance estimation}, requiring models to be robust to all potential adversarial samples in order to score.
In our framework, we introduce the \textbf{average performance estimation} for a more comprehensive robustness evaluation.
Specifically, for each original sample, we compute the percentage of cases that models can correctly classify among all potential adversarial samples.
Then we average over all original samples to get the average performance estimation score.






\subsection{Reliable Robustness Evaluation}
\label{sec:reliable_eval}
\paragraph{Motivation.} 
Previous work chases for higher attack success rate, while the validity of adversarial samples may be sacrificed\footnote{We give a detailed explanation for adversarial samples validity in Appendix~\ref{appendix:validity}.}. 
The consequence of this practice is unreliable and inaccurate robustness evaluation.
We showcase adversarial samples crafted by three popular methods on SST-2~\cite{socher-etal-2013-recursive} in Table~\ref{tab:adv_invalid_case}. 
While all samples successfully flip the predictive label, they are not good choices for robustness evaluation because the ground truth label is changed (e.g., BERT-Attack) or the meaning of the original sentence is changed (e.g., GA, Textbugger).
\citet{morris-etal-2020-reevaluating,wang2021adversarial, hauser2021bert} show that there are many such invalid cases in adversarial samples that successfully mislead models' predictions.
We further conduct a human evaluation to support this conclusion.
We hire annotators to evaluate adversarial samples validity of three representative attack methods, namely contextual-based~\citep{li-etal-2020-bert-attack}, synonym-based~\citep{zang-etal-2020-word}, and typo-based attacks~\citep{karpukhin-etal-2019-training}. 
The results show that on average only \textbf{25.5\%}, \textbf{20.0\%}, and \textbf{31.5\%} generated samples are valid.
Thus, if directly employing original adversarial samples for robustness evaluation, the results are unreliable and don't convey too much useful information to practitioners.




\paragraph{Potential Solutions.}
For reliable robustness evaluation, we need to consider how to ensure the validity of constructed adversarial samples. 
We can approach this problem in two different ways: 
(1) Verify generated adversarial samples; 
(2) Incorporating the validity criterion in robustness evaluation. 
All existing work focuses on verification. 
For example, in the implementation of OpenAttack~\cite{zeng-etal-2021-openattack} and TextFlint~\cite{wang-etal-2021-textflint}, an embedding similarity threshold is set for filtering adversarial samples. 
However, we argue that \textbf{a unified sample selection standard without considering the specific trait of the attack method can not perform effective filtering.} 
For example, consider the adversarial sample crafted by adding typos: ``I love the way that it took \color[HTML]{FE0000} chancs  \color[HTML]{000000} and really asks you to \color[HTML]{FE0000}takke \color[HTML]{000000}  these great leaps of faith and pays off.''  
This sample may be filtered out by the similarity or perplexity threshold due to its unnatural expression.
However, it well simulates the input from real-world users and retains the original meaning, thus should be considered in the evaluation.



%

\paragraph{Our Method.}
In our framework, we consider incorporating the validity criterion into robustness evaluation. 
We hold a basic intuition that there is an inverse correlation between the perturbation degree and the validity of adversarial samples.
Thus, we rely on the perturbation degree to measure the adversarial sample validity. 
Note that the perturbation degree is defined according to the concrete transformation level\footnote{The computational details are described in Appendix~\ref{appendix:compute_degree}.}.
%
We justify our intuition and demonstrate the superiority of this filtering strategy compared to previous heuristic rules (e.g., grammar error, sentence similarity, perplexity) in Sec.~\ref{sec:verify}.

\textbf{We propose to measure models' robustness under the specific attack method in various perturbation degrees and compute a robustness score for each degree. }
The robustness score is the model's worst performance estimation or average performance estimation. 
We put more emphasis on the robustness scores computed at lower perturbation degrees\footnote{Note that the perturbation degree computation methods are different for different dimensions (See Appendix~\ref{appendix:compute_degree}).} and employ the exponentially weighted moving average~\cite{hunter1986exponentially} to compute the final score for each robustness dimension.
%
Formally, we use $\theta_1, \theta_2, ..., \theta_n$ to denote robustness scores computed at $n$ perturbation degrees from high to low. 
Set $\mathcal{V}_1 = \theta_1$. To compute the \textbf{final robustness score $\mathcal{V}_n$}:
\begin{equation}
\label{eq:l_c}
    \mathcal{V}_t = \beta * \mathcal{V}_{t-1} + (1-\beta)*\theta_t, \quad t=2,..., n,
\end{equation}
where $\beta$ controls the weights on scores computed at different degrees.
Empirically, it should be chosen depending on the risk level of the considered task, 
and smaller $\beta$ will more emphasize the importance of evaluation on high-perturbed samples, which is essential for high-stake applications.
In our framework, we set $\beta$=0.5 for demonstration.

\section{RobTest}
We implement an automatic robustness evaluation toolkit named \textbf{RobTest} to realize our proposed framework.
We highlight four features of RobTest. 

\looseness=-1
\paragraph{Basic Adversarial Attack Methods.}
We implement eight attack methods, corresponding to eight robustness evaluation dimensions in our framework. 
We also include three attack types that assume different information available from evaluated models, namely rule-based, score-based, and gradient-based attacks.
RobTest allows practitioners to customize evaluated models and datasets and design new attack methods to test specified robustness dimensions.
Also, it supports the multi-process running of adversarial attacks for efficiency.

\begin{figure*}
\centering
\includegraphics[width=0.77\linewidth]{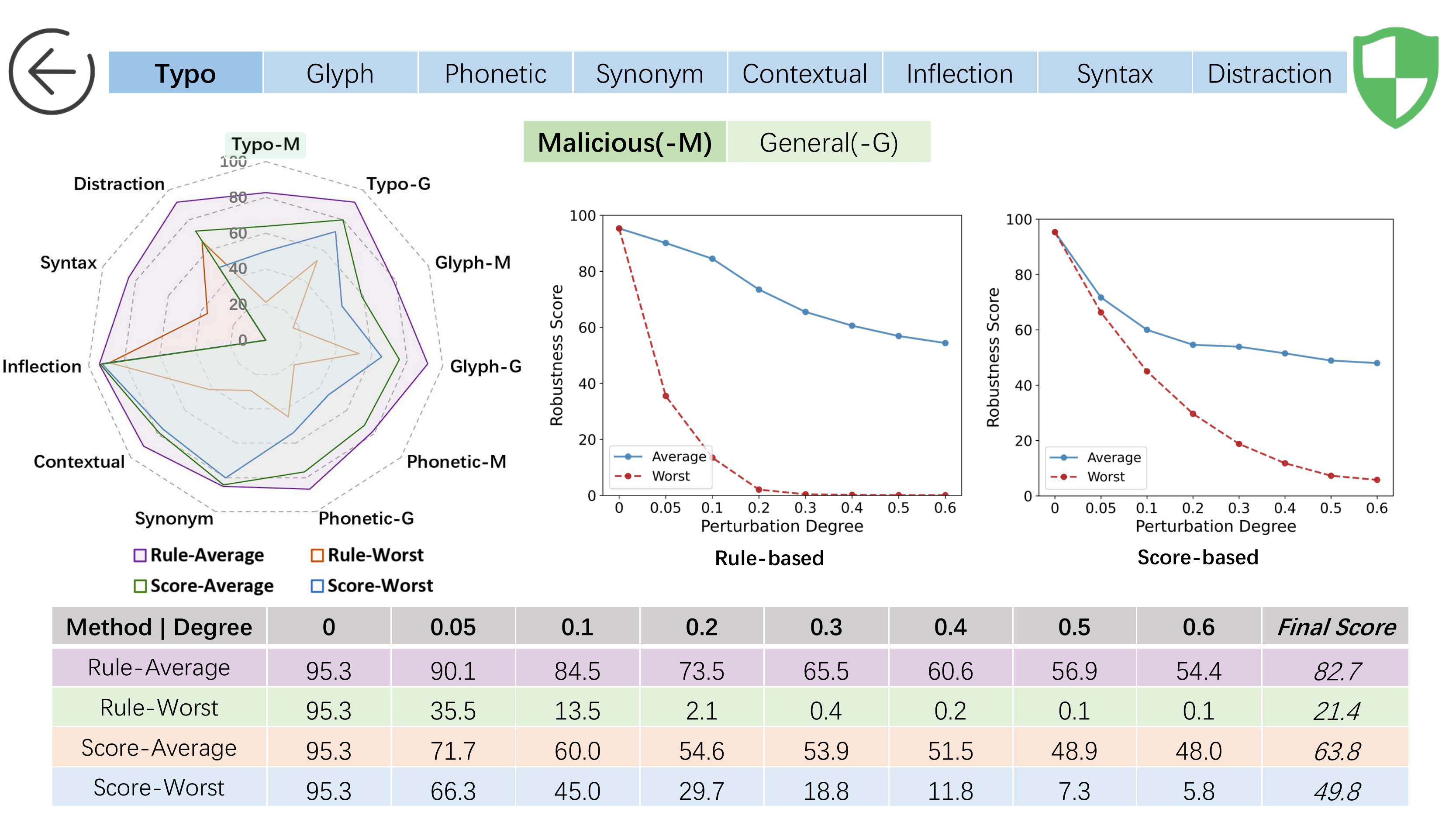}

\caption{\label{fig:report_example} Example of one single page of the robustness report of RoBERTa-base on SST-2, regarding the Typo (Malicious) dimension. The full report is shown in Figure~\ref{fig:robust_base_sst}. We use Rule- and Score- to denote two evaluation settings, and use -Average and -Worst to denote two metrics.}
\vspace{-10pt}
\end{figure*}

\begin{figure}
\centering
\includegraphics[width=0.99\linewidth]{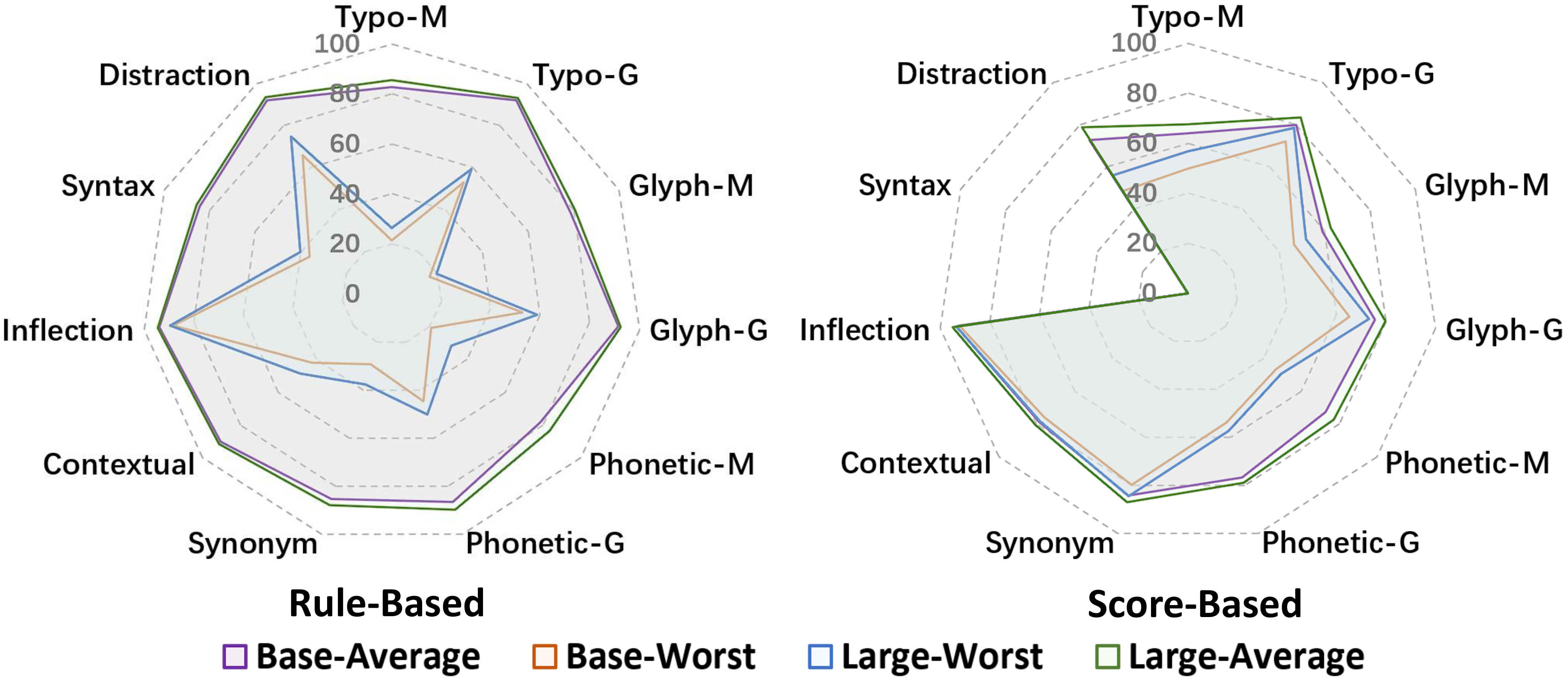}
\caption{\looseness=-1 Radar map to compare the robustness of RoBERTa-base and -large considering all dimensions on SST-2. We use Base- and Large- to denote two models, and other denotations are the same as Figure~\ref{fig:report_example}.} 
\label{fig:comparison} 
\end{figure}

\paragraph{Robustness Report.}
RobTest provides comprehensive robustness reports for evaluated models.
See Figure~\ref{fig:report_example} and Appendix~\ref{sec:rob_report} for examples of single-model robustness reports. 
See Figure~\ref{fig:comparison} and Appendix~\ref{sec:rob_compar_report} for examples of the robustness comparison of the two models.
We further discuss the details of robustness reports in Sec.~\ref{sec:exp}.

\looseness=-1
\paragraph{General Instructions.}
Existing toolkits that implement various attack methods don't provide detailed guidance on how to conduct robustness evaluation~\citep{morris-etal-2020-textattack, zeng-etal-2021-openattack, wang-etal-2021-textflint}.
In RobTest, we provide general instructions for practitioners.
Two kinds of instructions are included: 
(1) How to select appropriate robustness dimensions to evaluate, and which accessibility (e.g., score-based) should be considered.  
We introduce detailed descriptions of all robustness dimensions in RobTest, including the real-world distributions they consider;
%
(2) How to understand the robustness report.
We give detailed explanations for the figures and tables in the report. 

\paragraph{Data Augmentation.}
Practitioners may identify several weak robustness dimensions of evaluated models. 
RobTest supports generating adversarial samples under the specified perturbation degree for data augmentation to improve the robustness.

\section{Experiment} 
\label{sec:exp}
We conduct experiments to demonstrate the effectiveness of our automatic robustness evaluation framework using RobTest.
We aim to show how our framework fulfills the characteristics of model-centric robustness evaluation\footnote{We leave the detailed evaluation and analysis of various model architectures and robustness-enhanced algorithms for future work.}. 

\subsection{Experimental Setting}
\paragraph{Dataset and Evaluated Model.}
In our experiments, we choose the general, common, and application-driven tasks that our task-agnostic robustness dimensions can be applied to\footnote{Task-specific robustness dimensions can be designed for certain tasks, e.g., name entity robustness for reading comprehension~\citep{yan2021robustness}. We leave it for future work.}. 
We consider sentiment analysis, news classification, and hate-speech detection tasks.
We choose SST-2~\cite{socher-etal-2013-recursive}, AG's News~\cite{zhang2015character}, and Jigsaw\footnote{\url{https://www.kaggle.com/c/jigsaw-toxic-comment-classification-challenge}} as evaluation datasets. 
We choose RoBERTa-base and RoBERTa-large~\cite{liu2019roberta} as evaluated models.

\looseness=-1
\paragraph{Evaluation Setting.}
For each dataset, we sample 1,000 samples from the test set for experiments and generate at least 100 testing cases for each sample under each perturbation degree. 
%
%
In pilot experiments, we found no advantage of employing gradient information to generate saliency maps, and thus we only consider rule-based and score-based accessibility in experiments.
Further research is needed for more effective utilization of gradients.

\begin{figure*}[htbp]
\centering

\subfigure[Base-Rule-Average]{
\begin{minipage}[t]{0.24\linewidth}
\centering
\includegraphics[width=1.7in]{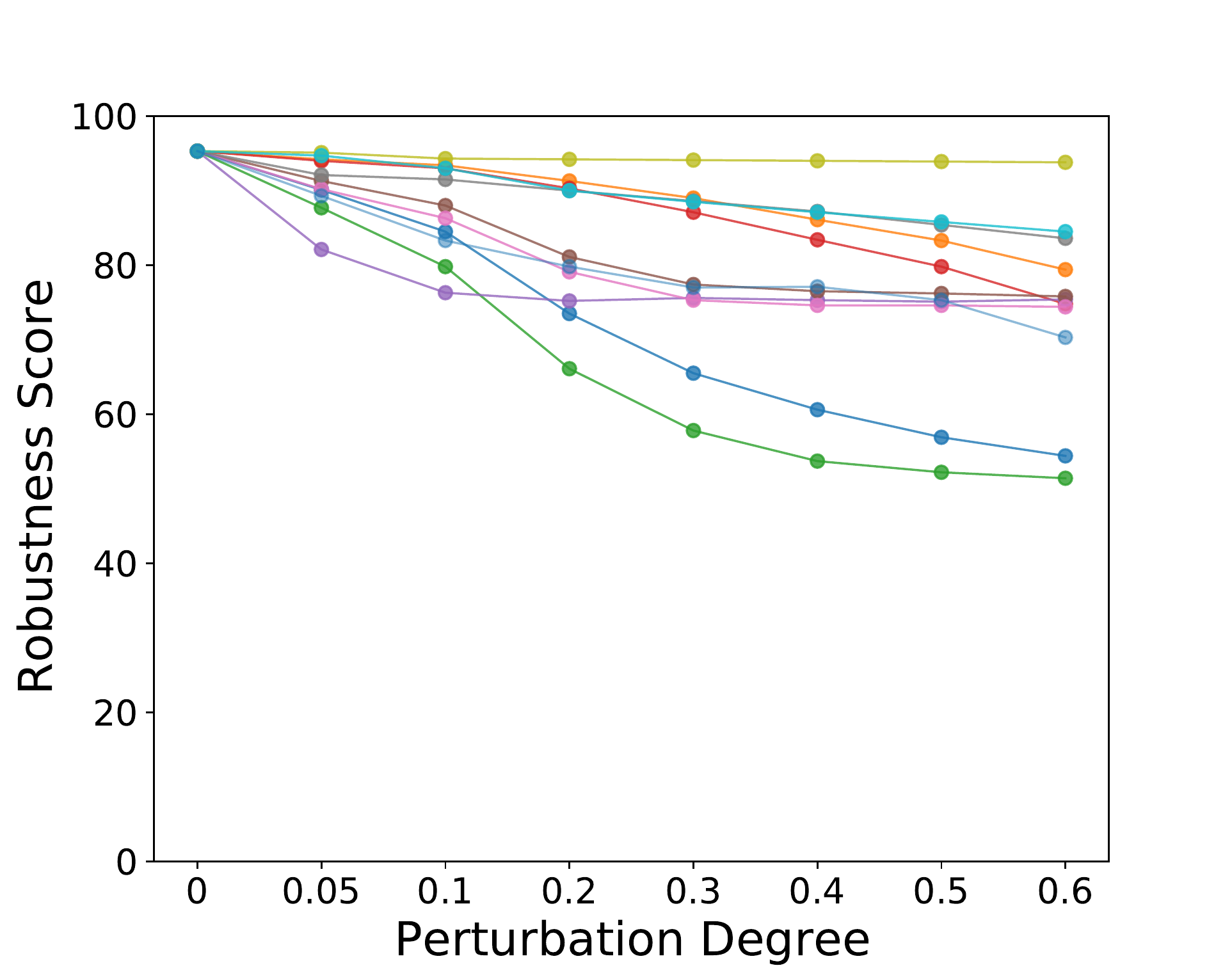}
\end{minipage}%
}%
\subfigure[Base-Rule-Worst]{
\begin{minipage}[t]{0.24\linewidth}
\centering
\includegraphics[width=1.7in]{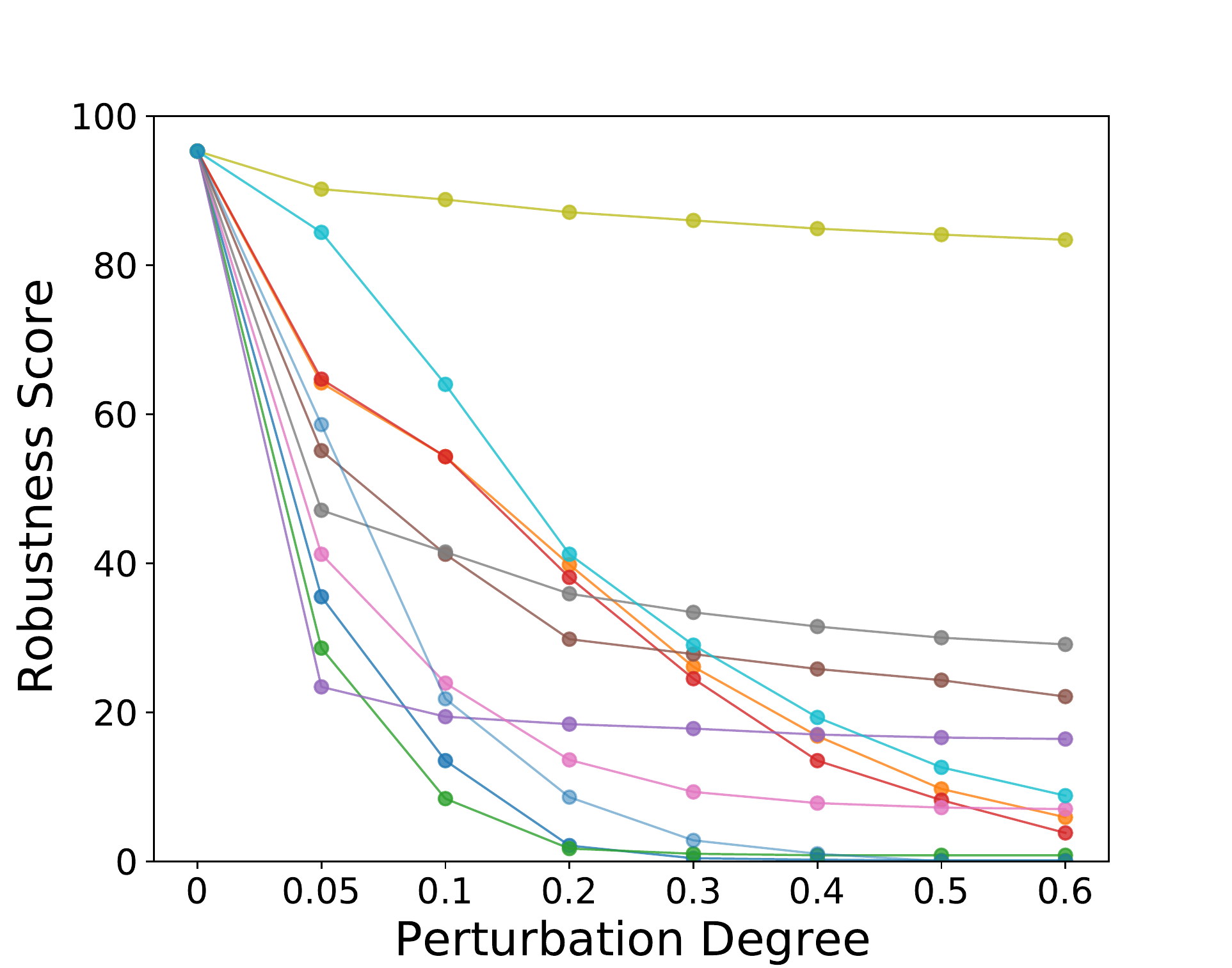}
\end{minipage}%
}%
\subfigure[Large-Rule-Average]{
\begin{minipage}[t]{0.24\linewidth}
\centering
\includegraphics[width=1.7in]{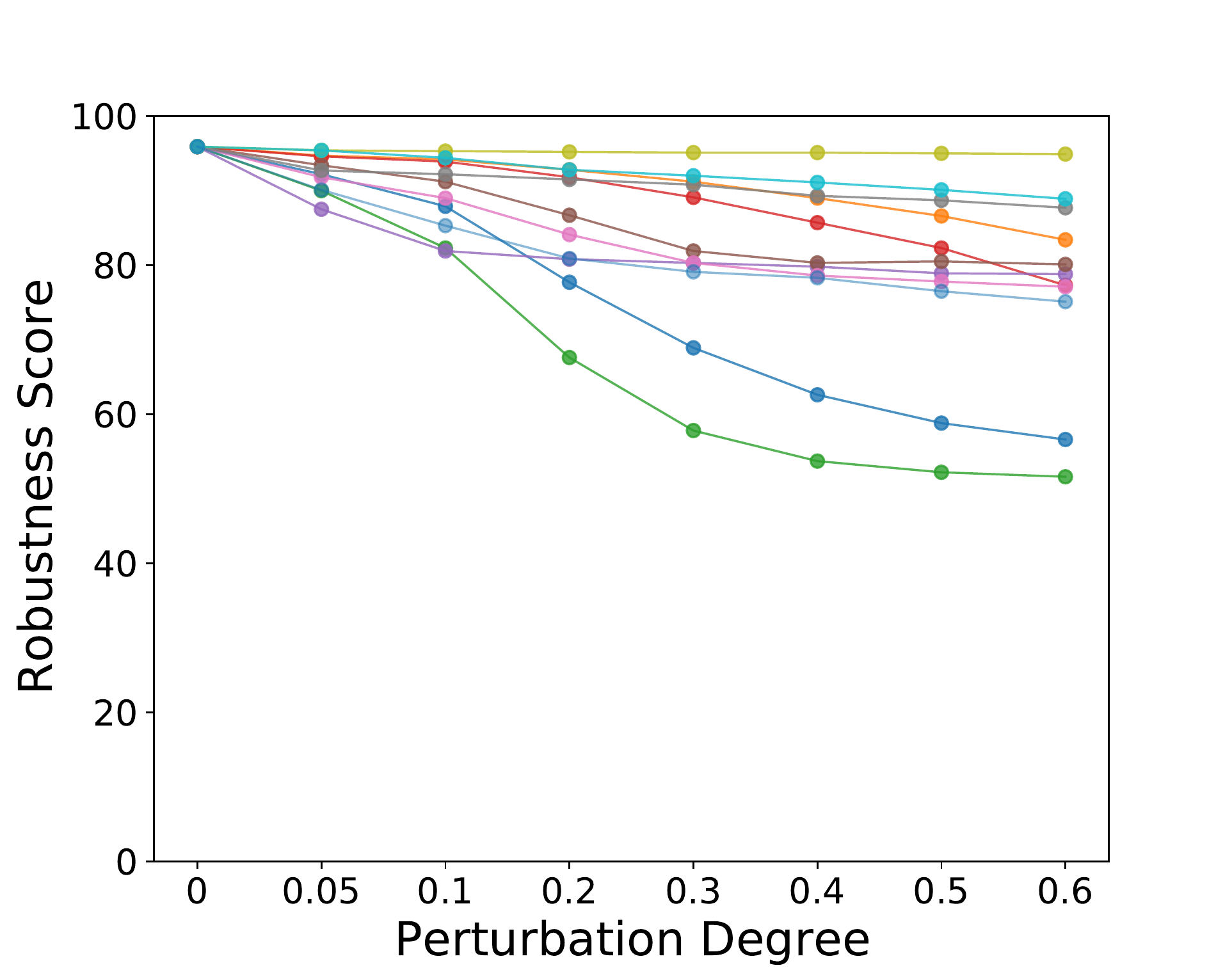}
\end{minipage}%
}%
\subfigure[Large-Rule-Worst]{
\begin{minipage}[t]{0.24\linewidth}
\centering
\includegraphics[width=1.7in]{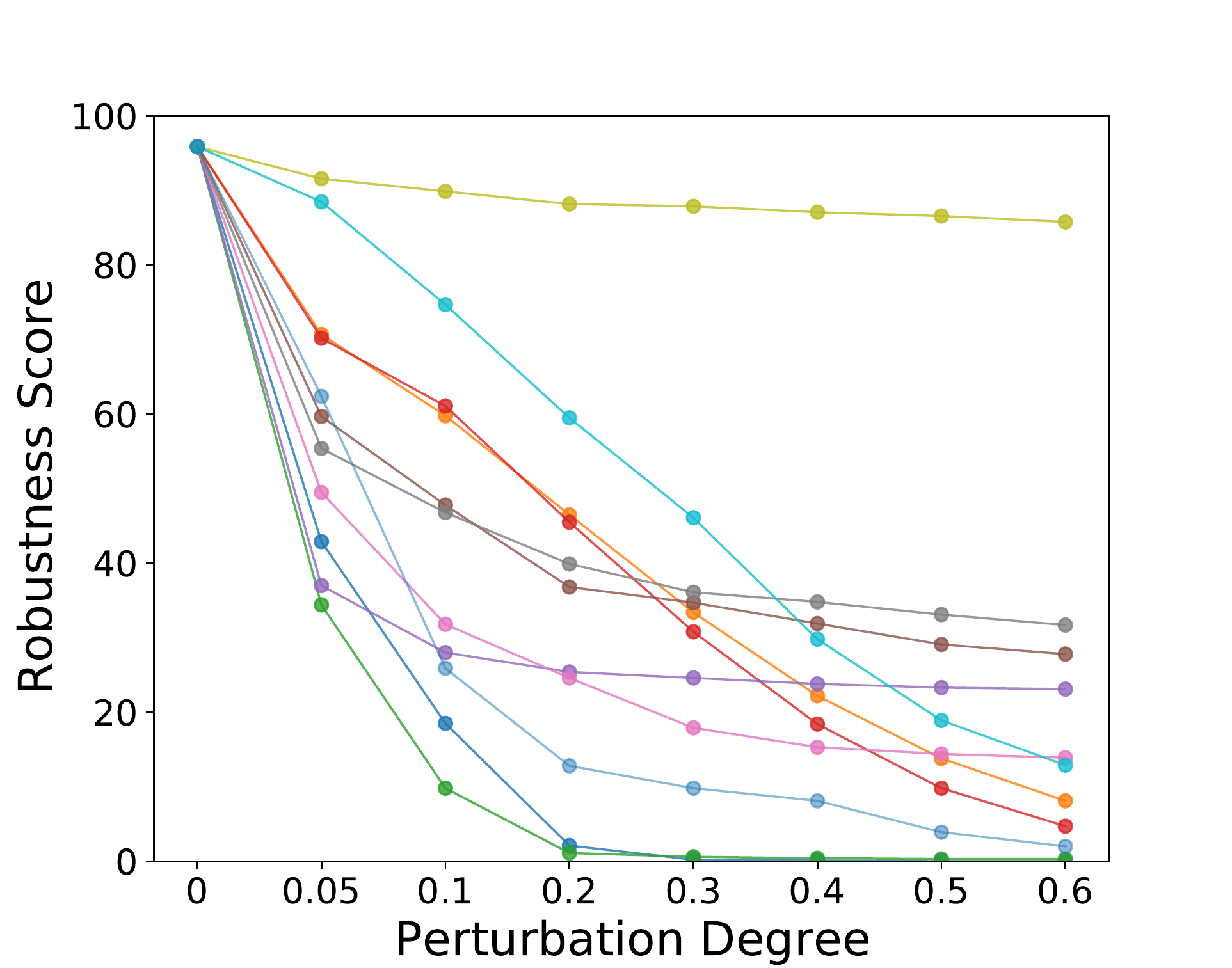}
\end{minipage}%
}%
    \quad             
\subfigure[Base-Score-Average]{
\begin{minipage}[t]{0.24\linewidth}
\centering
\includegraphics[width=1.7in]{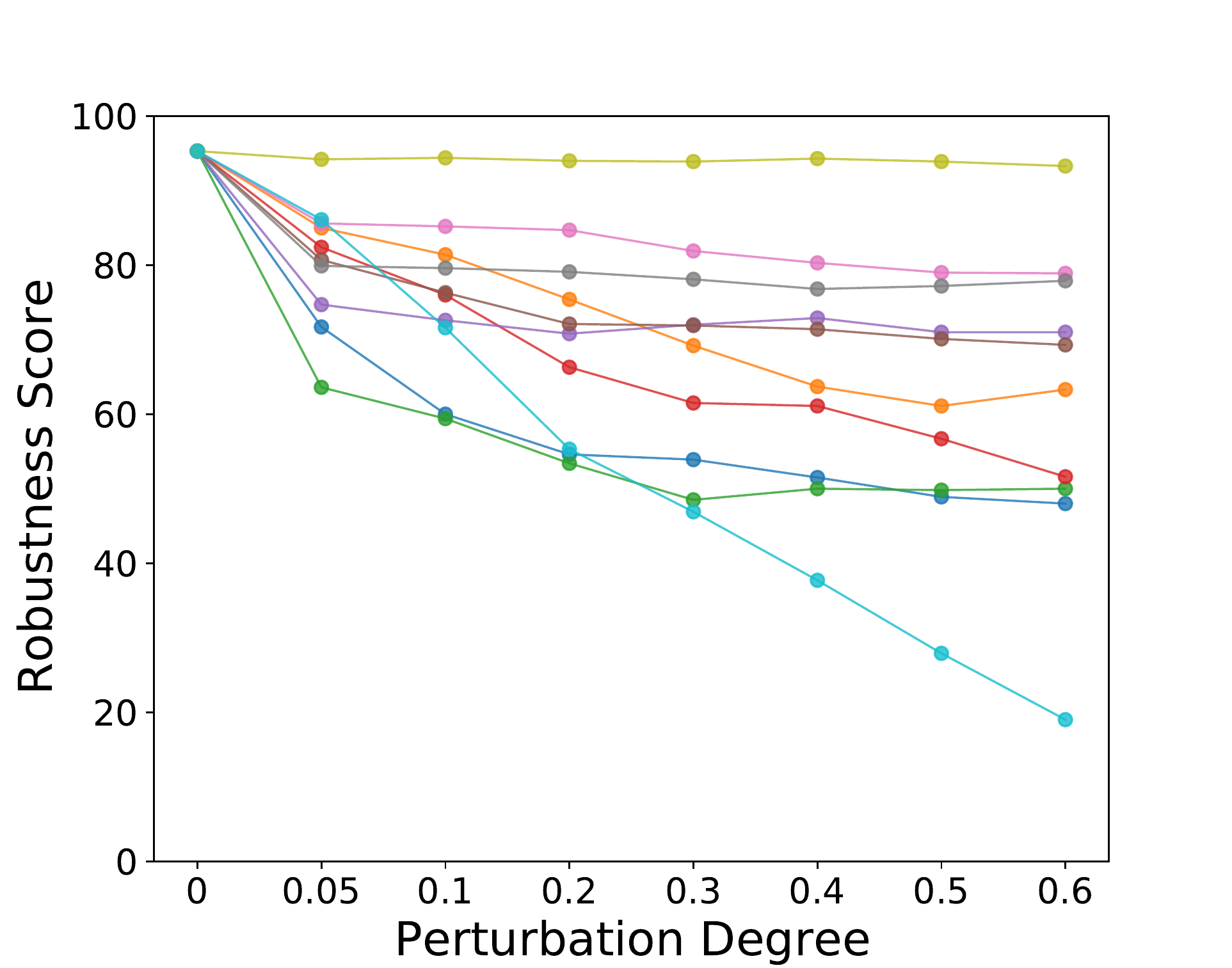}
\end{minipage}%
}%
\subfigure[Base-Score-Worst]{
\begin{minipage}[t]{0.24\linewidth}
\centering
\includegraphics[width=1.7in]{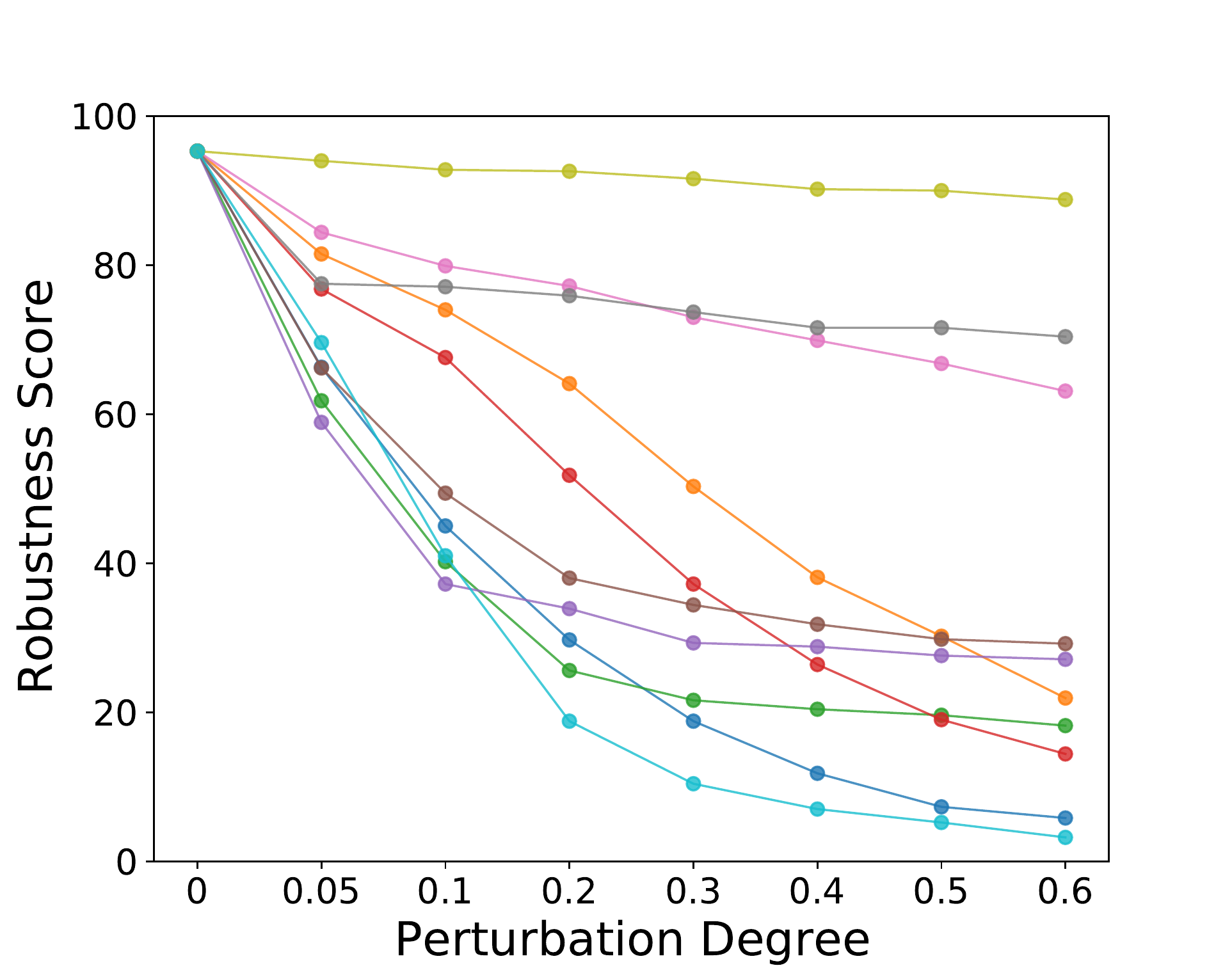}
\end{minipage}
}%
\subfigure[Large-Score-Average]{
\begin{minipage}[t]{0.24\linewidth}
\centering
\includegraphics[width=1.7in]{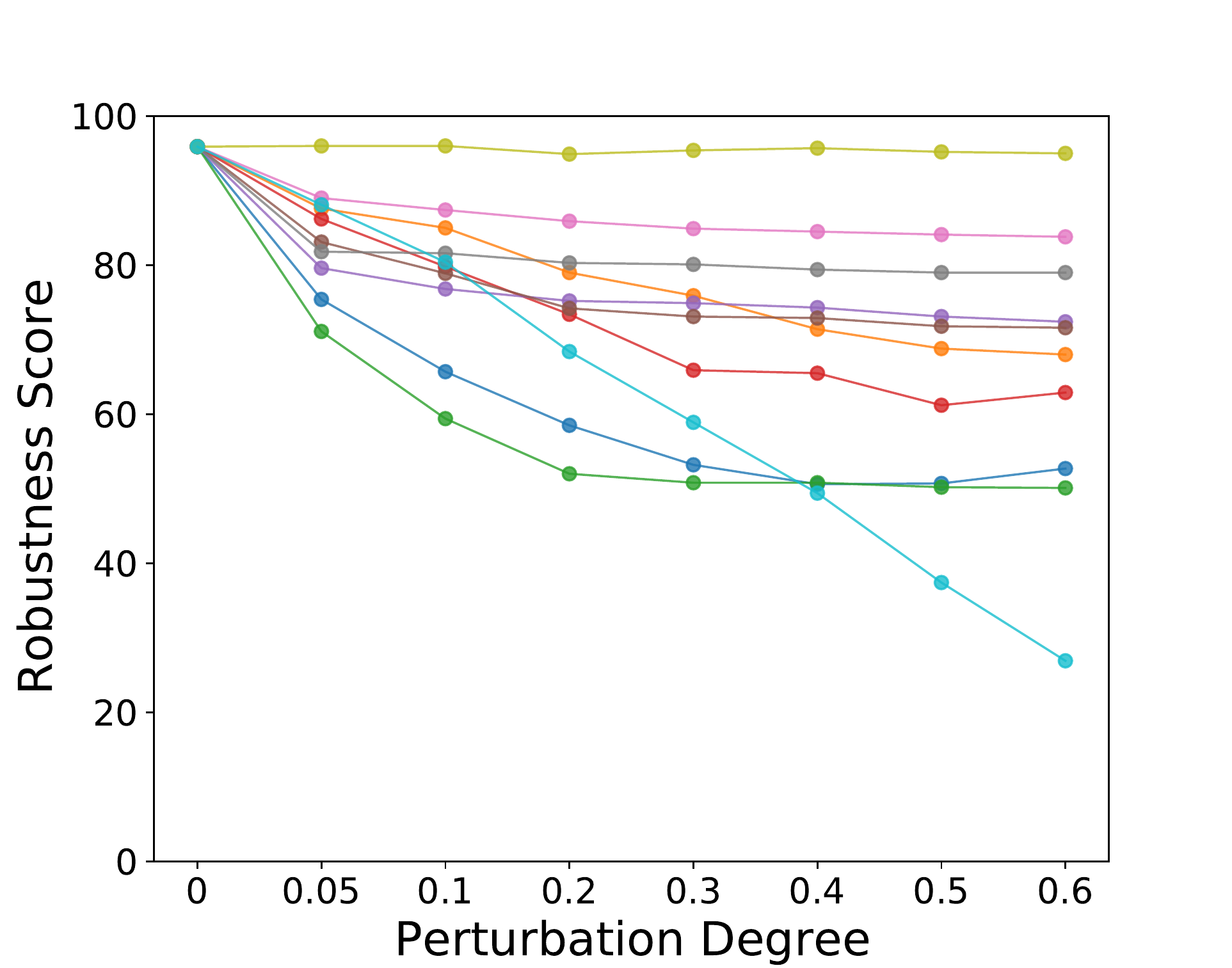}
\end{minipage}
}%
\subfigure[Large-Score-Worst]{
\begin{minipage}[t]{0.24\linewidth}
\centering
\includegraphics[width=1.7in]{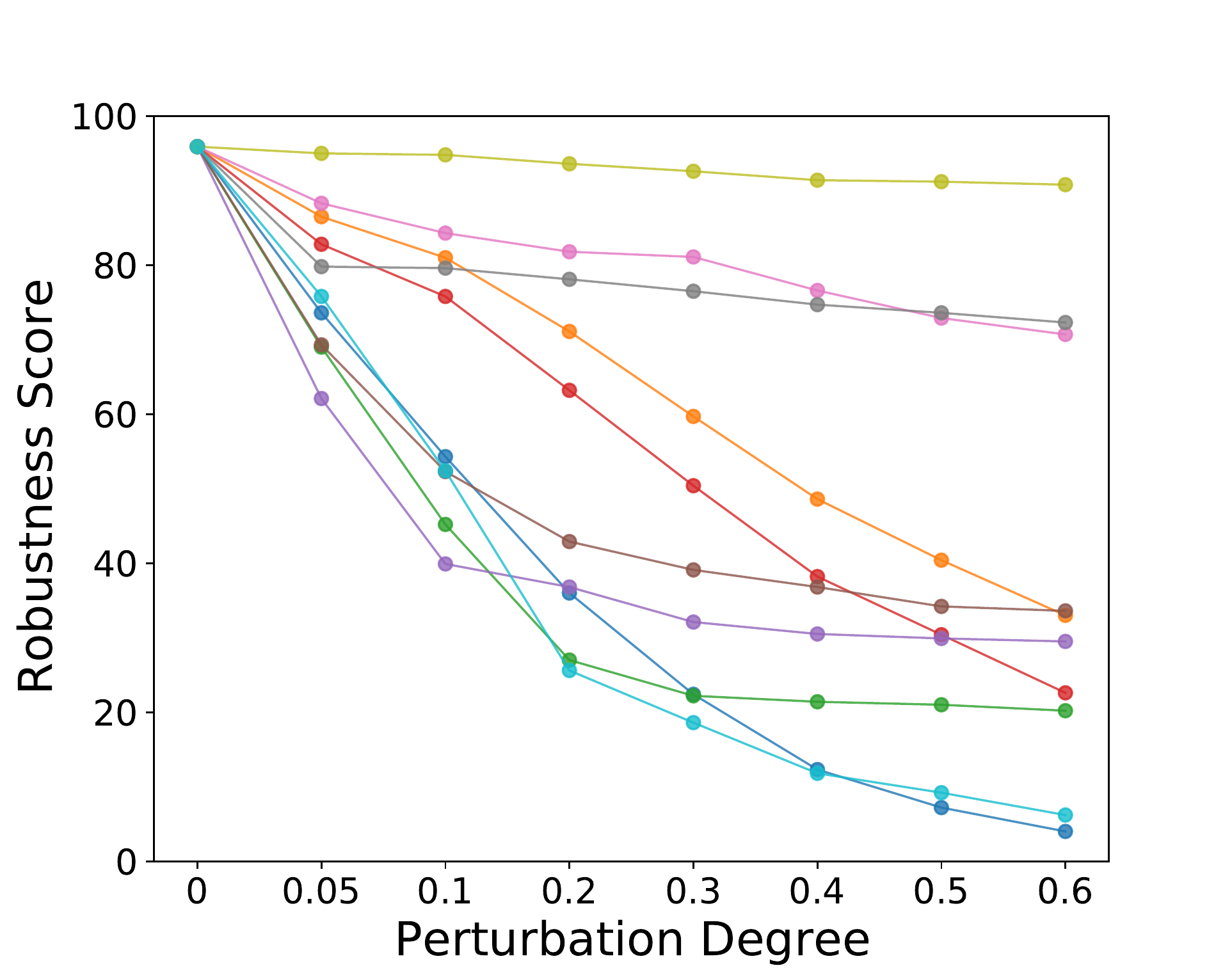}
\end{minipage}
}%
\\
\includegraphics[width=0.8\linewidth]{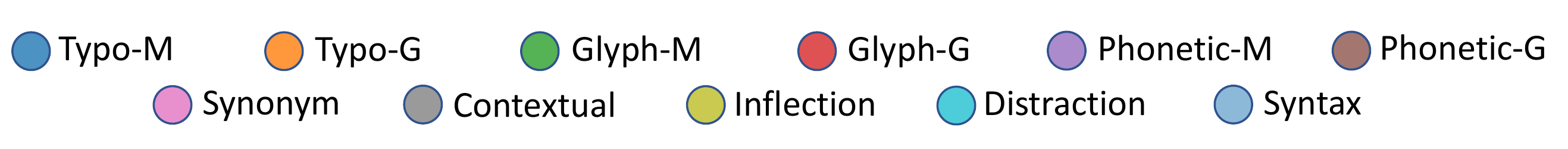}
\vspace{-10pt}
\centering
\caption{\label{fig:comprehensive_results} Comprehensive results of RoBERTa-base (Base) and RoBERTa-large (Large) on SST-2. We consider rule-based (Rule) and score-based (Score) attacks, and worst (Worst) and average (Average) performance estimation. }
\vspace{-10pt}
\end{figure*}

\subsection{Robustness Evaluation}
We consider two kinds of robustness evaluation: 
(1) Robustness evaluation of a given model; 
(2) Robustness comparison of two models. This can be easily extended to three or more models included. 

\looseness=-1
\paragraph{Single-model Robustness Evaluation.}
\looseness=-1
We generate robustness evaluation reports for given evaluated models. 
Figure~\ref{fig:report_example} shows an example of one single page of the robustness report of RoBERTa-base on SST-2, considering the Typo (Malicious) dimension.
Full reports for all datasets and models are in Appendix~\ref{sec:rob_report}. 
For each dimension, we show the robustness score computed at each robustness level considering two evaluation settings and two metrics, in both figures and the table.
We can observe that on average, the model can tolerate inputs with very small perturbation degrees (e.g., 0.05), but its performance degrades significantly in the worst performance estimation. 
This indicates that the model will be misled if malicious attackers try a little longer, even in small perturbation degrees. 
The final robustness scores for this dimension are derived by averaging over all robustness scores using Eq.~\ref{eq:l_c}, which will serve as overall estimations of the model's robustness in this dimension considering the validity criterion.
Also, we adopt the radar map to record the final robustness scores for all robustness dimensions, from which we can easily observe which dimension models fail. 
For example, we can observe from the radar map in Figure~\ref{fig:report_example} that RoBERTa-base fails frequently when users use various syntactic structures in their expressions or char-level transformations have been adopted for malicious attacks.
The implications are: (1) Practitioners should improve the model's capacity to capture syntax patterns or have extra mechanisms to deal with inputs with complex syntactic structures; 
(2) Practitioners should avoid deploying the model on security-related applications (e.g., hate-speech detection) to prevent hidden dangers.
%


\paragraph{Robustness Comparison.}

We can also generate reports to compare the two models' robustness.
Figure~\ref{fig:comparison} shows the core part of the report that compares the robustness of RoBERTa-base and RoBERTa-large considering all dimensions on SST-2.
We also employ radar maps to clearly show the robustness gap between the two models. 
The full report is in Appendix~\ref{sec:rob_compar_report} for demonstration.
We observe that RoBERTa-large consistently shows better robustness in all dimensions compared to RoBERTa-base.
This can be attributed to two potential factors: a) Larger models can generalize better beyond simple patterns (e.g., spurious correlations) in the in-distribution training dataset, thus more robust to distribution shifts~\cite{tu2020empirical}; b) Given the strong correlation between in-distribution and out-of-distribution performance~\cite{miller2021accuracy}, the robustness of larger models can be partially explained by better performance on in-distribution data. 
The quantification of these two factors is left for future work since the experiments in this paper are mainly for demonstration purposes.



\subsection{Analysis of Framework Components}
\label{sec:analysis_of_framework}
In this section, we analyze and prove the rationality of each component in our framework, including eight robustness dimensions, evaluation protocol, and our method to tackle the validity of adversarial samples.
For better demonstrations, we aggregate the results of eight dimensions considering two model sizes, two evaluation settings, and two metrics. 
The results on SST-2 are in Figure~\ref{fig:comprehensive_results}.
The results on AG's News and Jigsaw are in Appendix~\ref{sec:additional_results}.


\paragraph{Robustness Dimensions.}
We observe that models exhibit different capacities across all robustness dimensions, evidenced by substantially different robustness scores. 
This indicates the insufficiency in previous practice that adopts one single attack method to evaluate models' robustness. 
For example, only showing models' robustness to morphology inflection doesn't guarantee the same robustness transfer to inputs containing typos.
Thus, a multi-dimensional robustness evaluation in our framework is needed to reveal models' vulnerability in various circumstances, ensuring a more comprehensive evaluation of model capacities.
%
 
%

\paragraph{Evaluation Protocol.}
Our evaluation protocol includes two evaluation metrics (average and worst performance estimation) and two evaluation settings (rule-based and score-based). 
We show that the average performance estimation is in complementary to the worst performance estimation, showing the models' average success rates on the corresponding robustness dimension.
Thus, it can better reflect models' capacities since most attack methods can reduce models' worst performance estimation to near zero in high perturbation degrees, making it hard to compare different models.
%

Also, score-based and rule-based attacks consider different evaluation settings.
The score-based attacks are more effective than rule-based attacks considering average performance estimation. 
But the opposite is true considering worst performance estimation, probably because score-based attacks only perturb certain important words, limiting the search space.
Thus, incorporating these two evaluation settings is essential in robustness evaluation. 
%


\paragraph{Invalid Adversarial Samples Filtering.}
\label{sec:verify}

We observe that robustness scores drop along with the increase in the perturbation degrees across different models, datasets, and attack methods. 
However, as we argue, the robustness scores in higher perturbation degrees underestimate models' robustness since many successful but invalid adversarial samples exist. Thus, directly looking into the robustness curves without considering the influence of perturbation degrees on validity is unreliable. 

\looseness=-1
We justify our solution of incorporating the validity criterion into the robustness estimation process.
The basic intuition is that adversarial samples with higher perturbation degrees are more likely to become invalid. 
We conduct human annotation to verify it (See Table~\ref{tab:annotation}).
The annotation details are in Appendix~\ref{sec:annotate}.
We can observe that (1) attack methods have a large impact on sample validity, and (2) our intuition is justifiable since mostly a larger perturbation degree substantially harms the validity.

\begin{figure}[hpt!]
    \centering
    \includegraphics[width=0.70\linewidth]{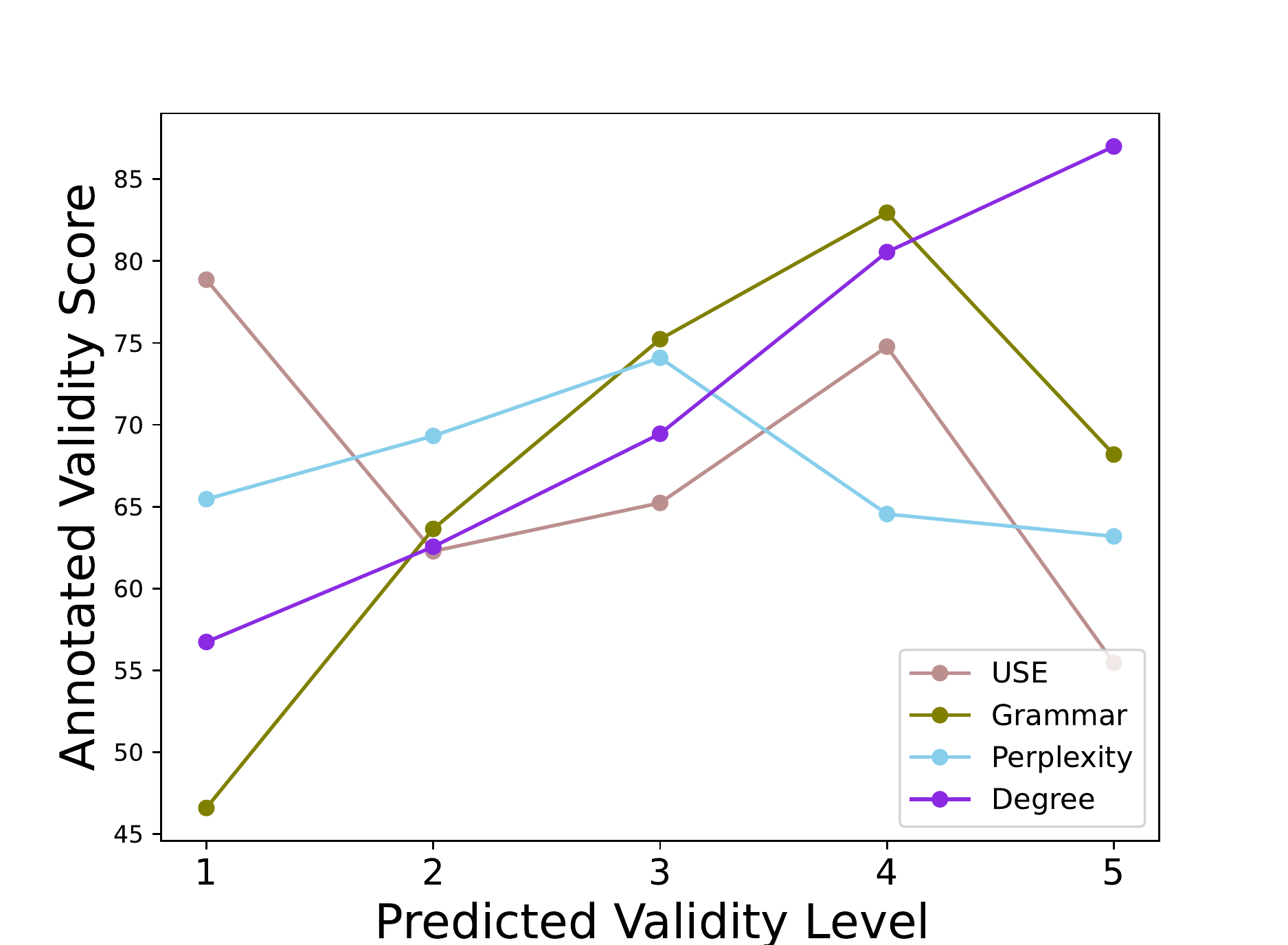}
       \caption{Results of the validity prediction. An ideal prediction should ensure the annotation validity score is proportional to the predicted validity level.}
    \label{fig:validity_eval}
    \vspace{-10pt}
\end{figure}

Also, we compare with previous heuristic filtering rules based on grammar errors (Grammar)~\citep{zang-etal-2020-word, chen2021multi}, sentence similarity (USE)~\citep{li-etal-2020-bert-attack, morris-etal-2020-reevaluating, wang-etal-2021-textflint, zeng-etal-2021-openattack}, and perplexity (Perplexity)~\citep{qi2021mind}.
We compute predictive validity scores for each adversarial sample based on the filtering rules (e.g., the perplexity rule will assign low validity scores to samples with high perplexity). 
For each filtering rule, we divide generated adversarial samples into five validity levels based on their validity scores and compute the average human annotated validity score of samples in five levels respectively (See Figure~\ref{fig:validity_eval}). 
Our method based on the perturbation degree better aligns with the ideal trend, while previous filtering methods show inconsistent trends and cannot effectively distinguish invalid cases.

%

\section{Related Work} 
Standard evaluation benchmarks~\cite{wang2018glue, wang2019superglue} follow the Independently Identical Distribution hypothesis that assumes the training and testing data come from the same distribution.
However, there is no such guarantee in practice, motivating the requirement to evaluate models' robustness beyond the standard accuracy. 
Various approaches have been proposed to simulate distribution shifts to construct static robustness evaluation benchmarks,
including stress test~\cite{naik-etal-2018-stress}, identifying and utilizing spurious correlations~\cite{mccoy2019right, zhang2019paws}, and domain shifts construction~\cite{hendrycks2020pretrained, yang2022glue}. 
Also, adversarial samples have been involved in robustness benchmarks, including machine-generated~\cite{wang2021adversarial} or human-in-the-loop generated~\cite{wallace2019trick, wallace2021analyzing, kiela2021dynabench} samples.

\looseness=-1
Compared to static benchmarks, we motivate to employ automatic attack methods to evaluate models' robustness dynamically, which is more comprehensive and rigorous.
Our work is built upon the long-lasting attack-and-defense arms race in adversarial NLP~\cite{wang2019towards, zhang2020adversarial}, mainly absorbing various attack methods. 
The attack methods can be roughly categorized into char-level, word-level, and sentence-level attacks, corresponding to the hierarchy in our framework.
Char-level attacks perturb the texts in the finest granularity, including deleting, inserting, replacing, swapping, and repeating characters \cite{karpukhin-etal-2019-training, gao2018blackbox}. 
Word-level attacks search for an optimal solution for word substitutions, using external knowledge bases \cite{ren-etal-2019-generating, zang-etal-2020-word} or contextual information \cite{li-etal-2020-bert-attack, garg2020bae, yuan2021bridge}. 
Sentence-level attacks transform the text considering syntactic patterns~\cite{iyyer-etal-2018-adversarial}, text styles~\cite{qi2021mind}, and domains \cite{wang-etal-2020-cat}.

\section{Conclusion}
We present a unified framework, providing solutions to three core challenges in automatic robustness evaluation.
%
We give a further discussion about robustness evaluation in Appendix~\ref{sec:discussion}.
In the future, we will selectively include more robustness dimensions in our framework.

%
%
%
%


\section*{Limitation}
Although we explore diverse robustness dimensions, there are more possible dimensions to cover, and we highly encourage future researchers to complete our paradigm for more comprehensive robustness evaluations. Moreover, our sample selection strategy is base on the perturbation degree. 
While being effective, this strategy is an approximate sub-optimal solution to the problem.
We leave finding better selection strategies as future work.

\section*{Ethical Consideration}
In this section, we discuss the intended use and energy saving considered in our paper.
\paragraph{Intended Use.} In this paper, we consider beyond the textual attack-and-defense arms race and highlight the role of adversarial attacks in robustness evaluation.
We design a systematic robustness evaluation paradigm to employ adversarial attacks for robustness evaluation.
We first summarize deficiencies in current works that limit the further use of adversarial attacks in practical scenarios.
Then we propose a standardized paradigm to evaluate the robustness of models using adversarial attacks.
We also develop an extensible toolkit to instantiate our paradigm.

\paragraph{Energy Saving. } 
We describe our experimental details to prevent other researchers from unnecessary hyper-parameter adjustments and to help them quickly reproduce our results.
We will also release all models we use in our experiments.


\section*{Acknowledgements}
This work is supported by the National Key R\&D Program of China (No. 2020AAA0106502), Major Project of the National SocialScience Foundation of China (No. 22\&ZD298), Institute Guo Qiang at Tsinghua University. 

Yangyi Chen and Ganqu Cui made the original research proposal and wrote the paper. 
Hongcheng Gao conducted experiments and helped to organize the paper. 
Lifan Yuan initiated the codebase and contributed to the proposal. 
Everyone else participated in the discussion, experiments, and paper writing of this study.



\bibliography{anthology,custom}
\bibliographystyle{acl_natbib}
\newpage

\newpage

\appendix
\section{Validity of Adversarial Samples}
\label{appendix:validity}

The original definition of adversarial samples in computer vision requires the perturbation to be imperceptible to human~\cite{goodfellow2014explaining}. However in NLP, texts are made of discrete tokens, where the changes are more apparent and difficult to measure. Therefore, the common practice in adversarial NLP recommend to evaluate the \textit{validity} of adversarial samples, which measures whether the transformed samples preserve the same meanings with the original samples, considering only the rationale part (a.k.a., the contents that determine the golden label). 
More precisely, valid adversarial samples preserve (1) the original labels and (2) the semantics of the rational part. 


\section{Justification of Perceptible Perturbations}
\label{sec:justify}
Consider the sample crafted by adding typos: ``I love the way that it took \color[HTML]{FE0000} chancs  \color[HTML]{000000} and really asks you to \color[HTML]{FE0000}takke \color[HTML]{000000}  these great leaps of faith and pays off.''
The common belief in adversarial NLP is to make the perturbations as small as possible.
So this sample with obvious perturbations highlighted in red will be dismissed in previous work.
But in our robustness evaluation framework, the requirement is to employ attack methods to simulate real-world inputs, which may contain some so-called perceptible perturbations like the above example.
Thus, we include various kinds of samples with perceptible perturbations in our framework provided that they can simulate real-world inputs well.

\begin{figure*}[htbp]
\centering

\subfigure[Base-Rule-Average]{
\begin{minipage}[t]{0.24\linewidth}
\centering
\includegraphics[width=1.7in]{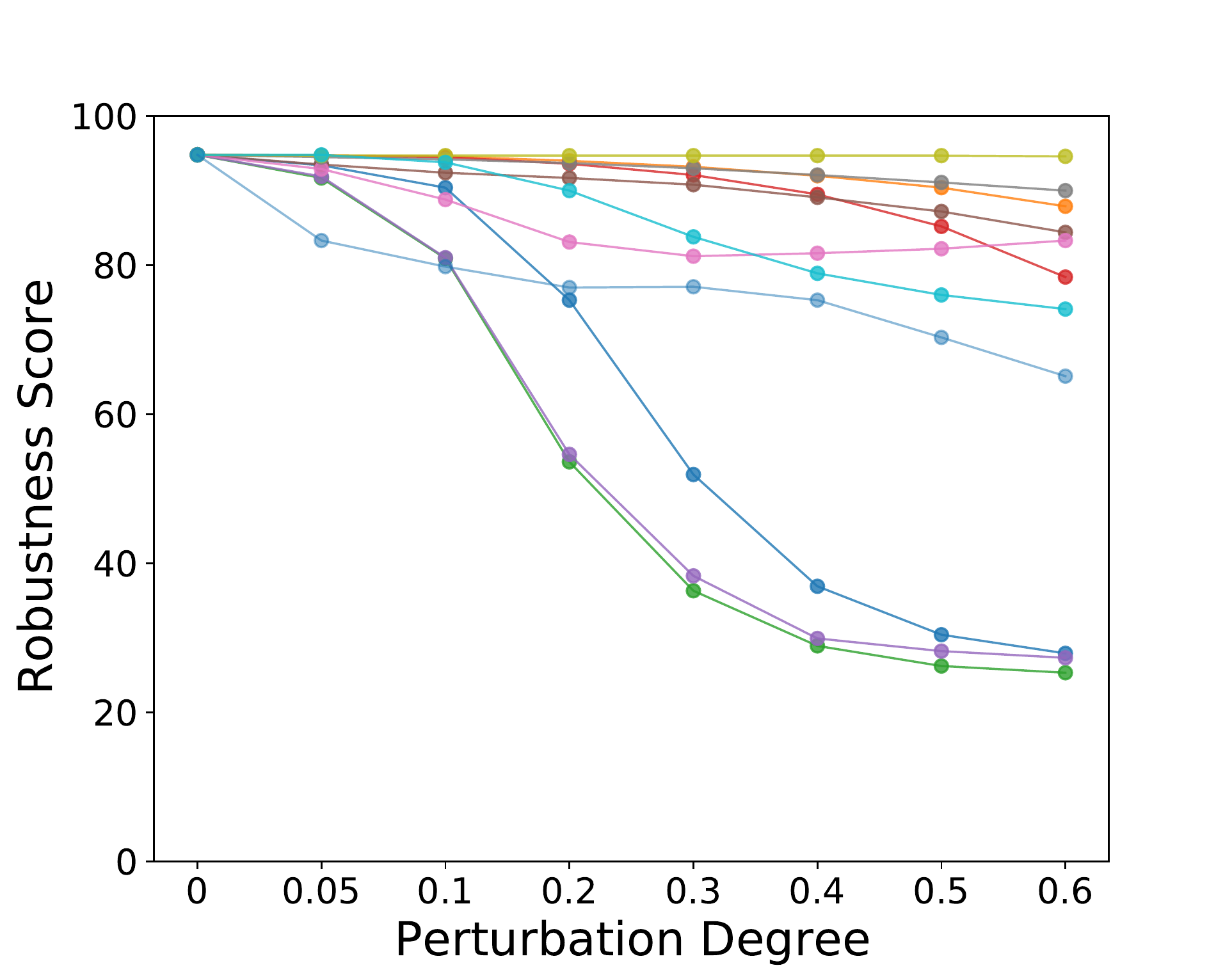}
\end{minipage}%
}%
\subfigure[Base-Rule-Worst]{
\begin{minipage}[t]{0.24\linewidth}
\centering
\includegraphics[width=1.7in]{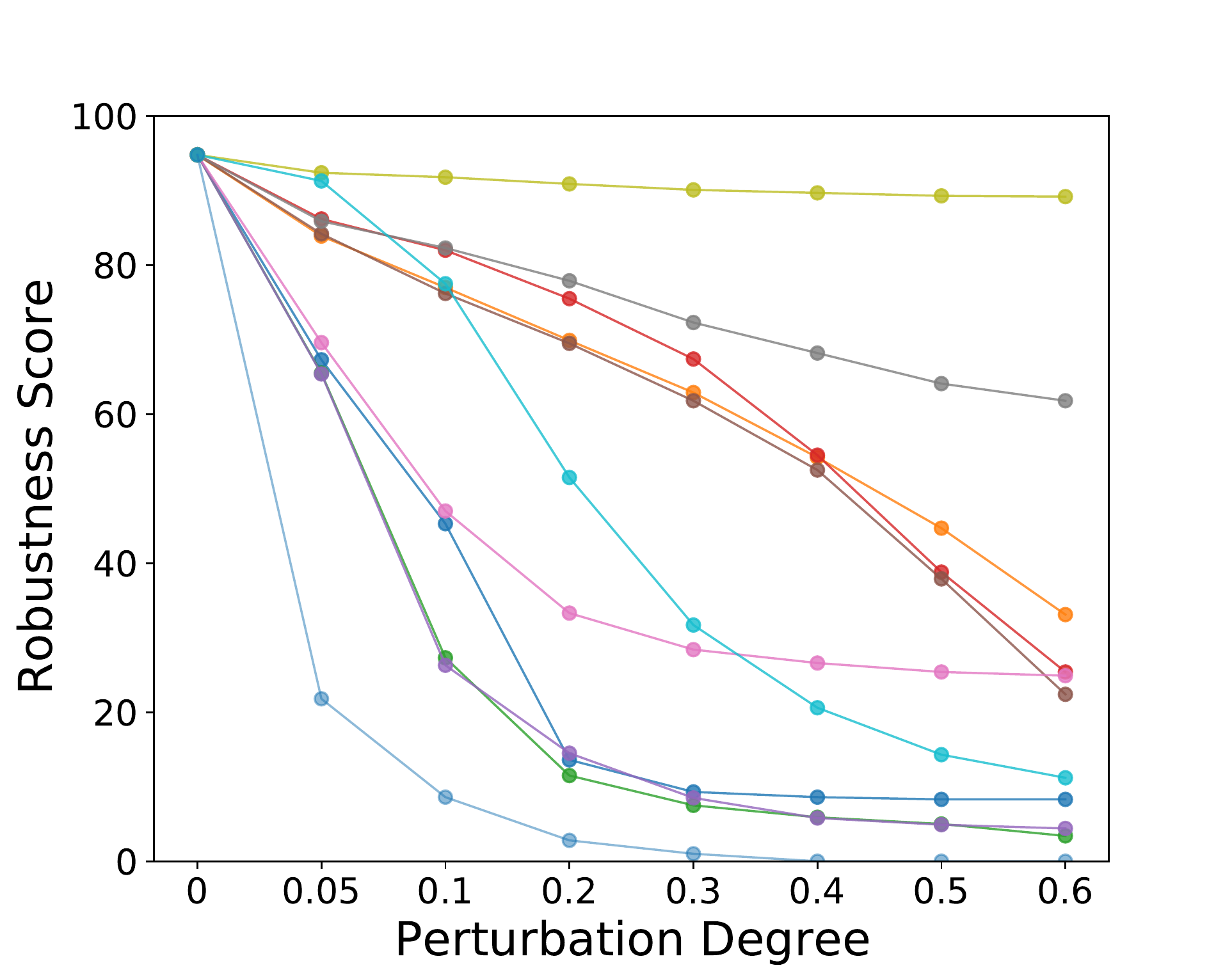}
\end{minipage}%
}%
\subfigure[Large-Rule-Average]{
\begin{minipage}[t]{0.24\linewidth}
\centering
\includegraphics[width=1.7in]{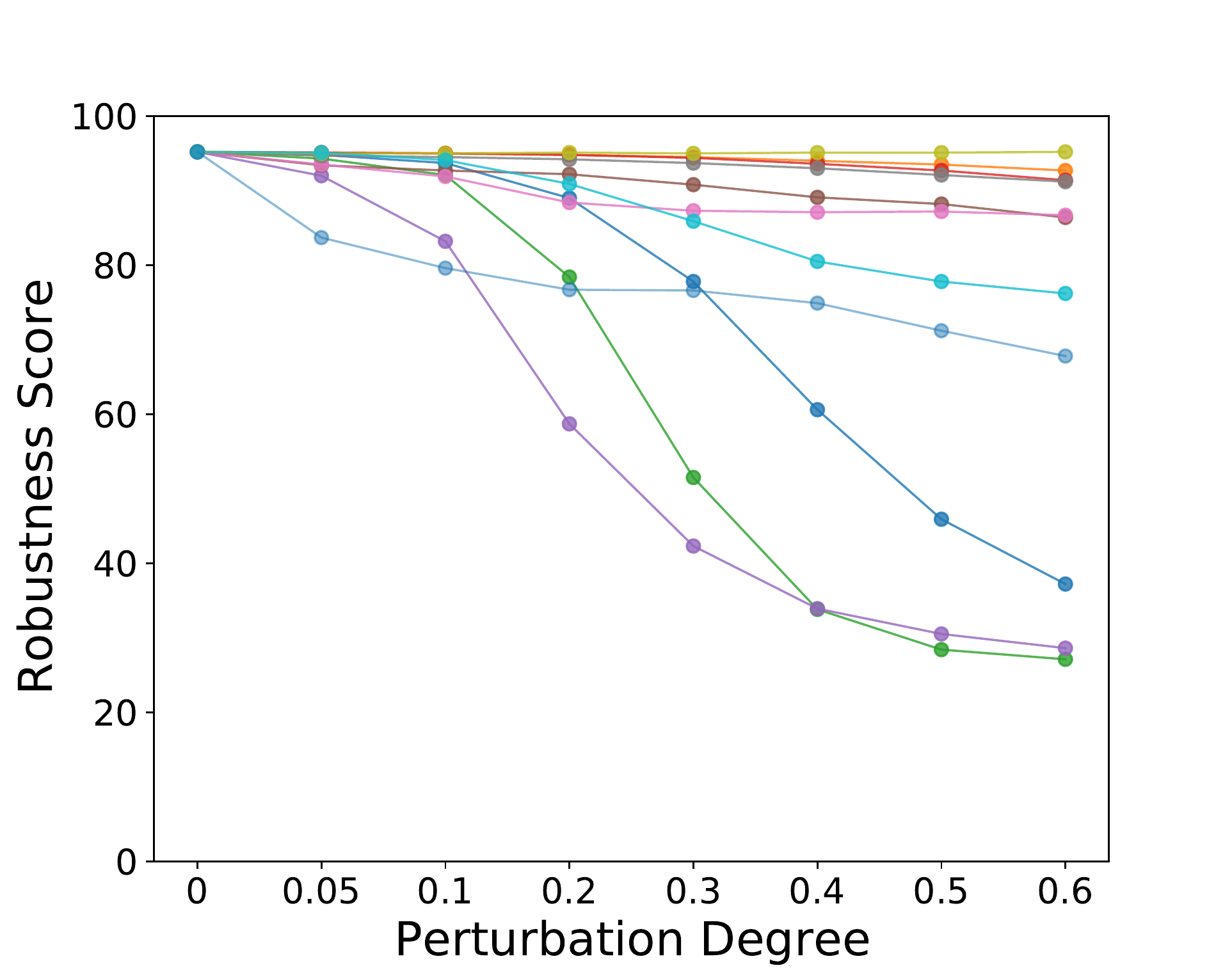}
\end{minipage}%
}%
\subfigure[Large-Rule-Worst]{
\begin{minipage}[t]{0.24\linewidth}
\centering
\includegraphics[width=1.7in]{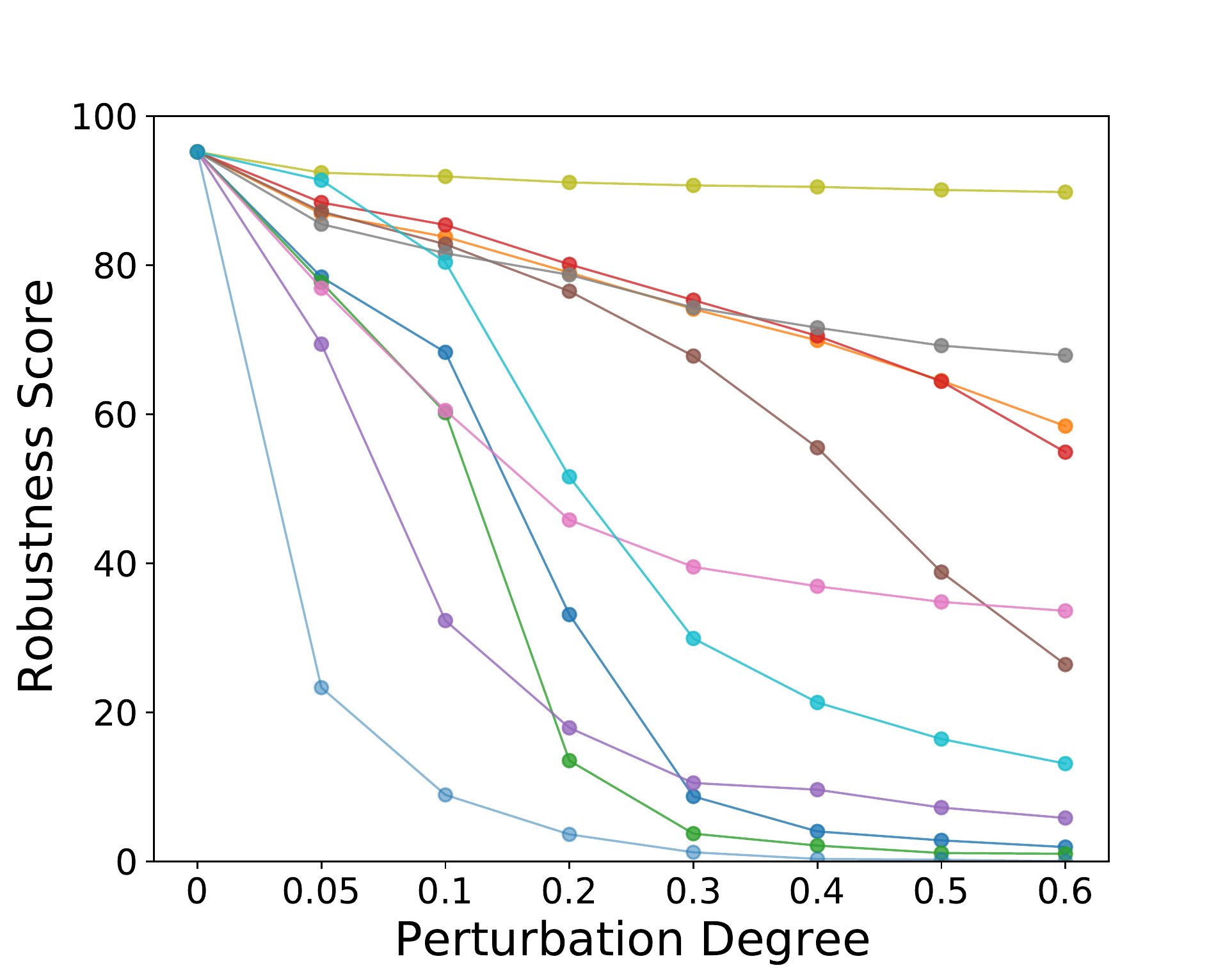}
\end{minipage}%
}%
    \quad             
\subfigure[Base-Score-Average]{
\begin{minipage}[t]{0.24\linewidth}
\centering
\includegraphics[width=1.7in]{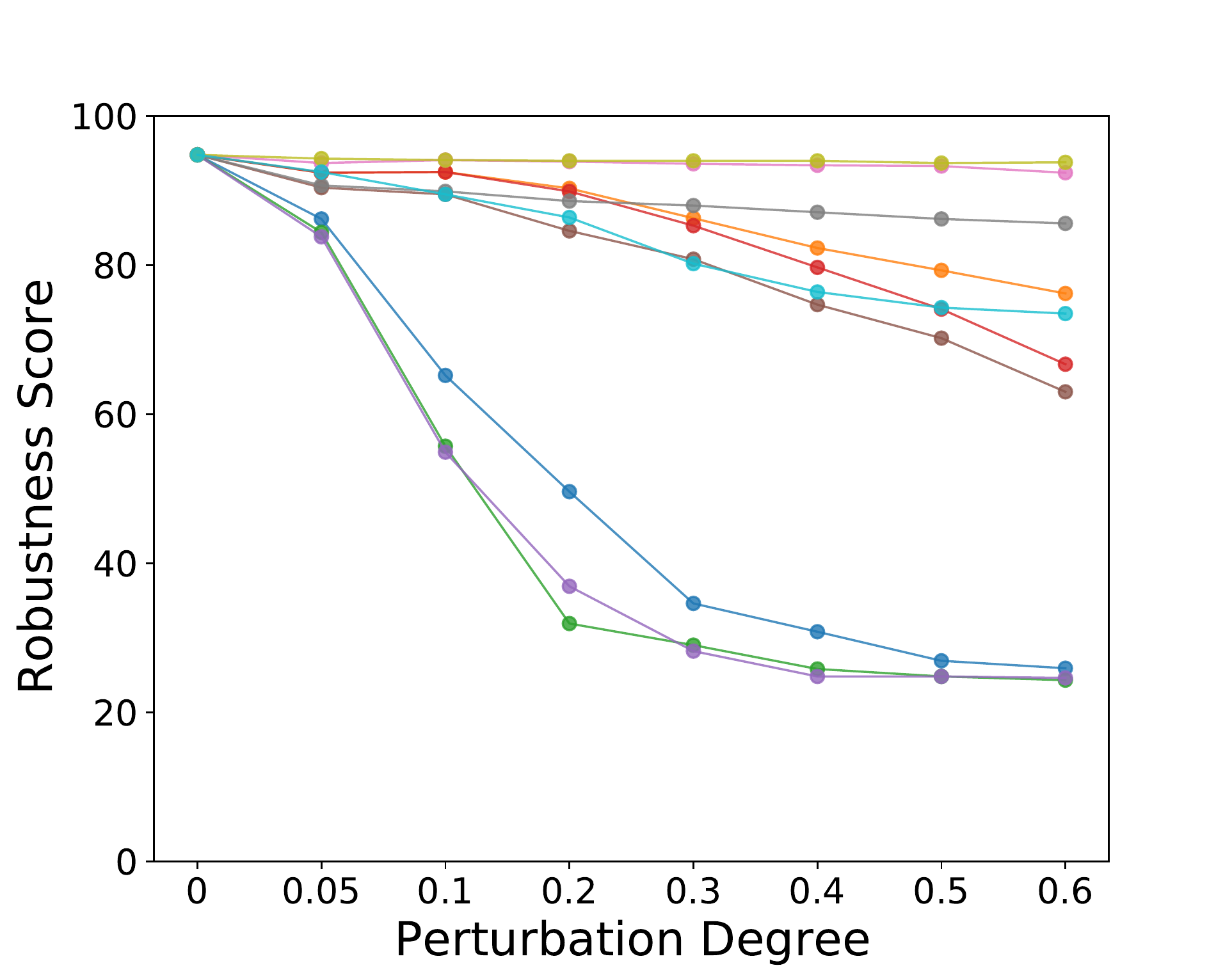}
\end{minipage}%
}%
\subfigure[Base-Score-Worst]{
\begin{minipage}[t]{0.24\linewidth}
\centering
\includegraphics[width=1.7in]{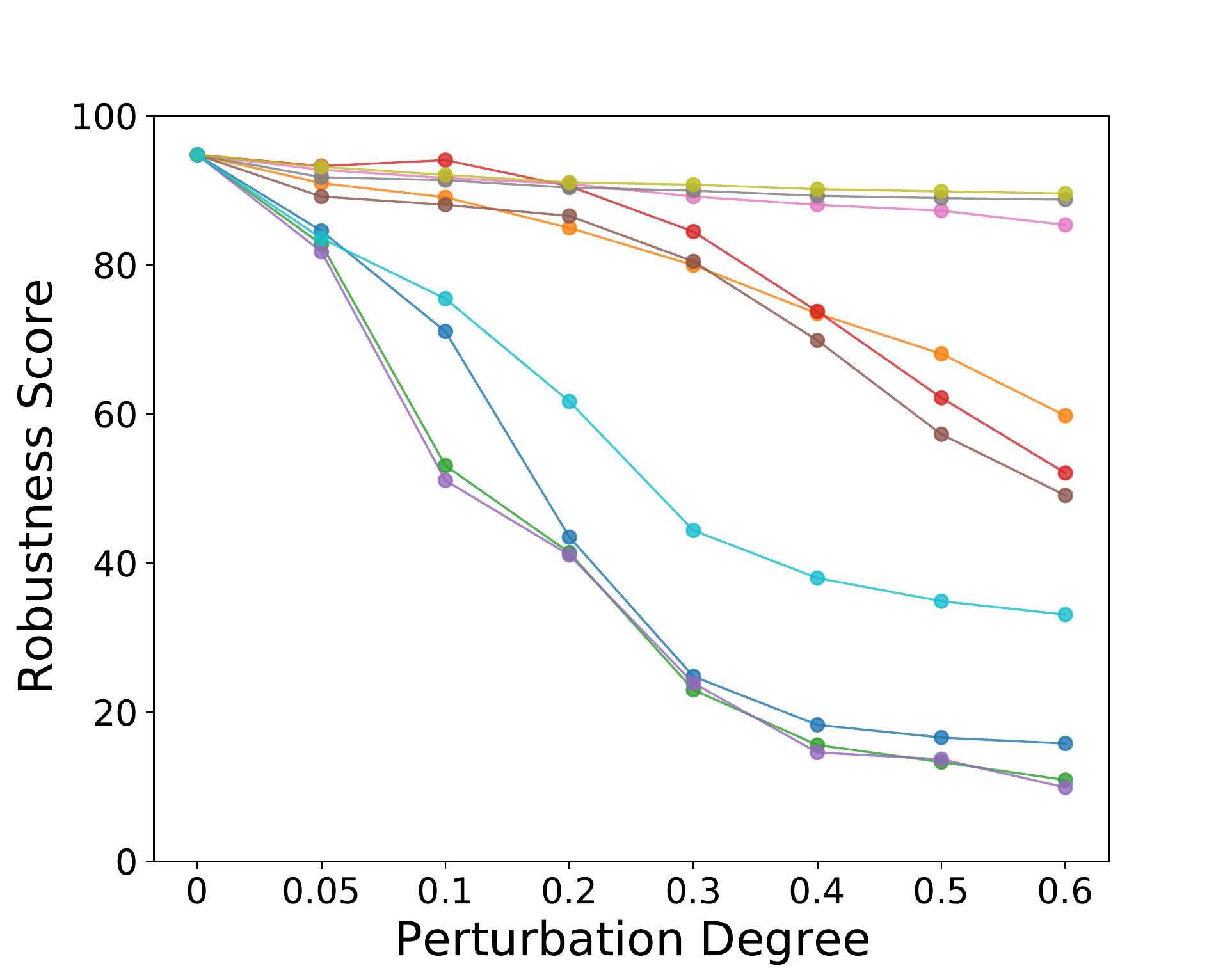}
\end{minipage}
}%
\subfigure[Large-Score-Average]{
\begin{minipage}[t]{0.24\linewidth}
\centering
\includegraphics[width=1.7in]{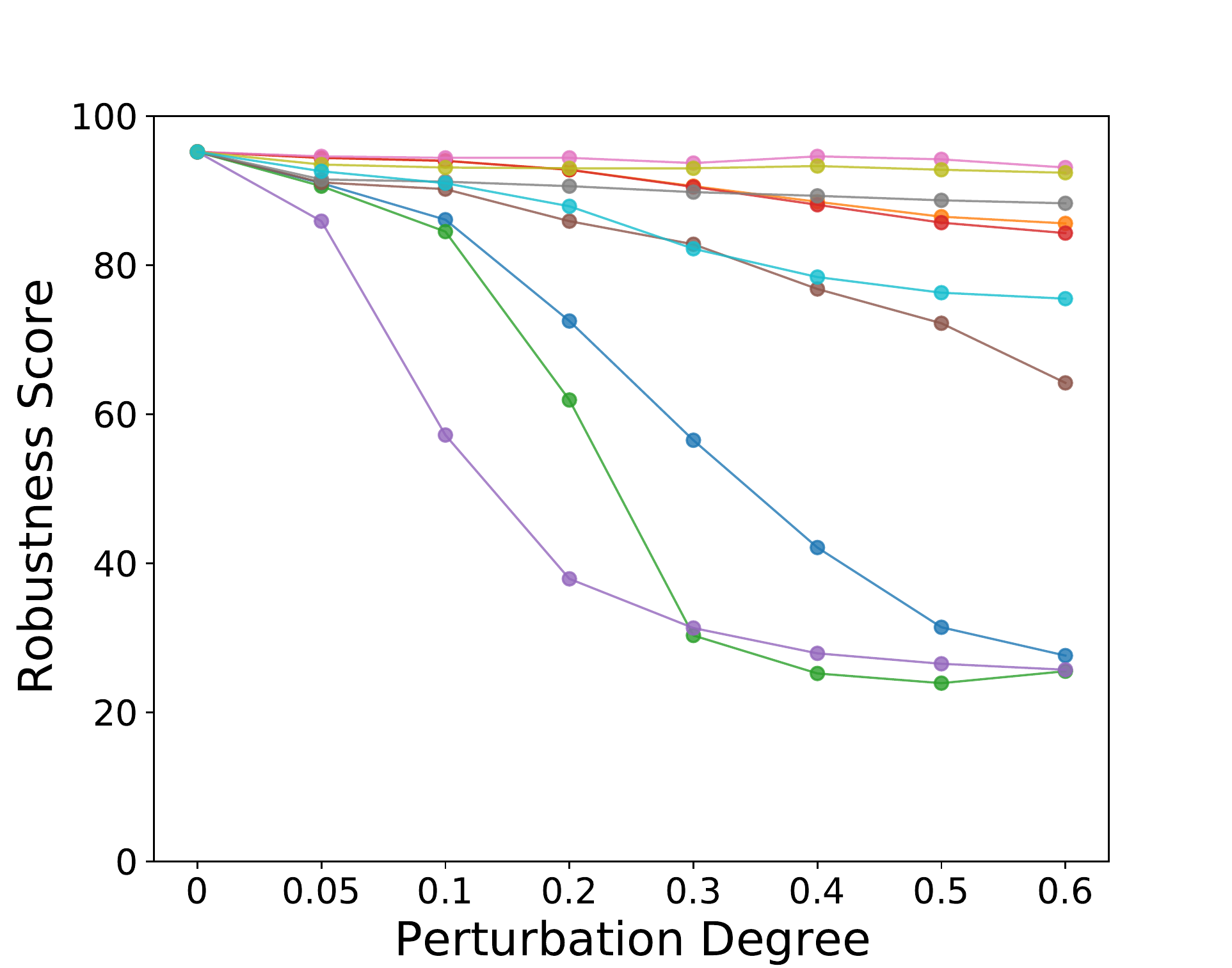}
\end{minipage}
}%
\subfigure[Large-Score-Worst]{
\begin{minipage}[t]{0.24\linewidth}
\centering
\includegraphics[width=1.7in]{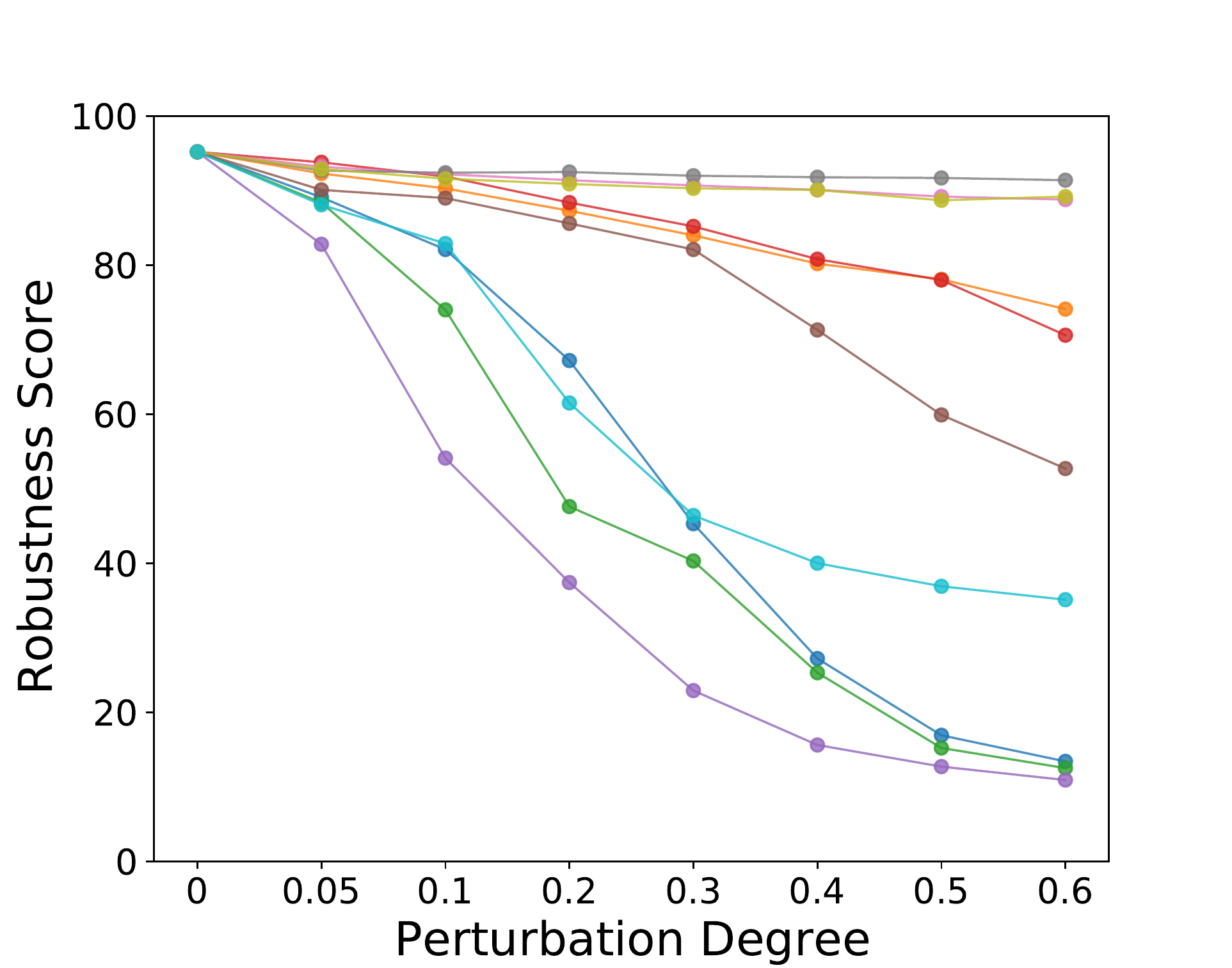}
\end{minipage}
}%
\\
\includegraphics[width=0.95\linewidth]{figs/Presentation1.pdf}

\centering
\caption{\label{fig:comprehensive_results_ag} Comprehensive results of RoBERTa-base (Base) and RoBERTa-large (Large) on AG's News. We consider rule-based (Rule) and score-based (Score) attacks, and worst (Worst) and average (Average) performance estimation. }
\end{figure*}

\section{Computation of Perturbation Degree}
\label{appendix:compute_degree}
\looseness=-1
For three transformation levels, we employ different computational methods to measure the perturbation degree.
For char-level transformations with the malicious tag, we adopt the relative Levenshtein Distance.
For char-level transformations with the general tag, we restrict the algorithms to perturb less than two characters for each word to better simulate inputs from benign users and adopt the word modification rate to measrue the perturbation degree.
For word-level transformations, we employ the word modification rate. 
For sentence-level transformations, we employ embedding similarity.
Next, we introduce how to compute these measurements.


%

\paragraph{Relative Edit Distance.} 
We use relative edit distance to measure the perturbation degree of char-level attacks with the malicious tag. Assume that the original text has $N_c$ characters in total. We modify $n_c$ characters in original text $X$ and get a new text $X'$. Then the Edit Distance between $X$ and $X'$ is $n_c$, and the perturbation degree is:
\[{D_c} = \frac{{{n_c}}}{{{N_c}}}.\]





\paragraph{Word Modification Rate.}
We use word modification rate to measure the perturbation degree of char-level attacks with the general tag and word-level attacks. Assume that the original text has $N_w$ words in total, and we perturb $n_w$ words. Then the perturbation degree is:
\[{D_w} = \frac{{{n_w}}}{{{N_w}}}.\]
Specifically for char-level attack, we only conduct one char-level modification for each perturbed word. 

\paragraph{Embedding Similarity.}
We adopt embedding similarity to measure the perturbation degree of sentence-level attack. We get the sentence embeddings with Sentence-Transformers~\cite{reimers-2019-sentence-bert}. Denote the sentence embedding of original sentence $\boldsymbol{x}$ , the transformed sentence embedding as $\boldsymbol{x'}$, and the embedding similarity between $\boldsymbol{x}$ and $\boldsymbol{x'}$ is calculated by cosine function $\cos\left( {\boldsymbol{x},\boldsymbol{x'}} \right)$. We compute the cosine similarity between two embeddings. Then the degree is:
\[D_s = 1 - \cos\left( {\boldsymbol{x},\boldsymbol{x'}} \right).\]


\begin{table*}[t]
\centering
\resizebox{\textwidth}{!}{
\begin{tabular}{l|ccccccccccc}
\toprule
Degree & \multicolumn{1}{l}{Typo-M} & \multicolumn{1}{l}{Glyph-M} & \multicolumn{1}{l}{Phonetic-M} & \multicolumn{1}{l}{Typo-G} & \multicolumn{1}{l}{Glyph-G} & \multicolumn{1}{l}{Phonetic-G} & \multicolumn{1}{l}{Synonym} & \multicolumn{1}{l}{Contextual} & \multicolumn{1}{l}{Inflection} & \multicolumn{1}{l}{Syntax} & \multicolumn{1}{l}{Distraction} \\ \midrule
0.05   & 0.96                                 & 1                                     & 1                                        & 1                                  & 1                                   & 1                                      & 0.44                        & 0.46                           & 1                              & -                          & 0.98                            \\
0.1    & 0.94                                 & 0.98                                  & 1                                        & 1                                  & 1                                   & 1                                      & 0.32                        & 0.44                           & 1                              & 0.28                       & 0.94                            \\
0.3    & 0.26                                 & 0.94                                  & 1                                        & 1                                  & 1                                   & 1                                      & 0.20                        & 0.32                           & 1                              & 0.06                       & 0.94                            \\
0.5    & 0.06                                 & 0.86                                  & 1                                        & 0.82                               & 1                                   & 1                                      & 0.14                        & 0.20                           & 0.98                           & 0.02                       & 0.82                            \\
0.8    & 0.02                                 & 0.70                                  & 0.98                                     & 0.64                               & 1                                   & 0.98                                   & 0.14                        & 0.06                           & 0.98                           & 0                          & 0.64                            \\ \bottomrule
\end{tabular}
}
\caption{\label{tab:annotation} Human annotation of samples validity considering five perturbation degrees and all attack methods.
 }
\end{table*}

\section{Human Annotation}
\label{sec:annotate}
\subsection{Annotation Details}
We conduct human annotation to evaluate the validity of adversarial samples generated by different methods at different perturbation degrees. 
We employ 3 human annotators, and use the voting strategy to produce the annotation results.
For each method and perturbation degree, we sample 50 successful adversarial samples. 
The final score is averaged over all 50 adversarial samples. 
Specifically for the annotation, we show annotators the original sample, the perturbed sample, and the original label, and ask annotators to give a binary score. 
1 represents (1) the original label is the same in the perturbed sample, and (2) the semantic preservation of the rationale part is good. 
0 indicates that either rule is not satisfied, or the perturbed sample is hard to comprehend. 
Note that we don't let the annotators to predict the labels of the perturbed samples and check the label consistency since validity is a higher-standard task that requires semantics invariance.

In the annotation process, we first write an annotation document containing some cases and instructions for annotators. 
Then we compose some cases to test the annotators. 
Only qualified annotators are involved in the final annotation task.

\subsection{Annotation Results}
The human annotation results to verify the intuition that adversarial samples with higher perturbation degrees are more likely to become invalid are listed in Table~\ref{tab:annotation}.
Additionally, it is pertinent to mention that our evaluation methodology for assessing validity can also be applied to textual backdoor learning, which faces the same evaluation challenge~\citep{cui2022unified}.


\section{Additional Result}
\label{sec:additional_results}
\begin{figure*}[htbp]
\centering

\subfigure[Base-Rule-Average]{
\begin{minipage}[t]{0.24\linewidth}
\centering
\includegraphics[width=1.7in]{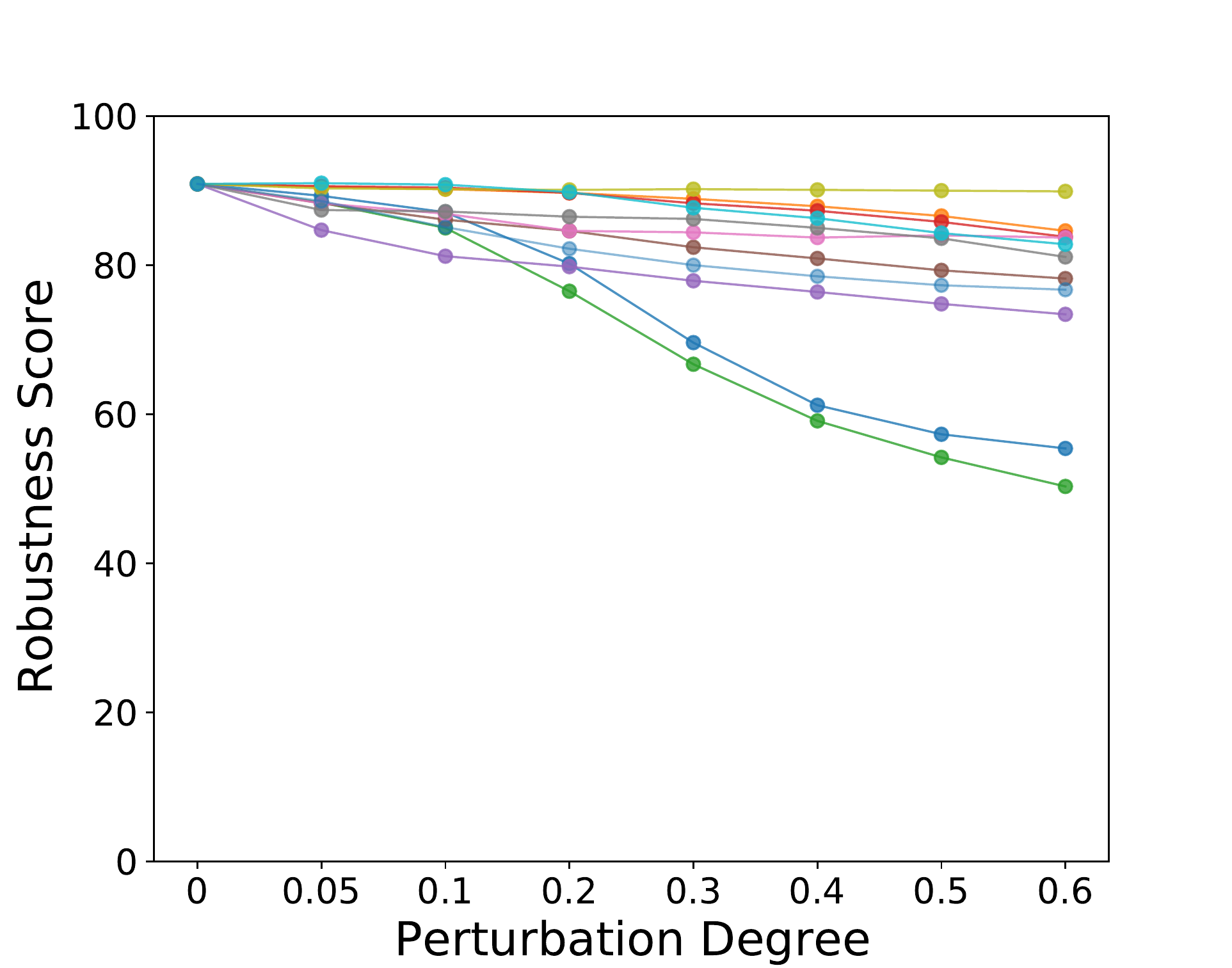}
\end{minipage}%
}%
\subfigure[Base-Rule-Worst]{
\begin{minipage}[t]{0.24\linewidth}
\centering
\includegraphics[width=1.7in]{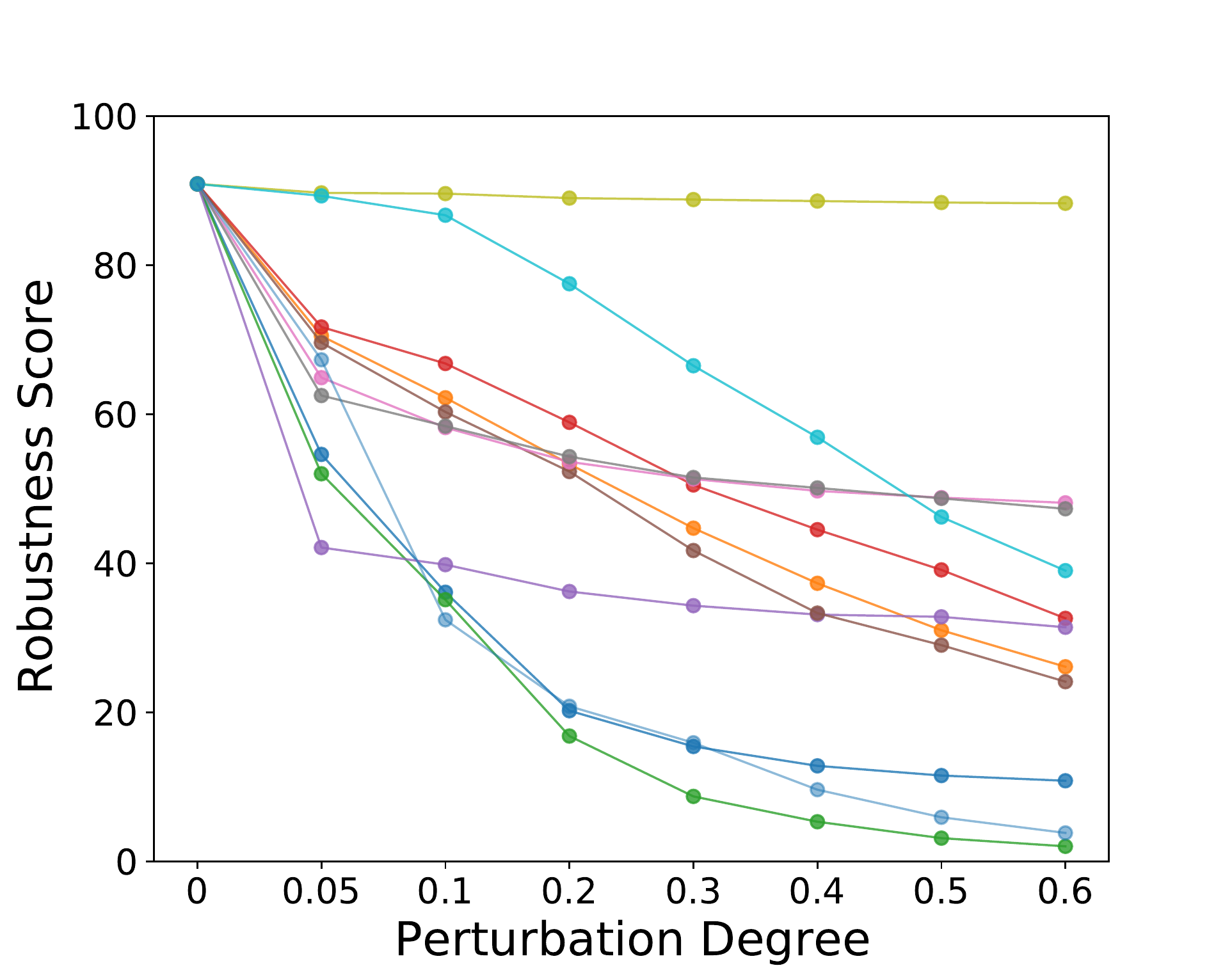}
\end{minipage}%
}%
\subfigure[Large-Rule-Average]{
\begin{minipage}[t]{0.24\linewidth}
\centering
\includegraphics[width=1.7in]{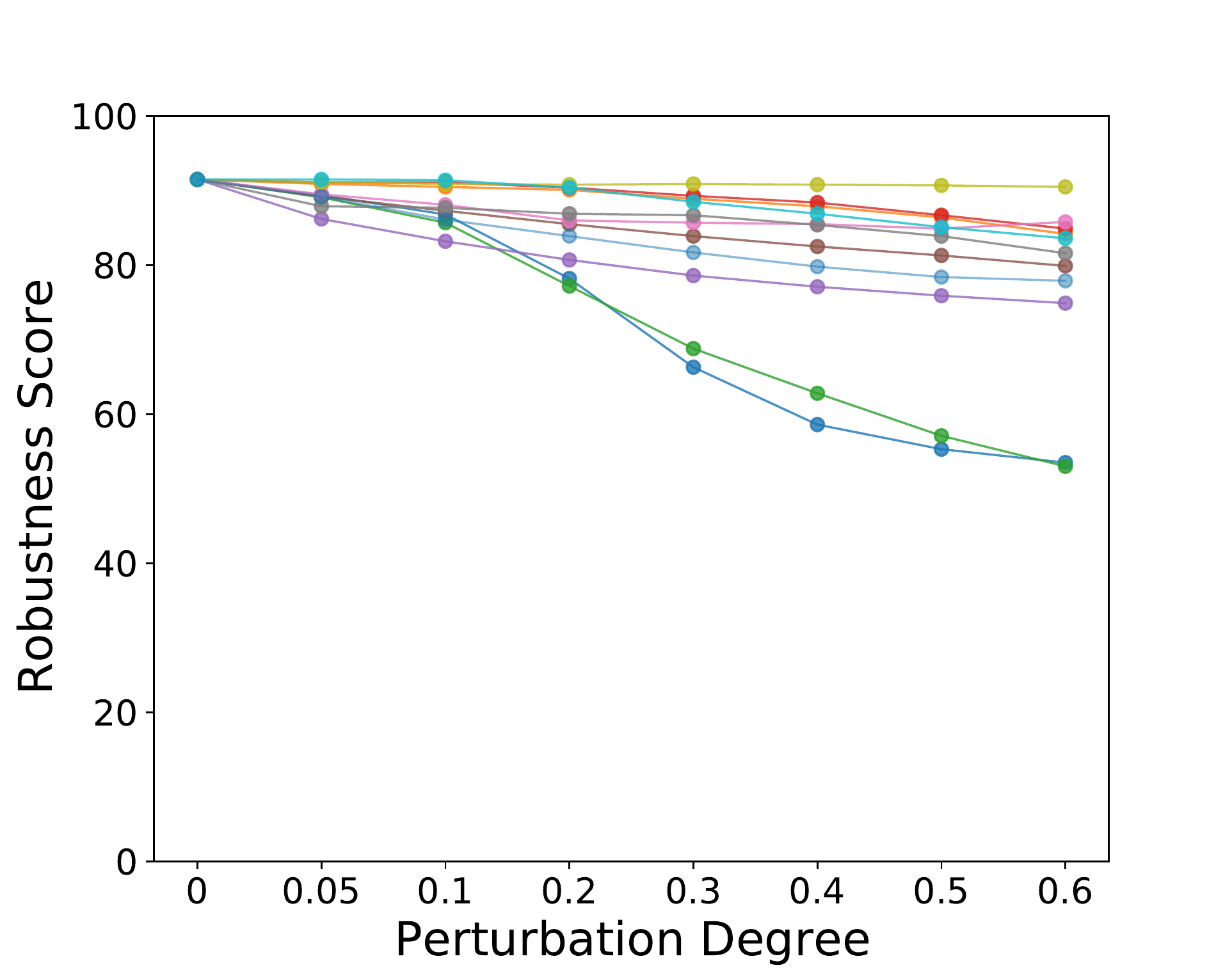}
\end{minipage}%
}%
\subfigure[Large-Rule-Worst]{
\begin{minipage}[t]{0.24\linewidth}
\centering
\includegraphics[width=1.7in]{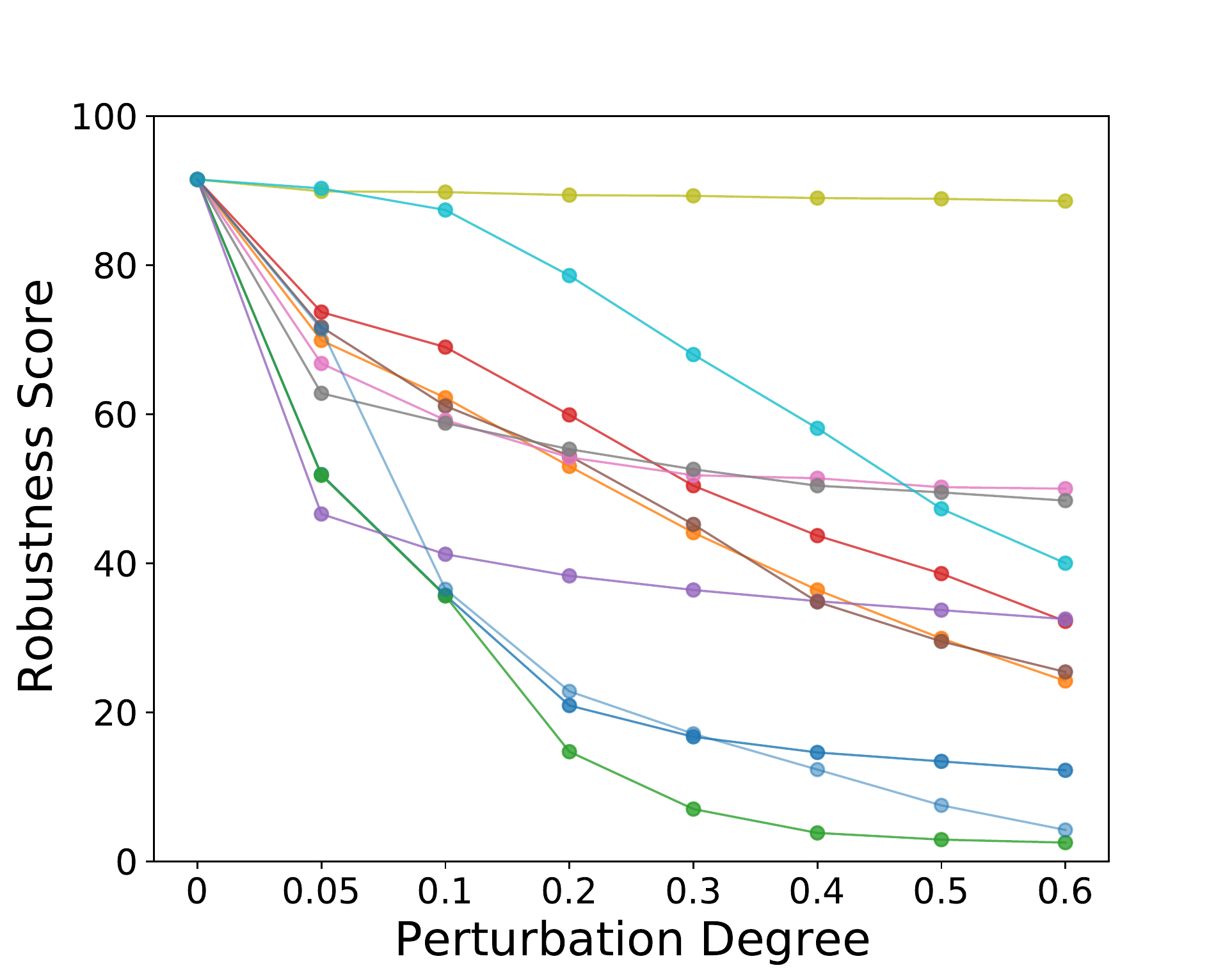}
\end{minipage}%
}%
    \quad             
\subfigure[Base-Score-Average]{
\begin{minipage}[t]{0.24\linewidth}
\centering
\includegraphics[width=1.7in]{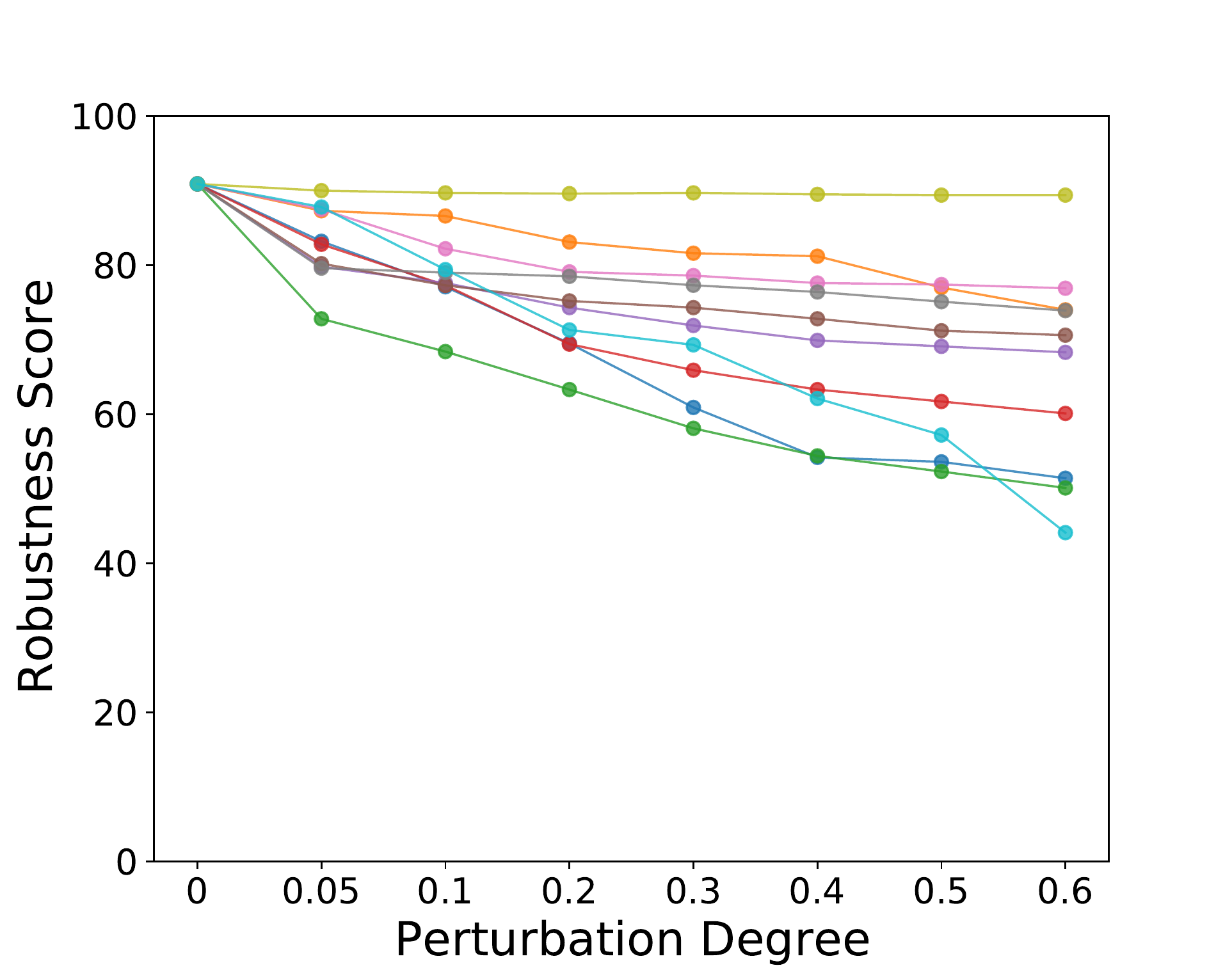}
\end{minipage}%
}%
\subfigure[Base-Score-Worst]{
\begin{minipage}[t]{0.24\linewidth}
\centering
\includegraphics[width=1.7in]{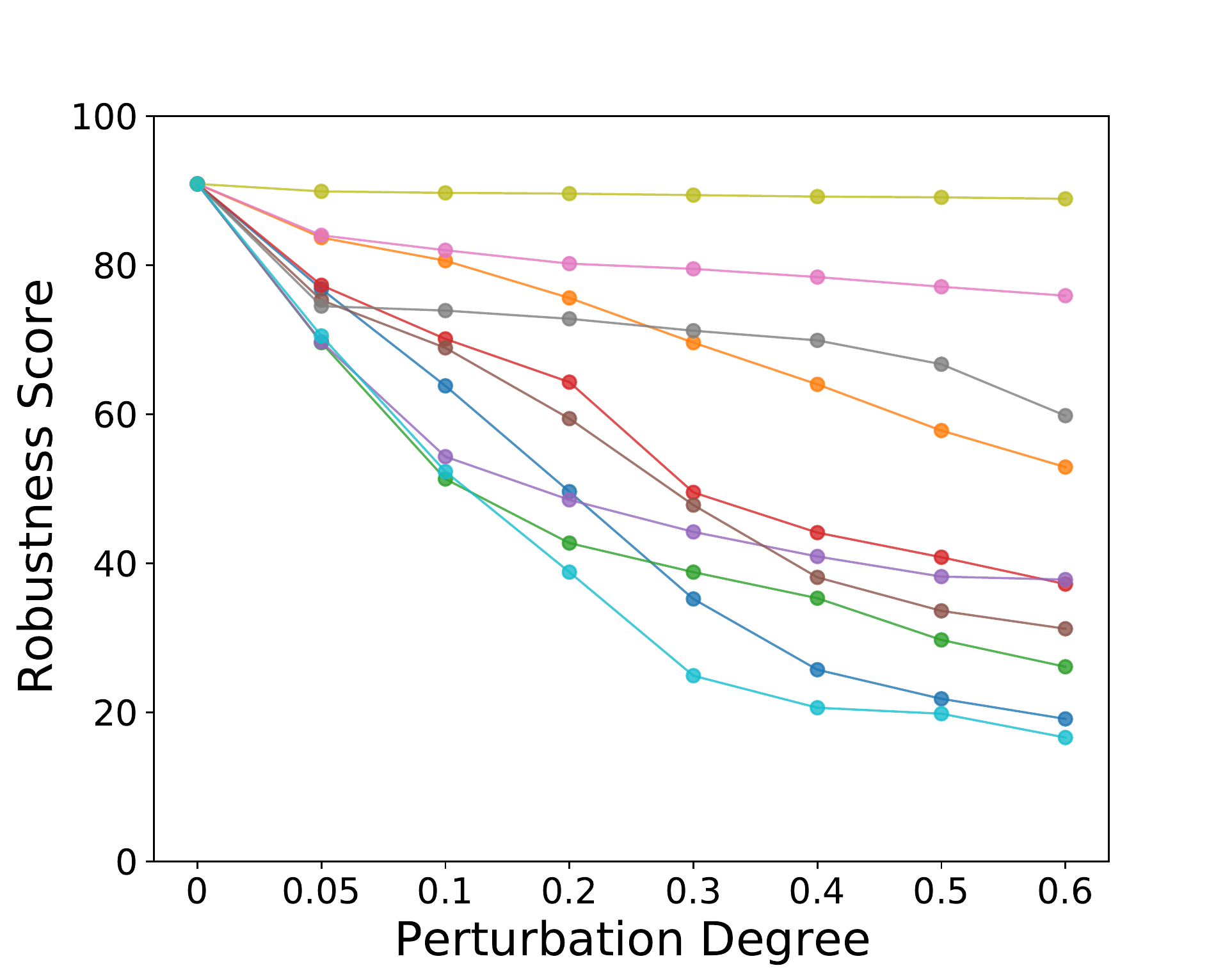}
\end{minipage}
}%
\subfigure[Large-Score-Average]{
\begin{minipage}[t]{0.24\linewidth}
\centering
\includegraphics[width=1.7in]{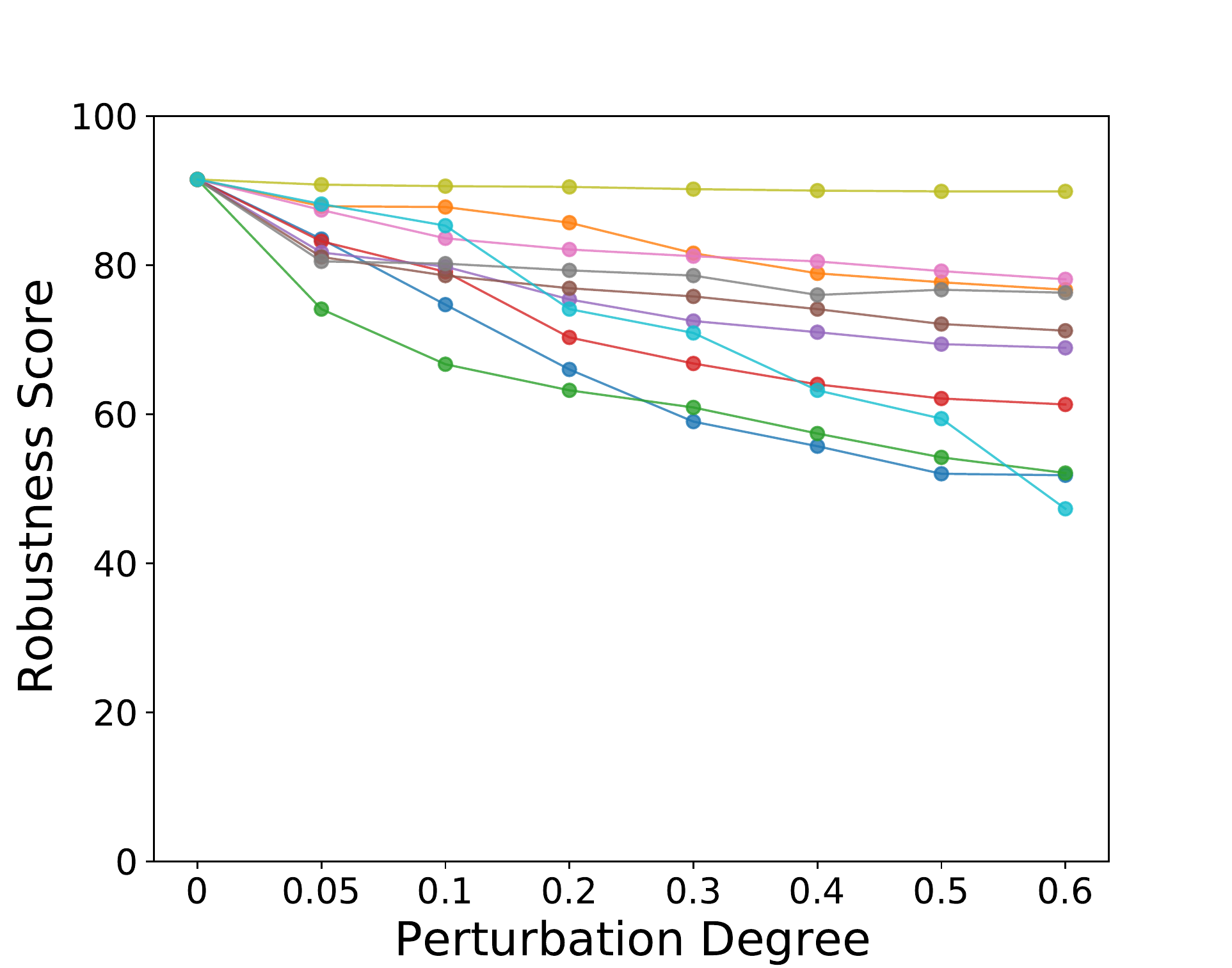}
\end{minipage}
}%
\subfigure[Large-Score-Worst]{
\begin{minipage}[t]{0.24\linewidth}
\centering
\includegraphics[width=1.7in]{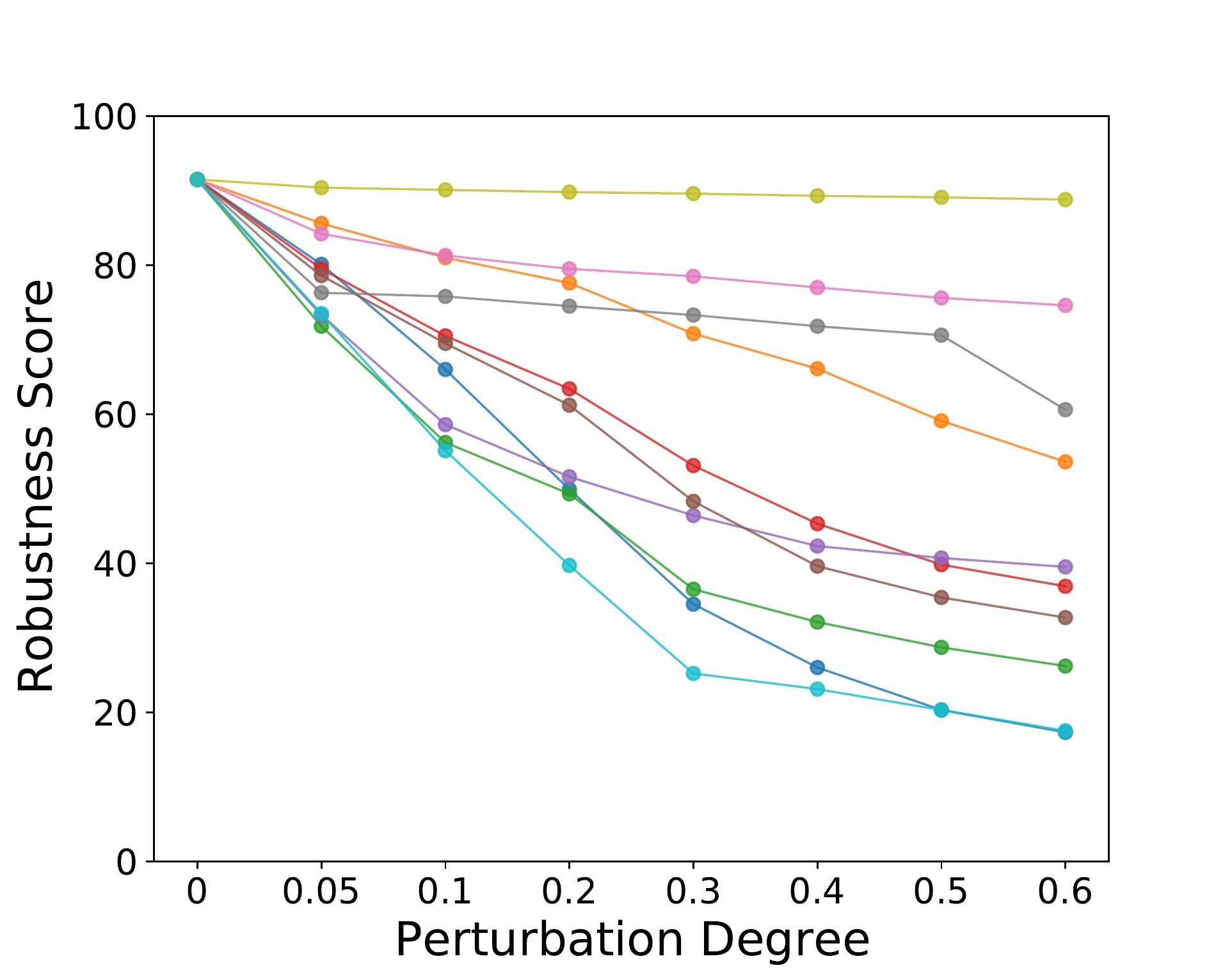}
\end{minipage}
}%
\\
\includegraphics[width=0.95\linewidth]{figs/Presentation1.pdf}

\centering
\caption{\label{fig:comprehensive_results_jigsaw} Comprehensive results of RoBERTa-base (Base) and RoBERTa-large (Large) on Jigsaw. We consider rule-based (Rule) and score-based (Score) attacks, and worst (Worst) and average (Average) performance estimation. }
\end{figure*}

We list results on AG's News in Figure~\ref{fig:comprehensive_results_ag} and results on Jigsaw in Figure~\ref{fig:comprehensive_results_jigsaw}.

\section{Discussion} 
\label{sec:discussion}
\citet{chen2022should} categorizes four different roles of textual adversarial samples.
In this paper, we consider how to employ adversarial attacks for automatic robustness evaluation, corresponding to the defined evaluation role. 
In this section, we give a further discussion about potential future directions on adversarial NLP for robustness evaluation, considering both the attack and the defense sides.

\subsection{Adversarial Attack}


\paragraph{Complemented robustness dimension} We consider general and representative robustness dimensions in our framework. 
We hope that future work can identify more important dimensions spanning three transformation levels to complement the framework. 
Specifically, task-specific dimensions can be explored for more specific and comprehensive evaluation.

\paragraph{Reliable evaluation} 
For invalid adversarial sample filtering, we employ a heuristic weighted average in our framework. 
Further improvement is needed for a more reliable robustness estimation.
The potential directions are:
(1) Identify specific metrics that are justifiable for expected valid adversarial samples; 
(2) Thoroughly investigate the problem of validity-aware robustness evaluation. For example, one can improve our method by using the human annotation results to better characterize the difference between various attack methods since there exist methods that can craft valid adversarial samples even in high perturbation degrees. 
Thus, the human annotation scores can serve as weights to average robustness scores computed at different perturbation degrees.

 \paragraph{Develop methods based on the model-centric evaluation.}
    The motivation of this paper is to bring out the more practical significance of attack methods.
    The core part is to shift towards model-centric robustness evaluation and consider how attack methods can actually contribute to the practitioners. 
    Thus, we recommend future research make a mild shift in method development to better fit the model-centric robustness evaluation scene.     
    For example, the central problem in the adversarial arms race era is how to make the attack methods stronger to achieve a higher attack success rate and beat the defense methods. 
    Now the model-centric evaluation requires that the attack methods can better reveal practical, important, and diversified vulnerabilities in models.

\paragraph{Additional work}
We note that there are some adversarial methods that don't fit into our paradigm because we cannot clearly describe the concrete distribution shift, including challenging samples generated by the human-in-the-loop process~\cite{wallace2019trick, wallace2021analyzing, kiela2021dynabench}, non-dimension-specified attack methods~\cite{bartolo2021improving, guo2021towards, deng2022valcat}.
Future works can explore characterizing the distribution shift through natural language~\cite{zhong2022summarizing} or model estimation~\cite{aharoni2020unsupervised,chronopoulou2021efficient} to include more dimensions in the evaluation framework.

\subsection{Adversarial Defense.}
In our evaluation framework, we don't approach the defense side.
We leave it for future work.
Here we discuss how we consider adversarial defense methods and how we can benefit from them.

Current practices often situate their defense methods in the scenario of malicious attacks.
We present an alternative perspective that accompanies our framework.
As adversarial attack methods can be employed to generate samples from different distributions, defense methods can also be employed to deal with out-of-distribution samples, which can address the challenge of diverse inputs from different users or attackers. 
However, the deficiency in current defense methods is that they mostly can only tackle a specific kind of distribution shift. 
For example, \citet{pruthi-etal-2019-combating} consider samples containing typos. 
\citet{wang2021natural} consider rich vocabulary of real-world users. 
Currently, a generalized and widely applicable  defense method is lacking.
The promising directions include: (1) Inference-time adaptation~\cite{antverg2022idani};
(2) Learning robust features from in-distribution data~\cite{ilyas2019adversarial, clark2019don, zhou2021examining};
(3) Distributionally robust optimization~\cite{hu2018does, oren2019distributionally}.

\section{Single-model Robustness Report}
\label{sec:rob_report}
\begin{figure*}[htbp]
\vspace{-1.25mm}
\captionsetup[subfigure]{labelformat=empty} 
\begin{subfigure}{
\includegraphics[width=2.9in,page=1]{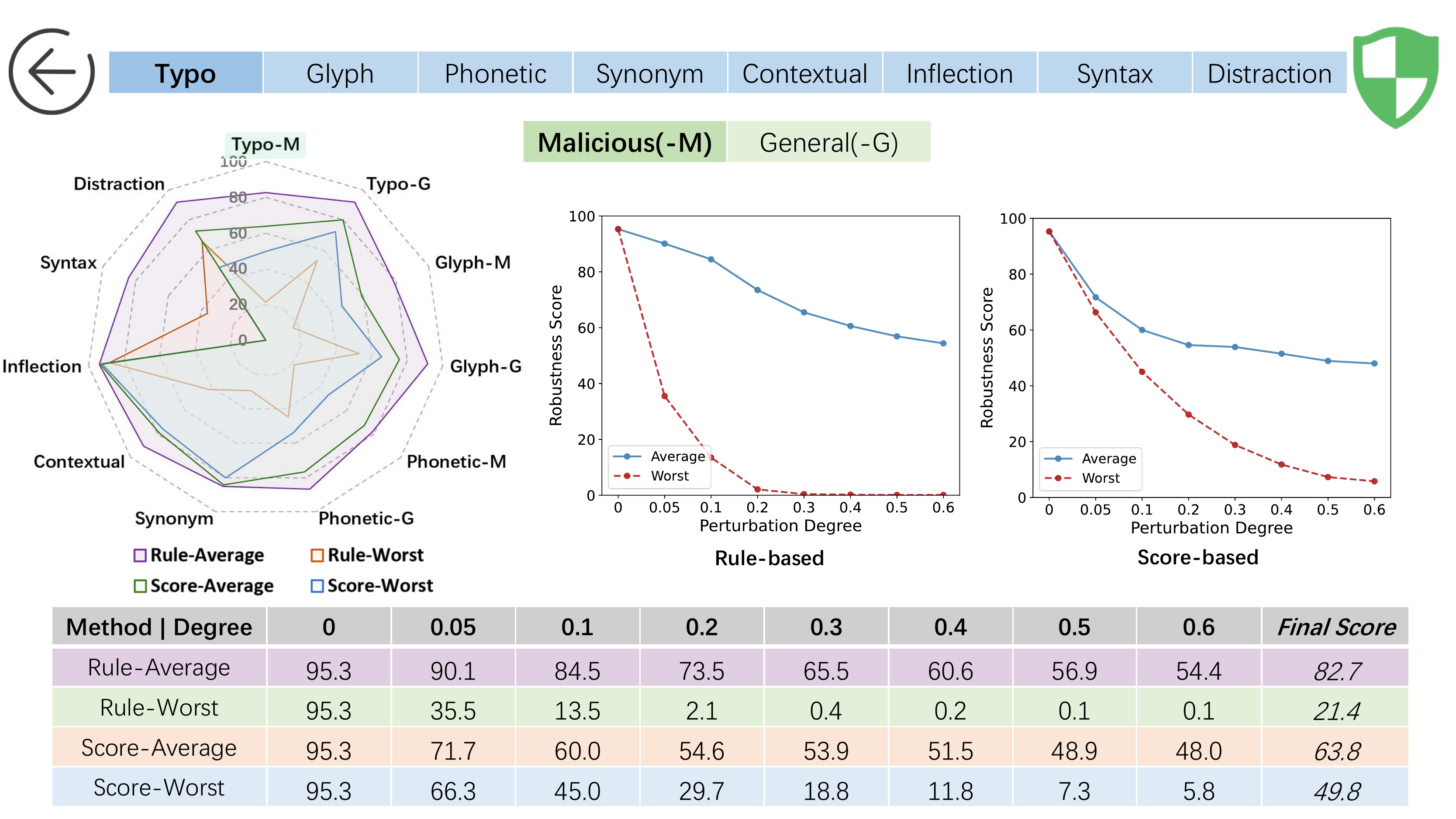}}
\end{subfigure}
\vspace{-1.25mm}
\begin{subfigure}{
\includegraphics[width=2.9in,page=2]{figs/rob_report_sst_base.pdf}}
\end{subfigure}
\vspace{-1.25mm}
\begin{subfigure}{
\includegraphics[width=2.9in,page=3]{figs/rob_report_sst_base.pdf}}
\end{subfigure}
\vspace{-1.25mm}
\begin{subfigure}{
\includegraphics[width=2.9in,page=4]{figs/rob_report_sst_base.pdf}}
\end{subfigure}
\vspace{-1.25mm}
\begin{subfigure}{
\includegraphics[width=2.9in,page=5]{figs/rob_report_sst_base.pdf}}
\end{subfigure}
\vspace{-1.25mm}
\begin{subfigure}{
\includegraphics[width=2.9in,page=6]{figs/rob_report_sst_base.pdf}}
\end{subfigure}
\vspace{-1.25mm}
\begin{subfigure}{
\includegraphics[width=2.9in,page=7]{figs/rob_report_sst_base.pdf}}
\end{subfigure}
\vspace{-1.25mm}
\begin{subfigure}{
\includegraphics[width=2.9in,page=8]{figs/rob_report_sst_base.pdf}}
\end{subfigure}
\vspace{-1.25mm}
\begin{subfigure}{
\includegraphics[width=2.9in,page=9]{figs/rob_report_sst_base.pdf}}
\end{subfigure}
\vspace{-1.25mm}
\begin{subfigure}{
\includegraphics[width=2.9in,page=10]{figs/rob_report_sst_base.pdf}}
\end{subfigure}
\vspace{-1.25mm}
\begin{subfigure}{
\includegraphics[width=2.9in,page=11]{figs/rob_report_sst_base.pdf}}
\end{subfigure}
\vspace{-1.25mm}
\centering
\caption{\label{fig:robust_base_sst} Robustness report for RoBERTa-base on SST-2.}
\end{figure*}

\begin{figure*}[htbp]
\vspace{-1.25mm}
\captionsetup[subfigure]{labelformat=empty} 
\begin{subfigure}{
\includegraphics[width=2.9in,page=1]{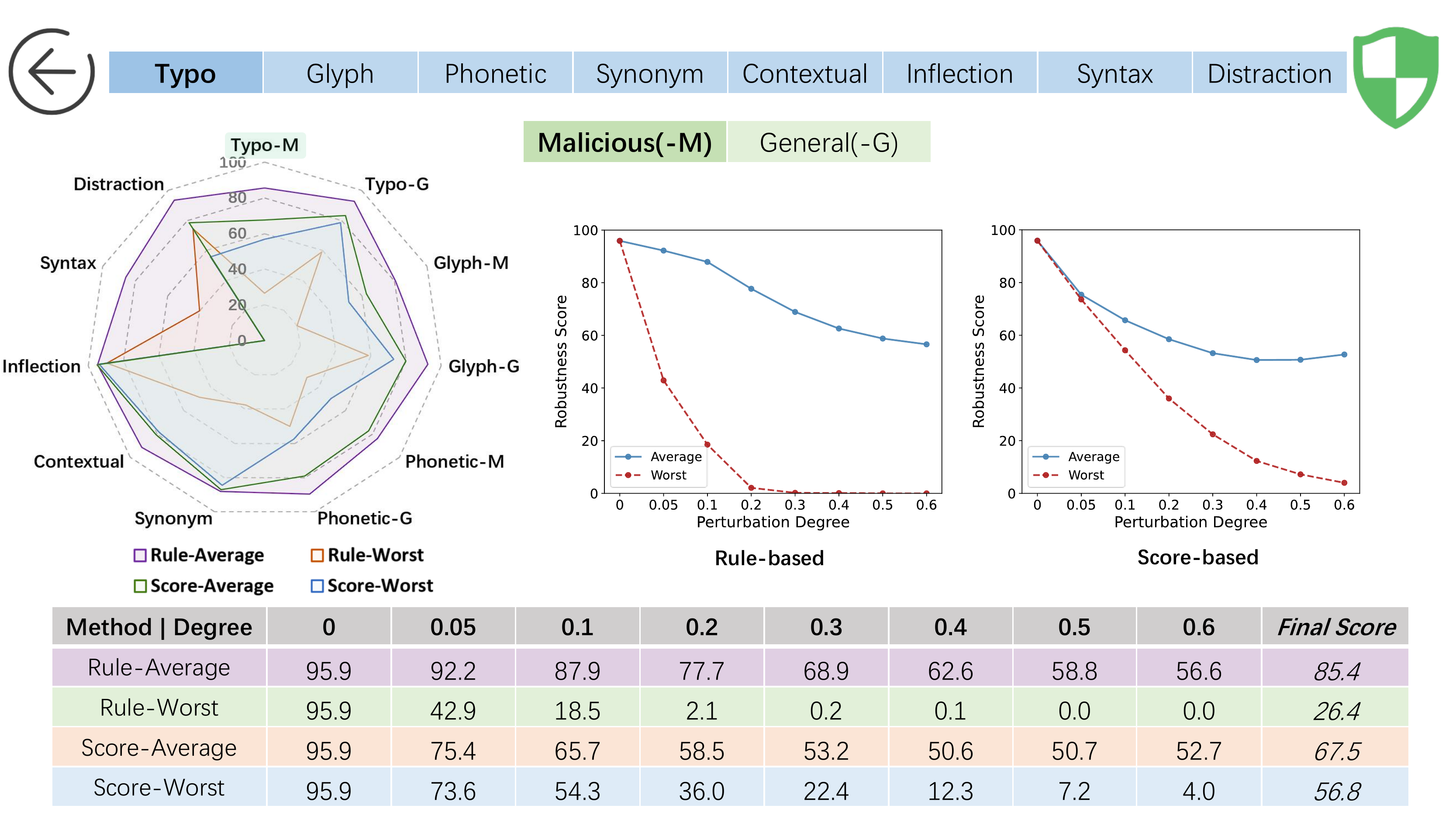}}
\end{subfigure}
\vspace{-1.25mm}
\begin{subfigure}{
\includegraphics[width=2.9in,page=2]{figs/rob_report_sst_big.pdf}}
\end{subfigure}
\vspace{-1.25mm}
\begin{subfigure}{
\includegraphics[width=2.9in,page=3]{figs/rob_report_sst_big.pdf}}
\end{subfigure}
\vspace{-1.25mm}
\begin{subfigure}{
\includegraphics[width=2.9in,page=4]{figs/rob_report_sst_big.pdf}}
\end{subfigure}
\vspace{-1.25mm}
\begin{subfigure}{
\includegraphics[width=2.9in,page=5]{figs/rob_report_sst_big.pdf}}
\end{subfigure}
\vspace{-1.25mm}
\begin{subfigure}{
\includegraphics[width=2.9in,page=6]{figs/rob_report_sst_big.pdf}}
\end{subfigure}
\vspace{-1.25mm}
\begin{subfigure}{
\includegraphics[width=2.9in,page=7]{figs/rob_report_sst_big.pdf}}
\end{subfigure}
\vspace{-1.25mm}
\begin{subfigure}{
\includegraphics[width=2.9in,page=8]{figs/rob_report_sst_big.pdf}}
\end{subfigure}
\vspace{-1.25mm}
\begin{subfigure}{
\includegraphics[width=2.9in,page=9]{figs/rob_report_sst_big.pdf}}
\end{subfigure}
\vspace{-1.25mm}
\begin{subfigure}{
\includegraphics[width=2.9in,page=10]{figs/rob_report_sst_big.pdf}}
\end{subfigure}
\vspace{-1.25mm}
\begin{subfigure}{
\includegraphics[width=2.9in,page=11]{figs/rob_report_sst_big.pdf}}
\end{subfigure}
\vspace{-1.25mm}

\centering
\caption{\label{fig:robust_large_sst} Robustness report for RoBERTa-large on SST-2.}
\end{figure*}

\begin{figure*}[htbp]
\vspace{-1.25mm}
\captionsetup[subfigure]{labelformat=empty} 
\begin{subfigure}{
\includegraphics[width=2.9in,page=1]{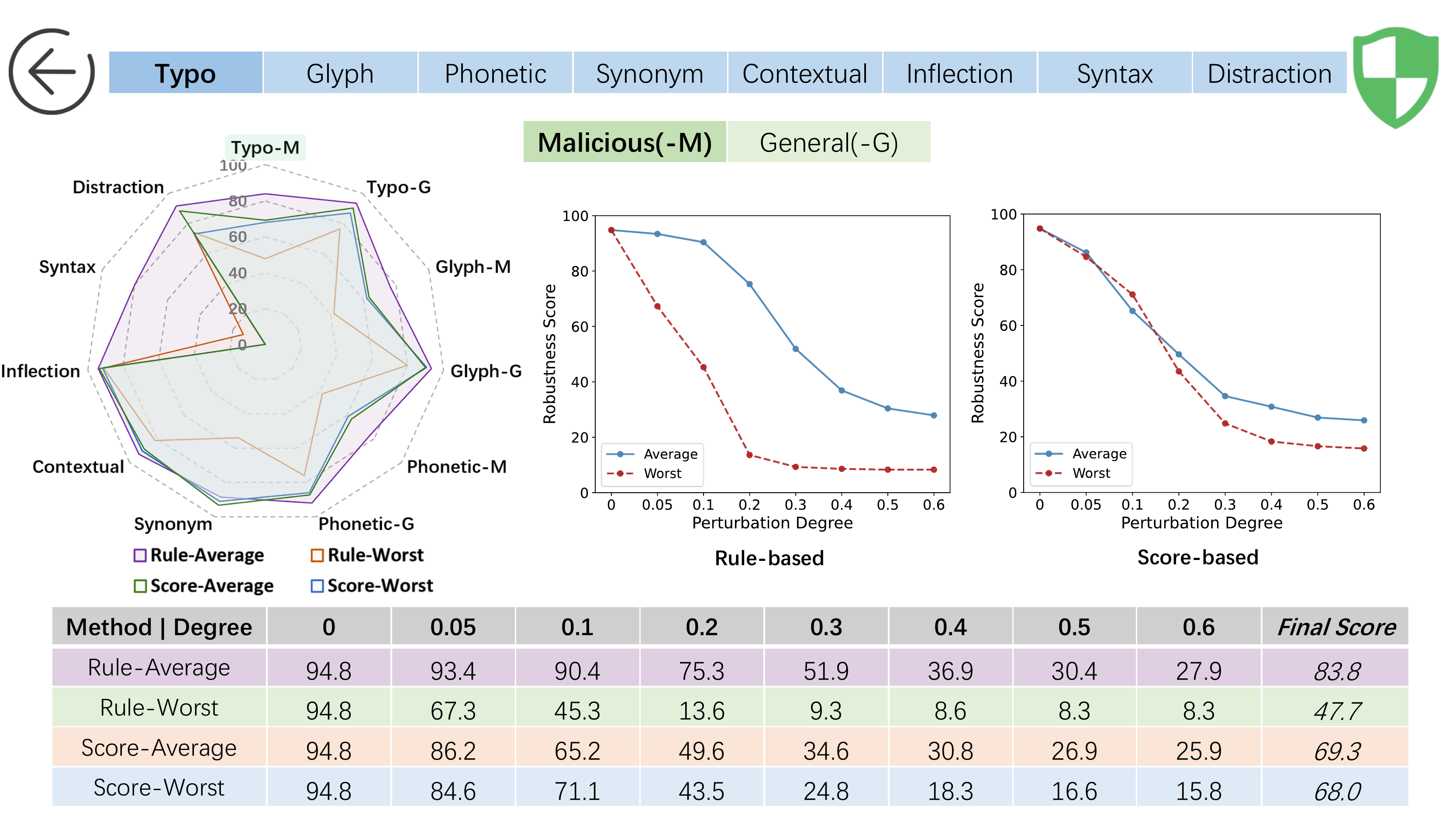}}
\end{subfigure}
\vspace{-1.25mm}
\begin{subfigure}{
\includegraphics[width=2.9in,page=2]{figs/rob_report_ag_base.pdf}}
\end{subfigure}
\vspace{-1.25mm}
\begin{subfigure}{
\includegraphics[width=2.9in,page=3]{figs/rob_report_ag_base.pdf}}
\end{subfigure}
\vspace{-1.25mm}
\begin{subfigure}{
\includegraphics[width=2.9in,page=4]{figs/rob_report_ag_base.pdf}}
\end{subfigure}
\vspace{-1.25mm}
\begin{subfigure}{
\includegraphics[width=2.9in,page=5]{figs/rob_report_ag_base.pdf}}
\end{subfigure}
\vspace{-1.25mm}
\begin{subfigure}{
\includegraphics[width=2.9in,page=6]{figs/rob_report_ag_base.pdf}}
\end{subfigure}
\vspace{-1.25mm}
\begin{subfigure}{
\includegraphics[width=2.9in,page=7]{figs/rob_report_ag_base.pdf}}
\end{subfigure}
\vspace{-1.25mm}
\begin{subfigure}{
\includegraphics[width=2.9in,page=8]{figs/rob_report_ag_base.pdf}}
\end{subfigure}
\vspace{-1.25mm}
\begin{subfigure}{
\includegraphics[width=2.9in,page=9]{figs/rob_report_ag_base.pdf}}
\end{subfigure}
\vspace{-1.25mm}
\begin{subfigure}{
\includegraphics[width=2.9in,page=10]{figs/rob_report_ag_base.pdf}}
\end{subfigure}
\vspace{-1.25mm}
\begin{subfigure}{
\includegraphics[width=2.9in,page=11]{figs/rob_report_ag_base.pdf}}
\end{subfigure}
\vspace{-1.25mm}

\centering
\caption{\label{fig:robust_base_ag} Robustness report for RoBERTa-base on AG's News.}
\end{figure*}

\begin{figure*}[htbp]
\vspace{-1.25mm}
\captionsetup[subfigure]{labelformat=empty} 
\begin{subfigure}{
\includegraphics[width=2.9in,page=1]{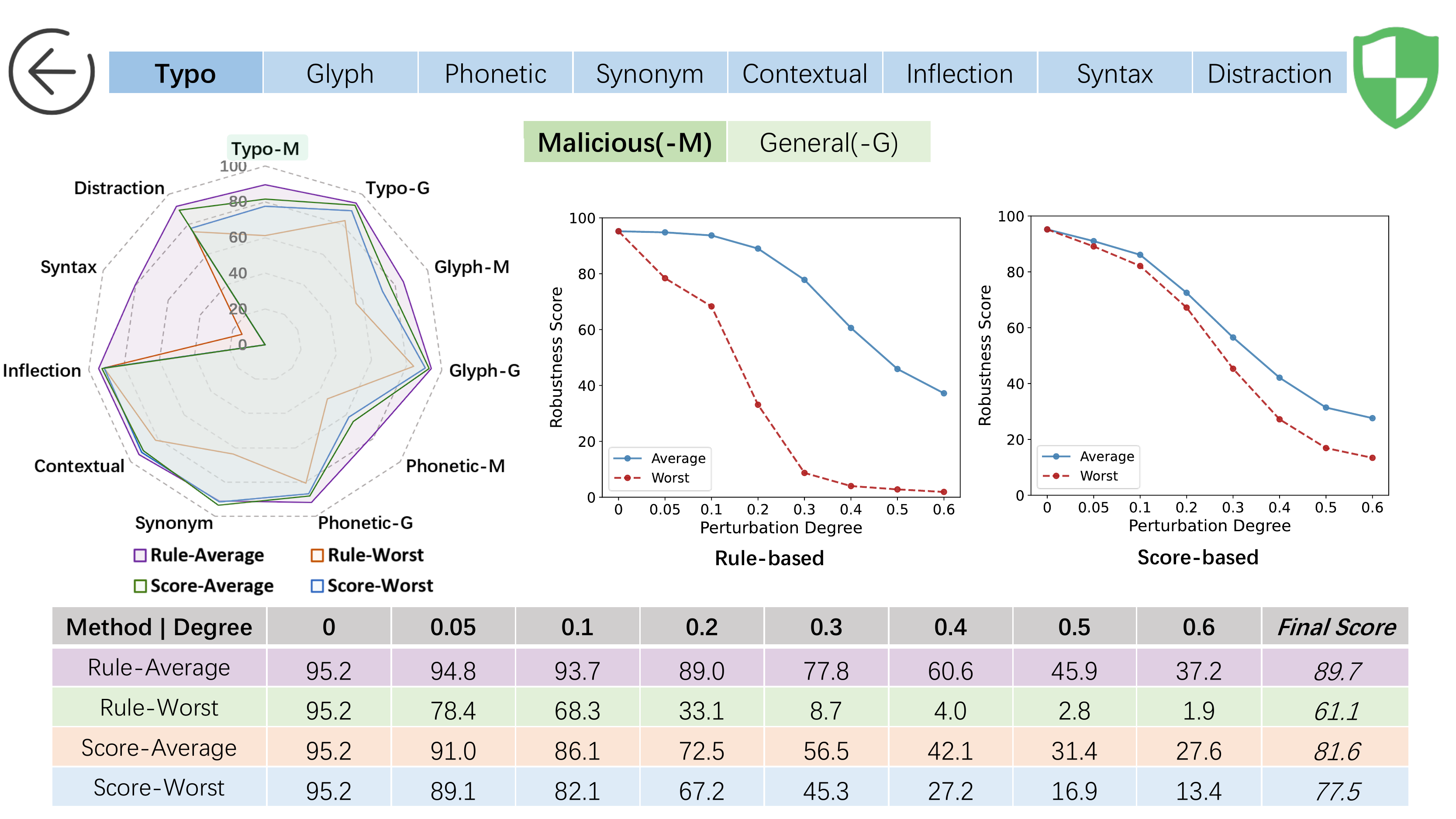}}
\end{subfigure}
\vspace{-1.25mm}
\begin{subfigure}{
\includegraphics[width=2.9in,page=2]{figs/rob_report_ag_big.pdf}}
\end{subfigure}
\vspace{-1.25mm}
\begin{subfigure}{
\includegraphics[width=2.9in,page=3]{figs/rob_report_ag_big.pdf}}
\end{subfigure}
\vspace{-1.25mm}
\begin{subfigure}{
\includegraphics[width=2.9in,page=4]{figs/rob_report_ag_big.pdf}}
\end{subfigure}
\vspace{-1.25mm}
\begin{subfigure}{
\includegraphics[width=2.9in,page=5]{figs/rob_report_ag_big.pdf}}
\end{subfigure}
\vspace{-1.25mm}
\begin{subfigure}{
\includegraphics[width=2.9in,page=6]{figs/rob_report_ag_big.pdf}}
\end{subfigure}
\vspace{-1.25mm}
\begin{subfigure}{
\includegraphics[width=2.9in,page=7]{figs/rob_report_ag_big.pdf}}
\end{subfigure}
\vspace{-1.25mm}
\begin{subfigure}{
\includegraphics[width=2.9in,page=8]{figs/rob_report_ag_big.pdf}}
\end{subfigure}
\vspace{-1.25mm}
\begin{subfigure}{
\includegraphics[width=2.9in,page=9]{figs/rob_report_ag_big.pdf}}
\end{subfigure}
\vspace{-1.25mm}
\begin{subfigure}{
\includegraphics[width=2.9in,page=10]{figs/rob_report_ag_big.pdf}}
\end{subfigure}
\vspace{-1.25mm}
\begin{subfigure}{
\includegraphics[width=2.9in,page=11]{figs/rob_report_ag_big.pdf}}
\end{subfigure}
\vspace{-1.25mm}

\centering
\caption{\label{fig:robust_large_ag} Robustness report for RoBERTa-large on AG's News.}
\end{figure*}

\begin{figure*}[htbp]
\vspace{-1.25mm}
\captionsetup[subfigure]{labelformat=empty} 
\begin{subfigure}{
\includegraphics[width=2.9in,page=1]{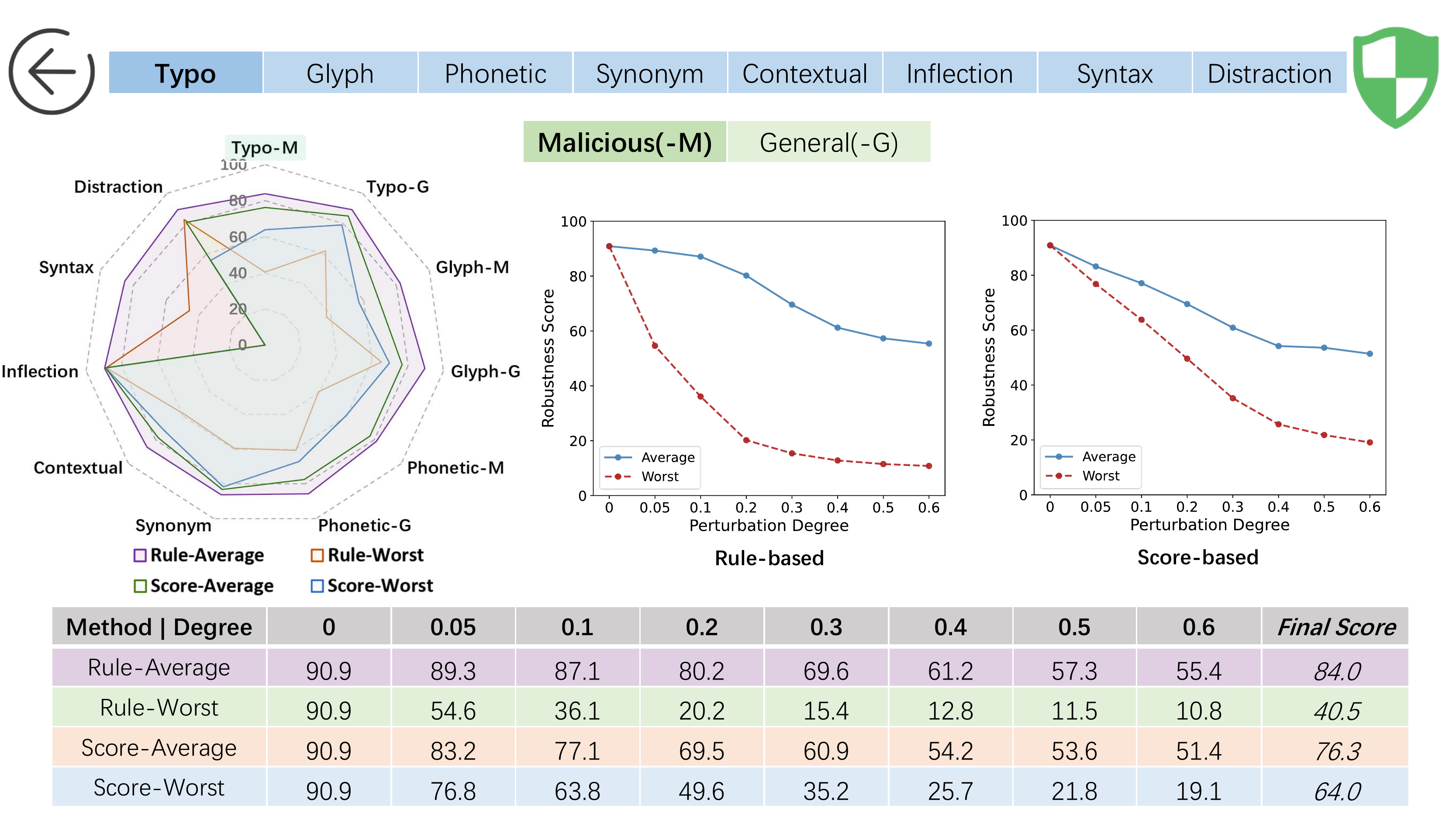}}
\end{subfigure}
\vspace{-1.25mm}
\begin{subfigure}{
\includegraphics[width=2.9in,page=2]{figs/rob_report_jigsaw_base.pdf}}
\end{subfigure}
\vspace{-1.25mm}
\begin{subfigure}{
\includegraphics[width=2.9in,page=3]{figs/rob_report_jigsaw_base.pdf}}
\end{subfigure}
\vspace{-1.25mm}
\begin{subfigure}{
\includegraphics[width=2.9in,page=4]{figs/rob_report_jigsaw_base.pdf}}
\end{subfigure}
\vspace{-1.25mm}
\begin{subfigure}{
\includegraphics[width=2.9in,page=5]{figs/rob_report_jigsaw_base.pdf}}
\end{subfigure}
\vspace{-1.25mm}
\begin{subfigure}{
\includegraphics[width=2.9in,page=6]{figs/rob_report_jigsaw_base.pdf}}
\end{subfigure}
\vspace{-1.25mm}
\begin{subfigure}{
\includegraphics[width=2.9in,page=7]{figs/rob_report_jigsaw_base.pdf}}
\end{subfigure}
\vspace{-1.25mm}
\begin{subfigure}{
\includegraphics[width=2.9in,page=8]{figs/rob_report_jigsaw_base.pdf}}
\end{subfigure}
\vspace{-1.25mm}
\begin{subfigure}{
\includegraphics[width=2.9in,page=9]{figs/rob_report_jigsaw_base.pdf}}
\end{subfigure}
\vspace{-1.25mm}
\begin{subfigure}{
\includegraphics[width=2.9in,page=10]{figs/rob_report_jigsaw_base.pdf}}
\end{subfigure}
\vspace{-1.25mm}
\begin{subfigure}{
\includegraphics[width=2.9in,page=11]{figs/rob_report_jigsaw_base.pdf}}
\end{subfigure}
\vspace{-1.25mm}

\centering
\caption{\label{fig:robust_base_jigsaw} Robustness report for RoBERTa-base on Jigsaw.}
\end{figure*}

\begin{figure*}[htbp]
\vspace{-1.25mm}
\captionsetup[subfigure]{labelformat=empty} 
\begin{subfigure}{
\includegraphics[width=2.9in,page=1]{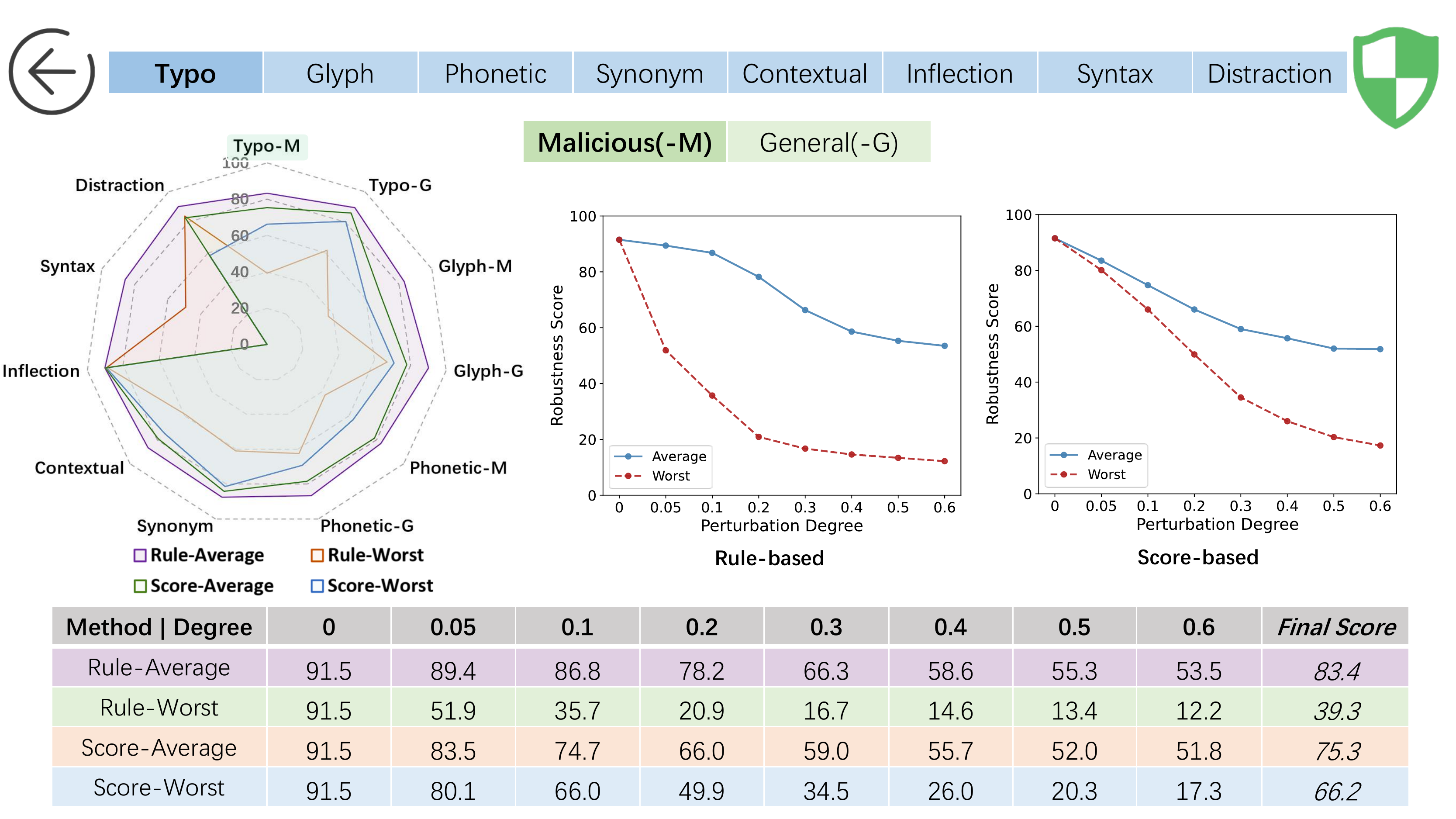}}
\end{subfigure}
\vspace{-1.25mm}
\begin{subfigure}{
\includegraphics[width=2.9in,page=2]{figs/rob_report_jigsaw_large.pdf}}
\end{subfigure}
\vspace{-1.25mm}
\begin{subfigure}{
\includegraphics[width=2.9in,page=3]{figs/rob_report_jigsaw_large.pdf}}
\end{subfigure}
\vspace{-1.25mm}
\begin{subfigure}{
\includegraphics[width=2.9in,page=4]{figs/rob_report_jigsaw_large.pdf}}
\end{subfigure}
\vspace{-1.25mm}
\begin{subfigure}{
\includegraphics[width=2.9in,page=5]{figs/rob_report_jigsaw_large.pdf}}
\end{subfigure}
\vspace{-1.25mm}
\begin{subfigure}{
\includegraphics[width=2.9in,page=6]{figs/rob_report_jigsaw_large.pdf}}
\end{subfigure}
\vspace{-1.25mm}
\begin{subfigure}{
\includegraphics[width=2.9in,page=7]{figs/rob_report_jigsaw_large.pdf}}
\end{subfigure}
\vspace{-1.25mm}
\begin{subfigure}{
\includegraphics[width=2.9in,page=8]{figs/rob_report_jigsaw_large.pdf}}
\end{subfigure}
\vspace{-1.25mm}
\begin{subfigure}{
\includegraphics[width=2.9in,page=9]{figs/rob_report_jigsaw_large.pdf}}
\end{subfigure}
\vspace{-1.25mm}
\begin{subfigure}{
\includegraphics[width=2.9in,page=10]{figs/rob_report_jigsaw_large.pdf}}
\end{subfigure}
\vspace{-1.25mm}
\begin{subfigure}{
\includegraphics[width=2.9in,page=11]{figs/rob_report_jigsaw_large.pdf}}
\end{subfigure}
\vspace{-1.25mm}

\centering
\caption{\label{fig:robust_large_jigsaw} Robustness report for RoBERTa-large on Jigsaw.}
\end{figure*}

We show robustness reports of two models and three datasets. 
The robustness reports for RoBERTa-base are shown in Figure~\ref{fig:robust_base_sst} (SST-2), Figure~\ref{fig:robust_base_ag} (AG's News), and Figure~\ref{fig:robust_base_jigsaw} (Jigsaw).
The robustness reports for RoBERTa-large are shown in Figure~\ref{fig:robust_large_sst} (SST-2), Figure~\ref{fig:robust_large_ag} (AG's News), and Figure~\ref{fig:robust_large_jigsaw} (Jigsaw).

\section{Robustness Comparison Report}
\label{sec:rob_compar_report}
\begin{figure*}[htbp]
\vspace{-1.25mm}
\captionsetup[subfigure]{labelformat=empty} 
\begin{subfigure}{
\includegraphics[page=1,width=2.9in]{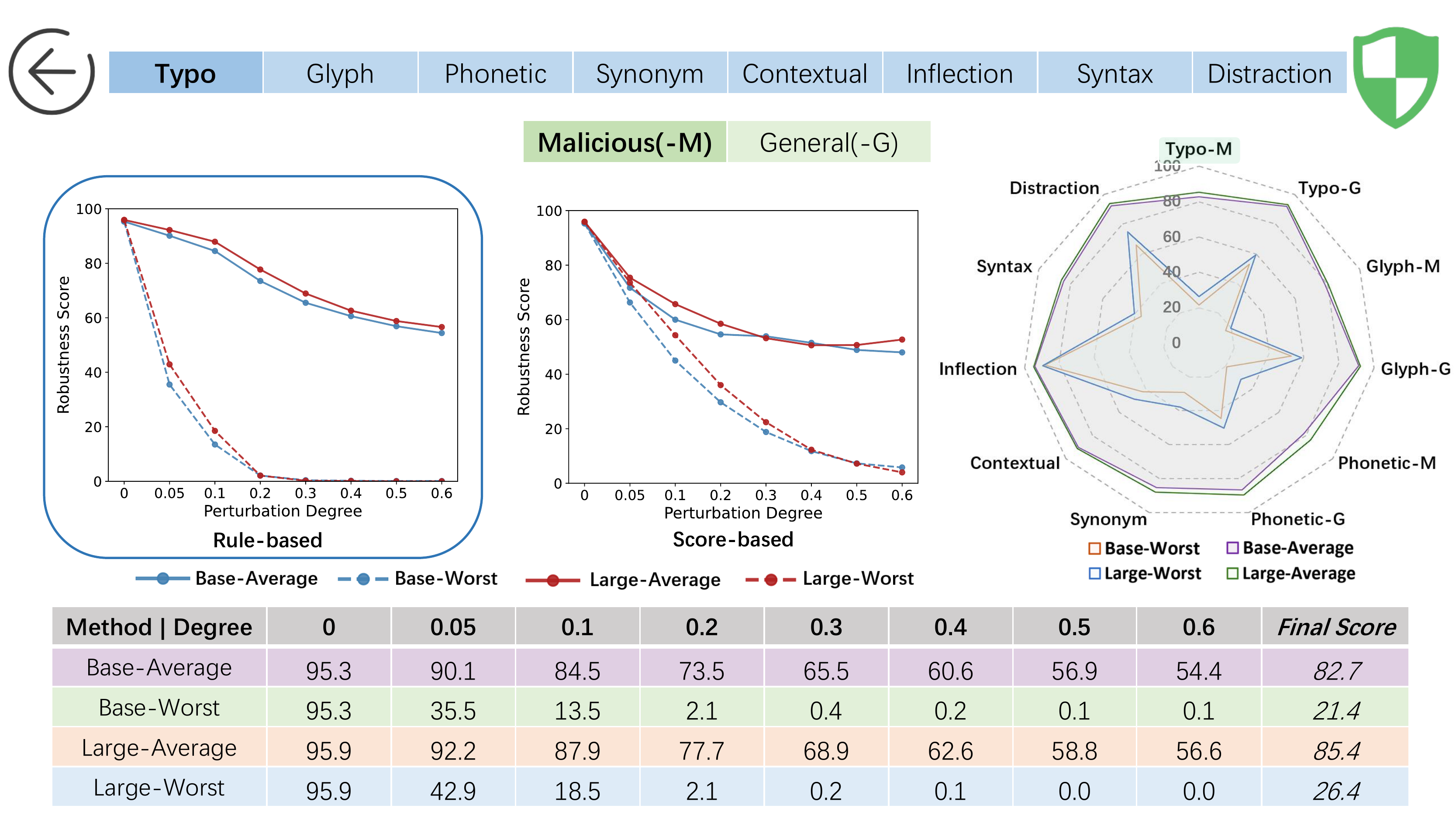}}
\end{subfigure}
\vspace{-1.25mm}
\begin{subfigure}{
\includegraphics[page=2,width=2.9in]{figs/rob_report_sst_com.pdf}}
\end{subfigure}
\vspace{-1.25mm}
\begin{subfigure}{
\includegraphics[width=2.9in,page=3]{figs/rob_report_sst_com.pdf}}
\end{subfigure}
\vspace{-1.25mm}
\begin{subfigure}{
\includegraphics[width=2.9in,page=4]{figs/rob_report_sst_com.pdf}}
\end{subfigure}
\vspace{-1.25mm}
\begin{subfigure}{
\includegraphics[width=2.9in,page=5]{figs/rob_report_sst_com.pdf}}
\end{subfigure}
\vspace{-1.25mm}
\begin{subfigure}{
\includegraphics[width=2.9in,page=6]{figs/rob_report_sst_com.pdf}}
\end{subfigure}
\vspace{-1.25mm}
\begin{subfigure}{
\includegraphics[width=2.9in,page=7]{figs/rob_report_sst_com.pdf}}
\end{subfigure}
\vspace{-1.25mm}
\begin{subfigure}{
\includegraphics[width=2.9in,page=8]{figs/rob_report_sst_com.pdf}}
\end{subfigure}
\vspace{-1.25mm}
\begin{subfigure}{
\includegraphics[width=2.9in,page=9]{figs/rob_report_sst_com.pdf}}
\end{subfigure}
\vspace{-1.25mm}
\begin{subfigure}{
\includegraphics[width=2.9in,page=10]{figs/rob_report_sst_com.pdf}}
\end{subfigure}
\vspace{-1.25mm}
\begin{subfigure}{
\includegraphics[width=2.9in,page=11]{figs/rob_report_sst_com.pdf}}
\end{subfigure}
\vspace{-1.25mm}

\centering
\caption{\label{fig:robust_com_rule} Robustness comparison report for rule-based evaluation on SST-2.}
\end{figure*}

\begin{figure*}[htbp]
\vspace{-1.25mm}
\captionsetup[subfigure]{labelformat=empty} 
\begin{subfigure}{
\includegraphics[width=2.9in,page=12]{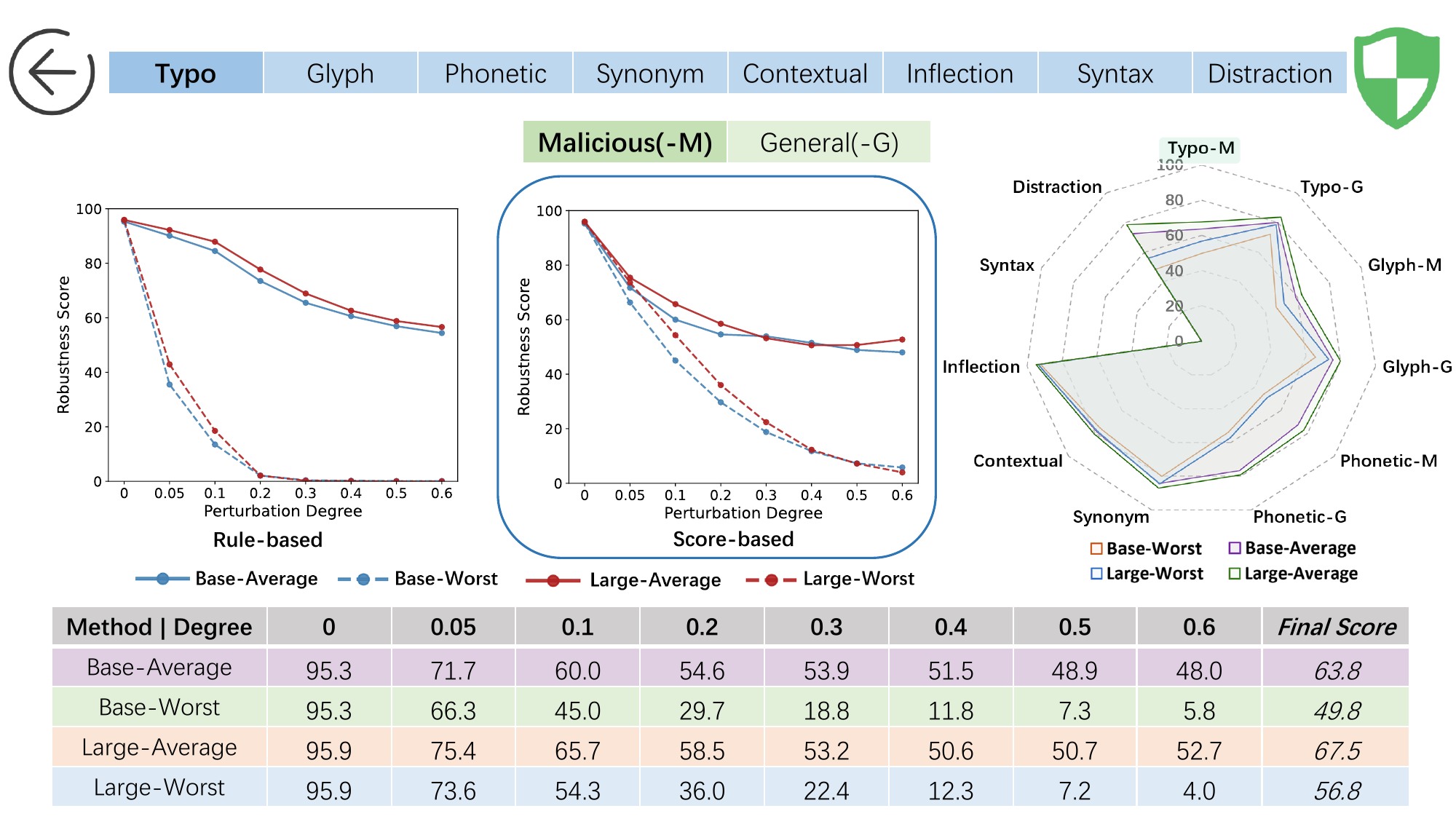}}
\end{subfigure}
\vspace{-1.25mm}
\begin{subfigure}{
\includegraphics[width=2.9in,page=13]{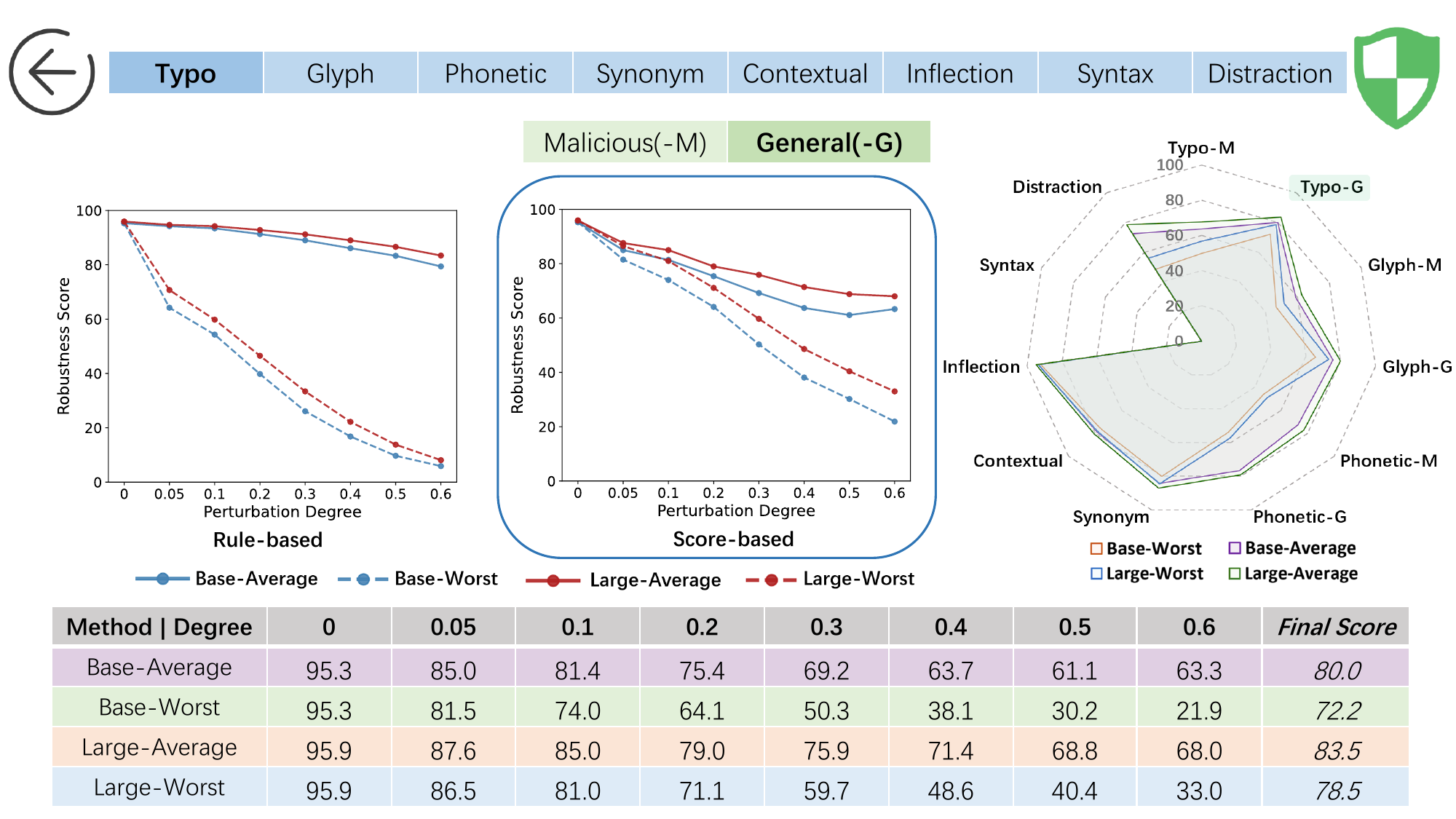}}
\end{subfigure}
\vspace{-1.25mm}
\begin{subfigure}{
\includegraphics[width=2.9in,page=14]{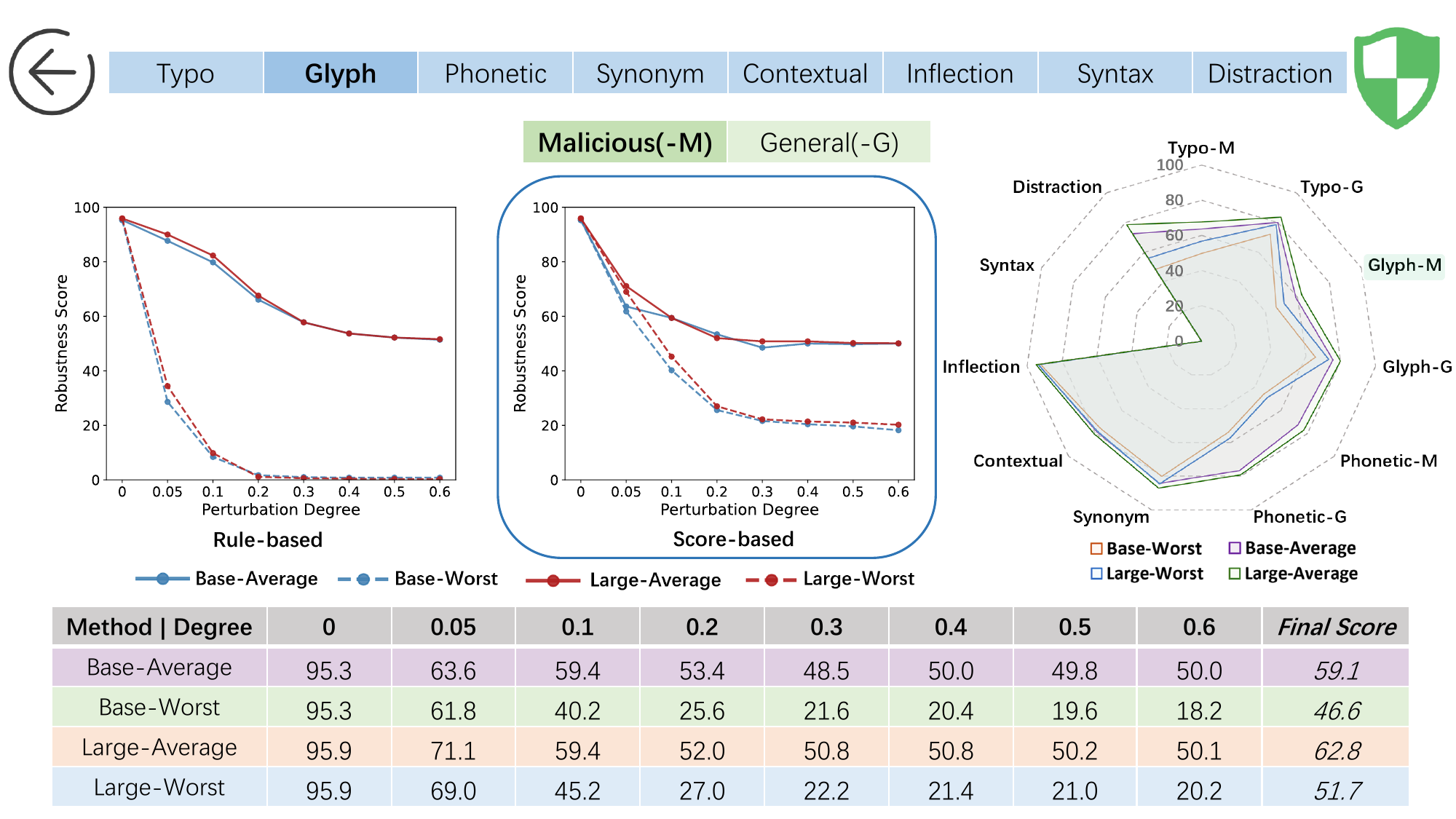}}
\end{subfigure}
\vspace{-1.25mm}
\begin{subfigure}{
\includegraphics[width=2.9in,page=15]{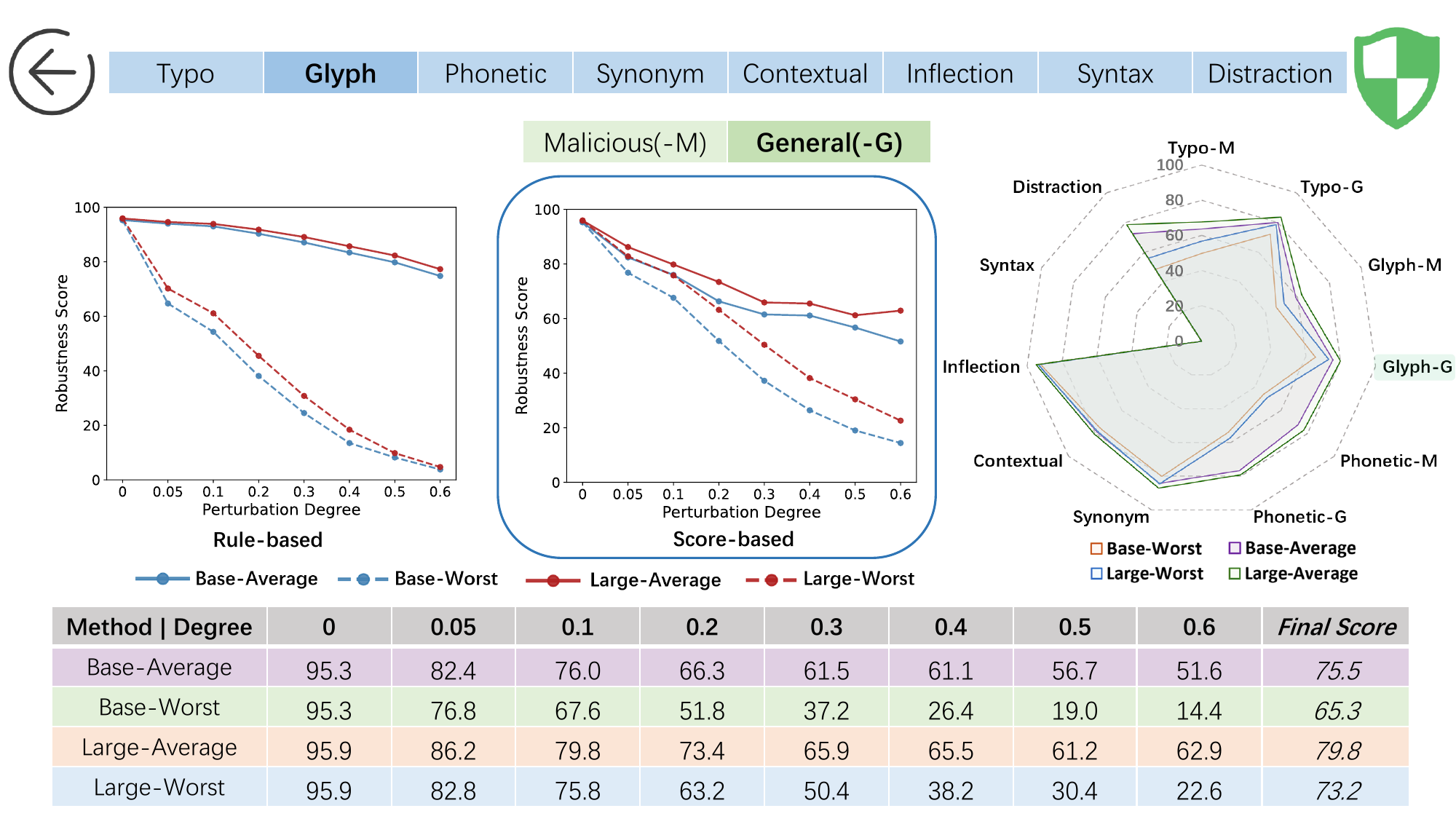}}
\end{subfigure}
\vspace{-1.25mm}
\begin{subfigure}{
\includegraphics[width=2.9in,page=16]{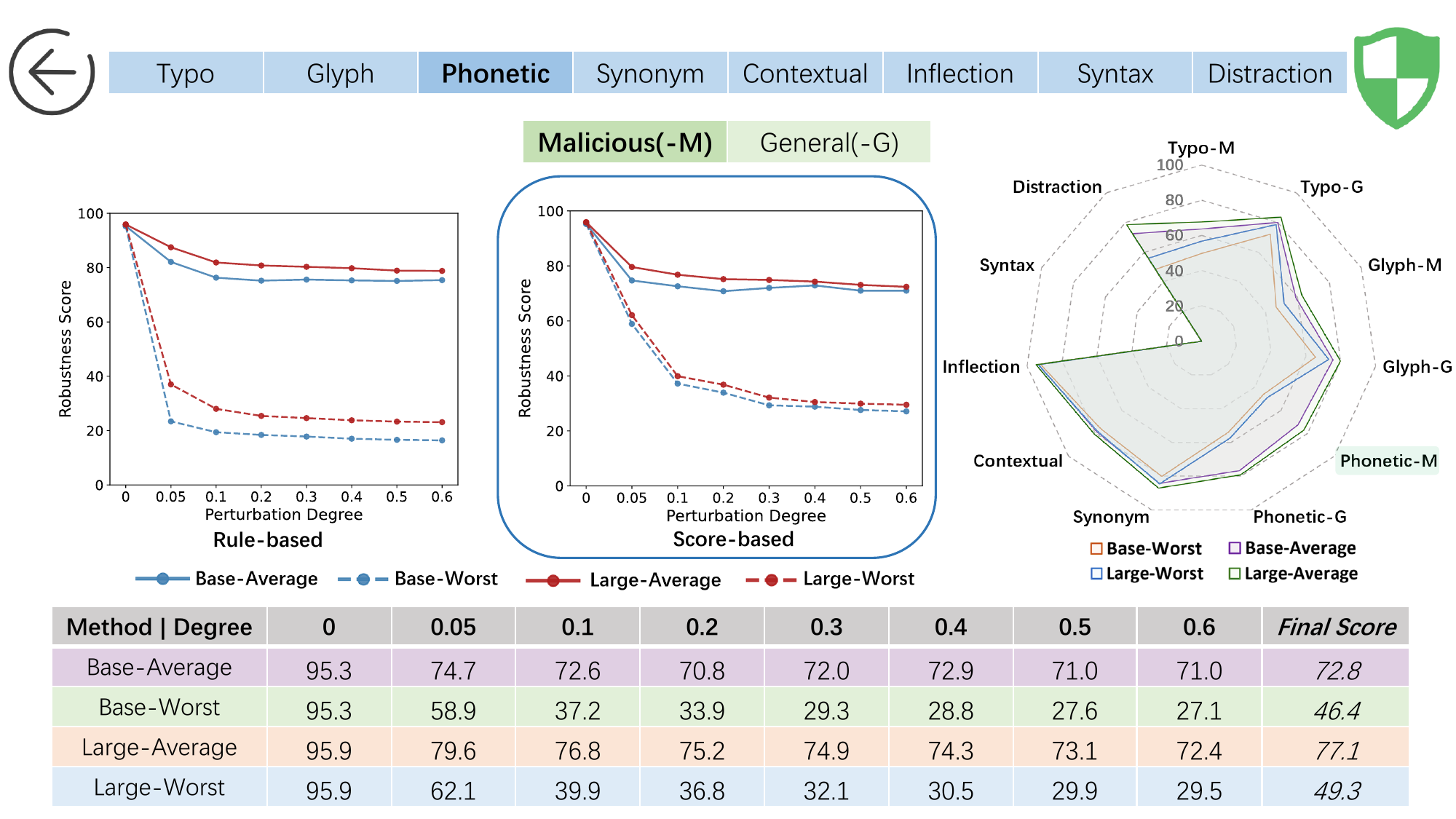}}
\end{subfigure}
\vspace{-1.25mm}
\begin{subfigure}{
\includegraphics[width=2.9in,page=17]{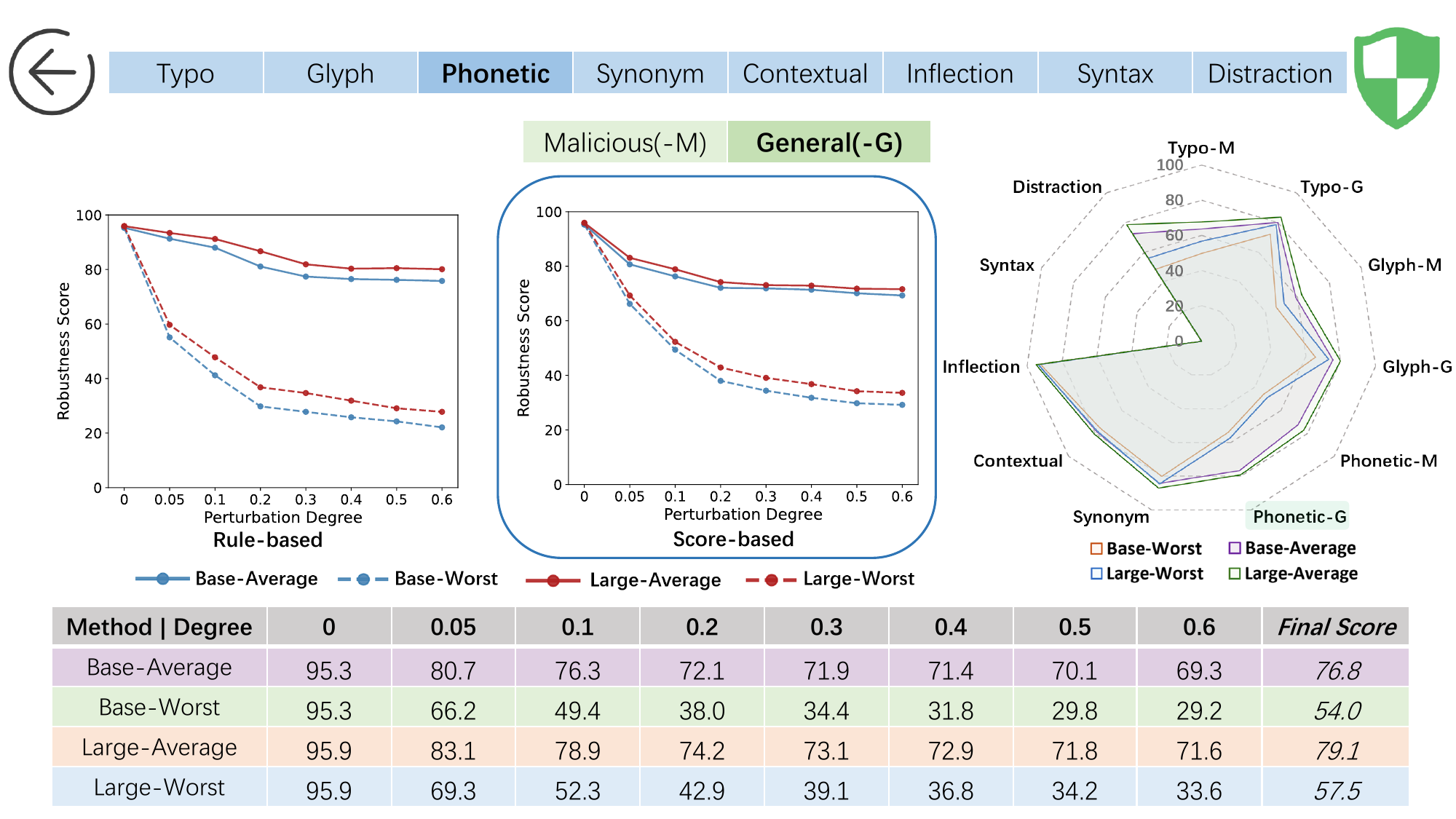}}
\end{subfigure}
\vspace{-1.25mm}
\begin{subfigure}{
\includegraphics[width=2.9in,page=18]{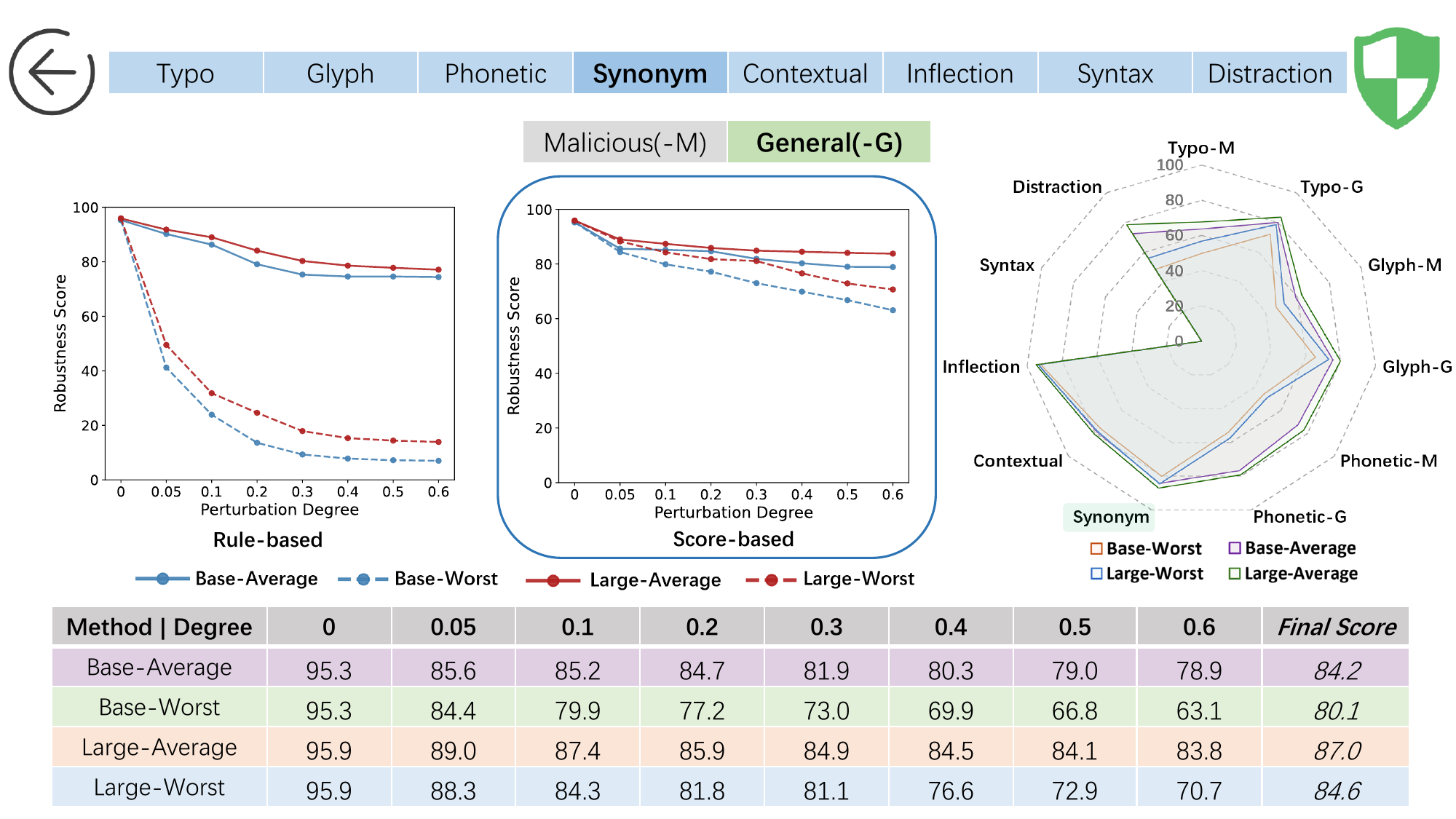}}
\end{subfigure}
\vspace{-1.25mm}
\begin{subfigure}{
\includegraphics[width=2.9in,page=19]{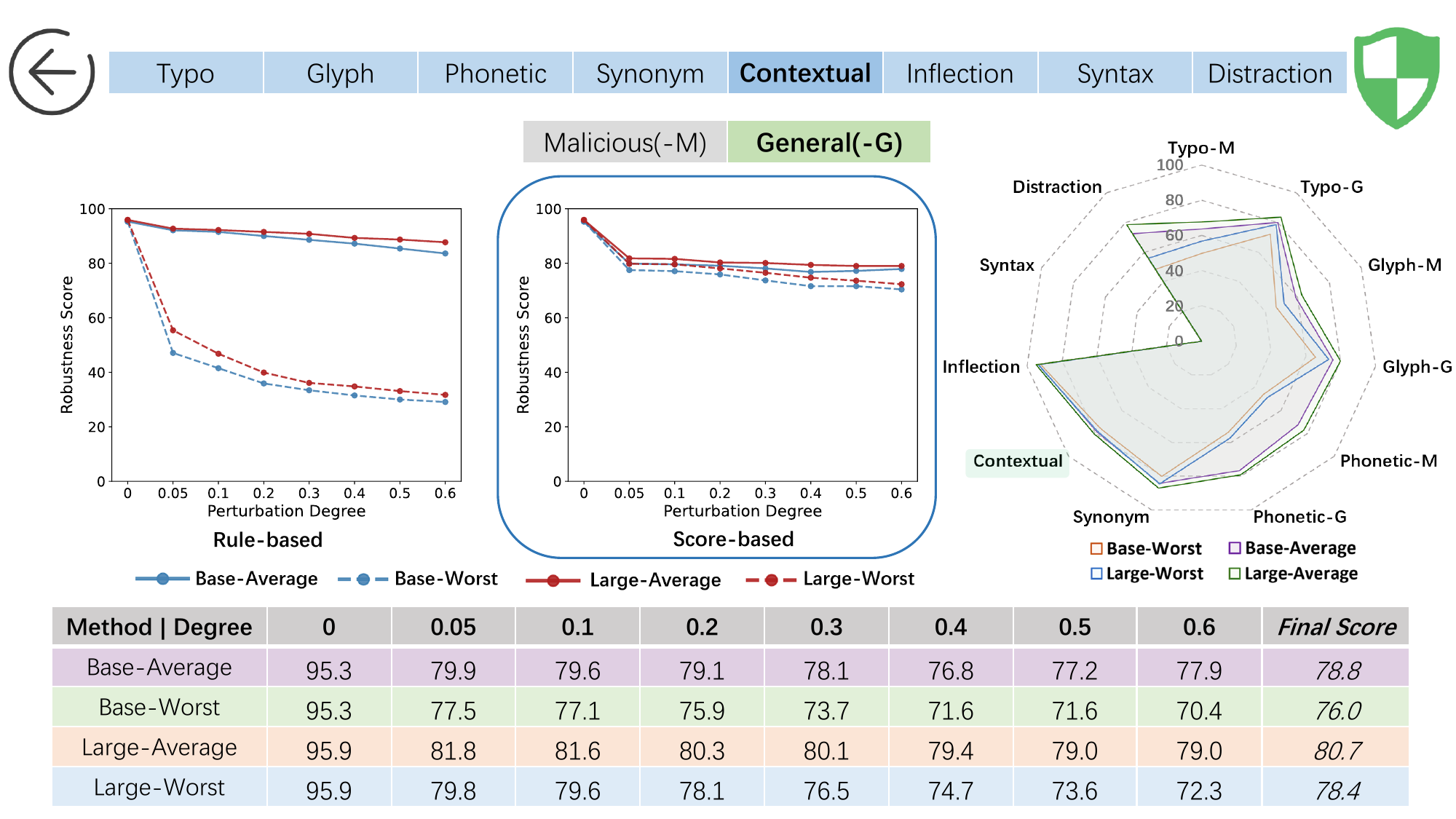}}
\end{subfigure}
\vspace{-1.25mm}
\begin{subfigure}{
\includegraphics[width=2.9in,page=20]{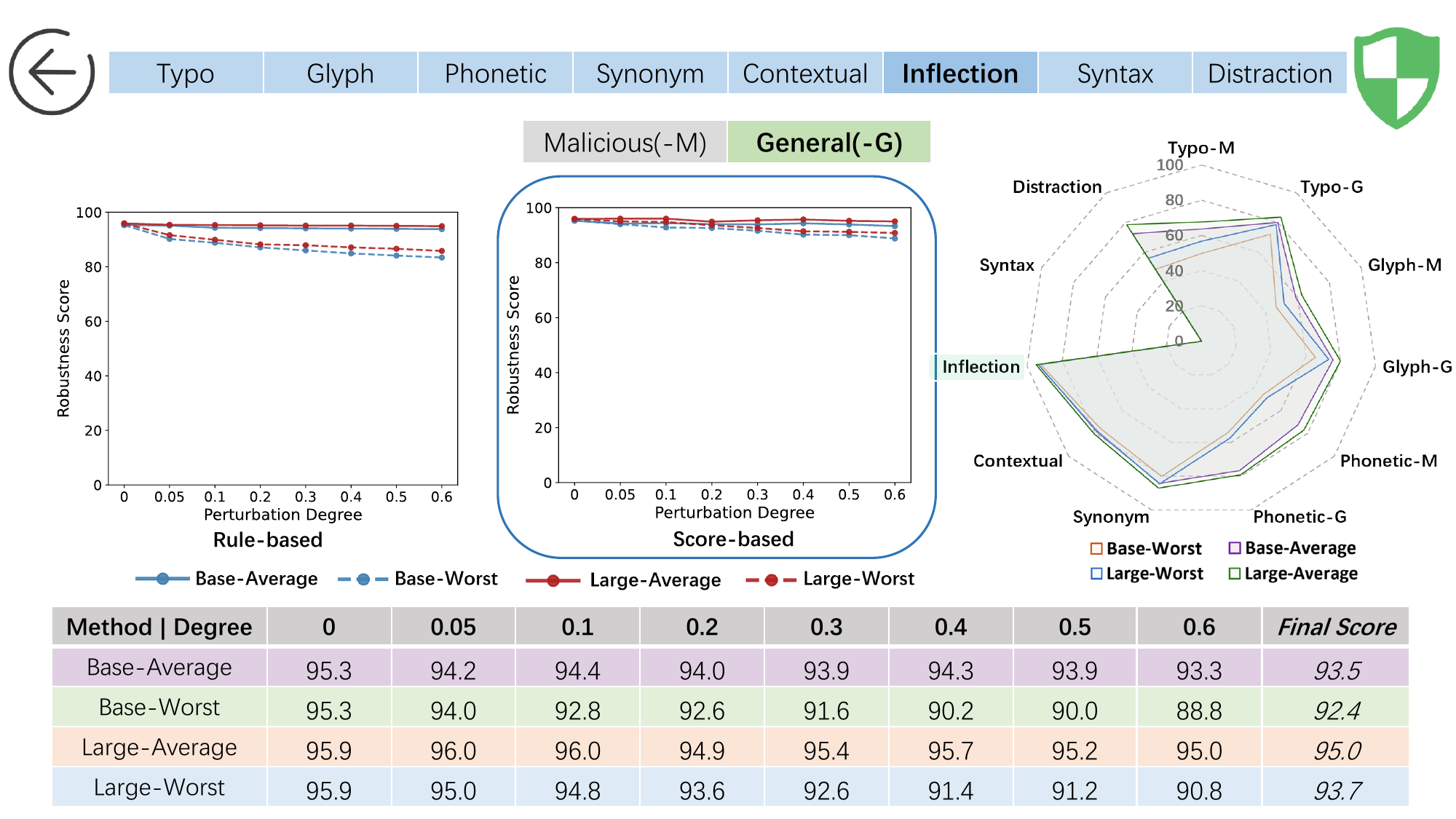}}
\end{subfigure}
\vspace{-1.25mm}
\begin{subfigure}{
\includegraphics[width=2.9in,page=21]{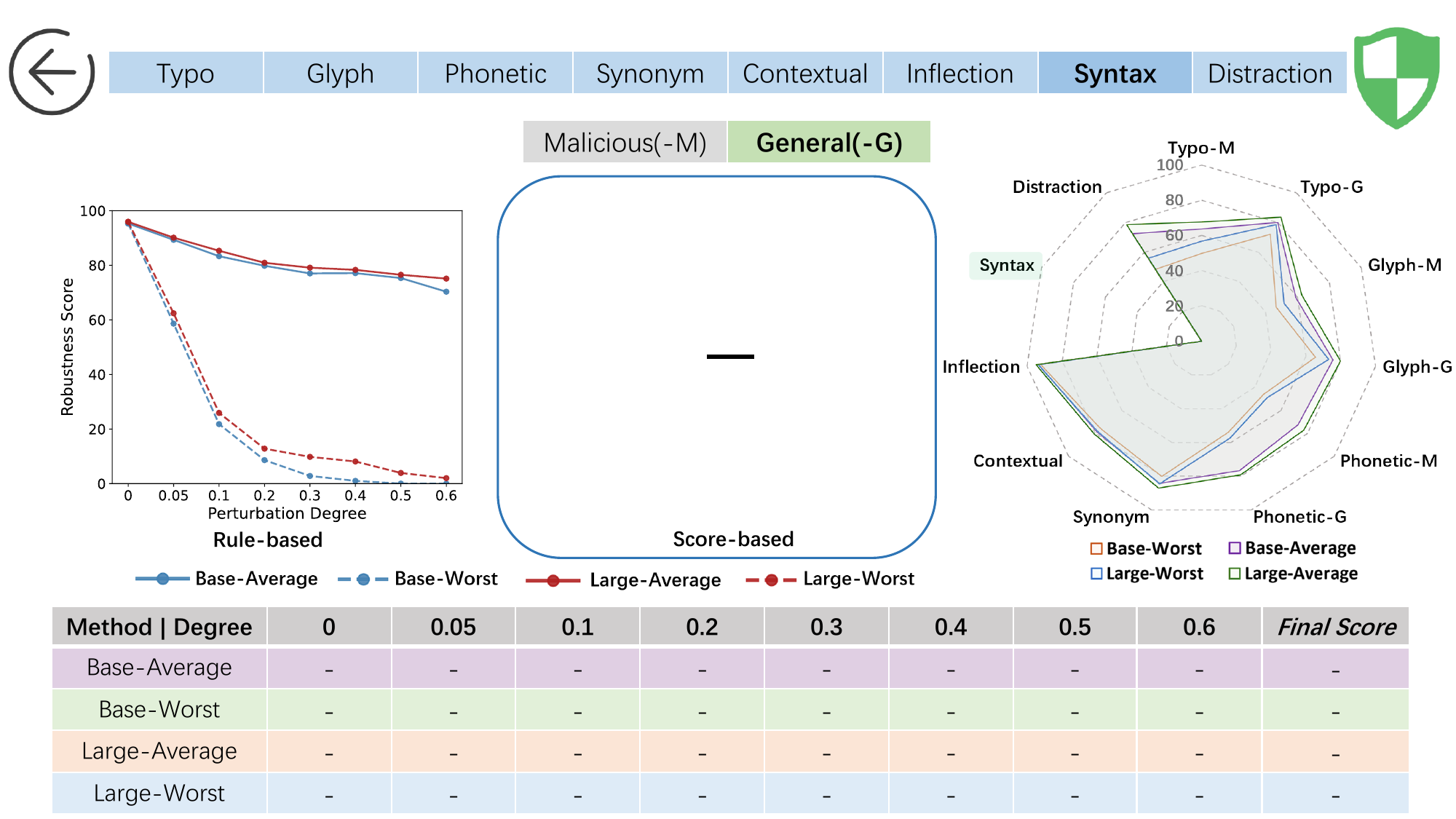}}
\end{subfigure}
\vspace{-1.25mm}
\begin{subfigure}{
\includegraphics[width=2.9in,page=22]{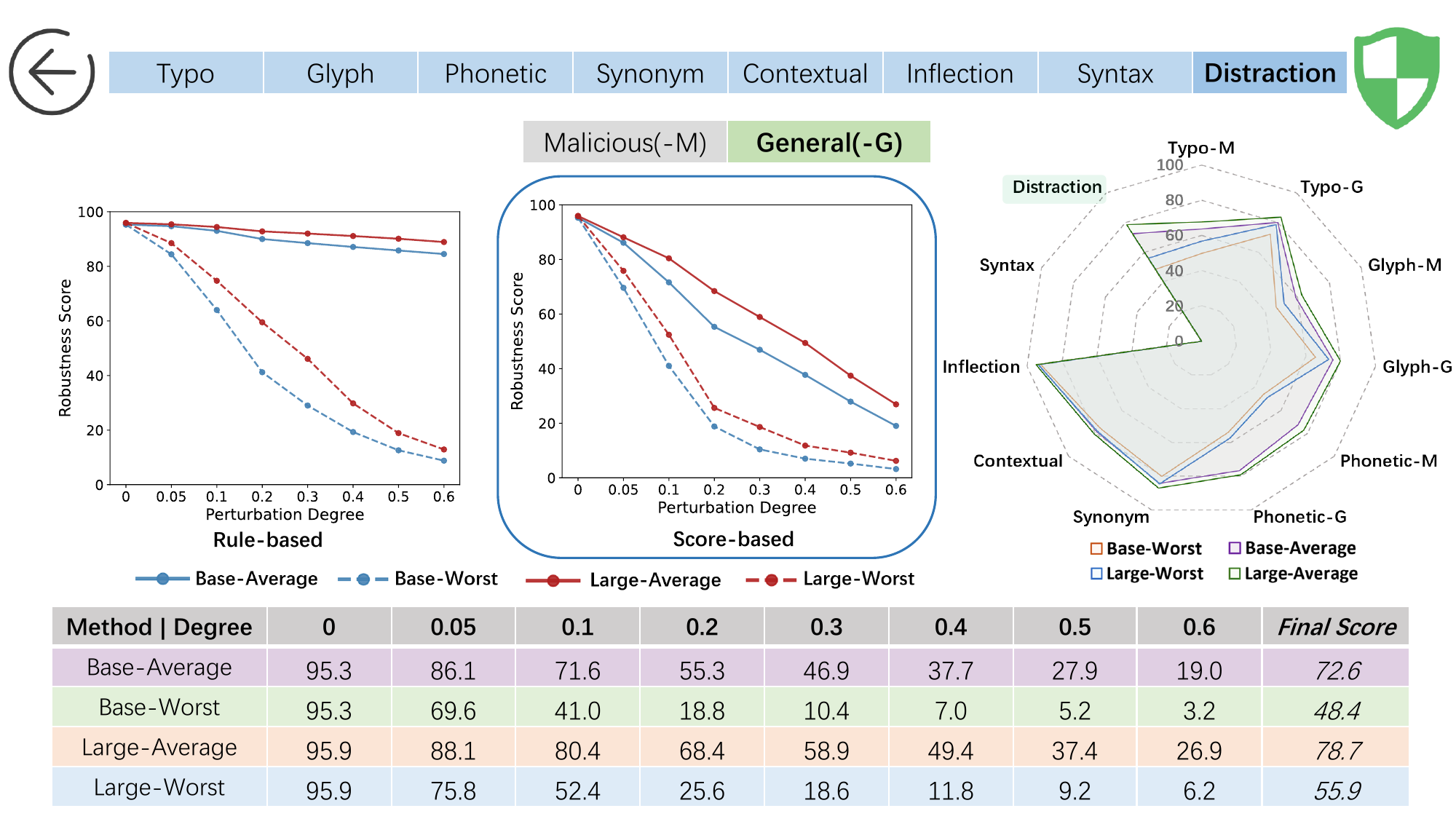}}
\end{subfigure}
\vspace{-1.25mm}

\centering
\caption{\label{fig:robust_com_score} Robustness comparison report for score-based evaluation on SST-2.}
\end{figure*}

We show the robustness report that compares the two models' robustness in Figure~\ref{fig:robust_com_rule} (rule-based evaluation) and Figure~\ref{fig:robust_com_score} (score-based evaluation).

\newpage

\end{document}